\documentclass[11pt,a4paper,twoside,openright]{report}
\usepackage[a4paper,inner=3.3cm,outer=2.4cm,top=3.5cm,bottom=3.5cm,pdftex]{geometry}

\usepackage[british]{babel}
\usepackage{graphicx}
\usepackage{xcolor}
\usepackage{wrapfig}
\usepackage[tight]{subfigure}
\usepackage{url}
\usepackage{excludeonly}
\usepackage{fancyhdr}
\usepackage[boxed,linesnumbered,algochapter]{algorithm2e}
\usepackage{setspace}
\usepackage{enumitem}
\usepackage{amssymb}
\definecolor{dark}{rgb}{0,0,0.6}
\usepackage[bookmarks,linkcolor=dark,citecolor=dark,urlcolor=dark,colorlinks,breaklinks,pdfpagelabels,pdftitle={Chromatic Aberration Recovery on Arbitrary Images},pdfauthor={Daniel J Blueman}]{hyperref}
\usepackage[Lenny]{fncychap}
\usepackage[sort,numbers]{natbib}
\usepackage{pgfplots}
\usepackage{tikz}
\usepackage{ctable}
\usepackage{multicol}
\usetikzlibrary{shapes,arrows,calc}

\usepackage{pxfonts}
\usepackage[T1]{fontenc}

\renewcommand{\MakeUppercase}[1]{#1}

\begin{document}
\pagenumbering{roman}


\begin{titlepage}
\begin{center}
\textbf{\huge Chromatic Aberration Recovery}\\ 
\vspace{0.3cm} 
\textbf{\huge on Arbitrary Images}\\
\vspace{2cm}
\textsc{by\\Daniel J Blueman}
\vspace{6.5cm}
\\
\includegraphics[width=0.35\textwidth]{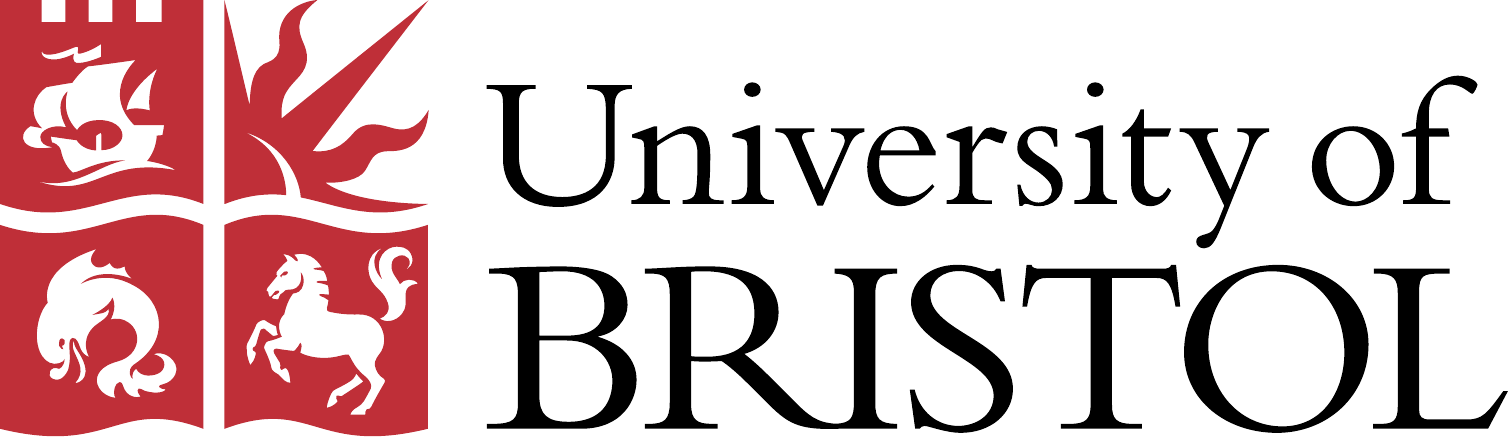}
\vspace{1cm}
\\
\textsc{A dissertation submitted to the University of Bristol \\
in accordance with the requirements of the degree \\
of Master of Research in the Faculty of Engineering} \\

\vfill
\textsc{September 2011}
\\
\vspace{1cm}
\textsc{Department of Computer Science}

\end{center}
\vspace{1cm}
\end{titlepage}

\pdfbookmark[0]{Abstract}{Abstract}
\section*{Abstract}
Digital imaging sensor technology has continued to outpace development in optical technology in modern imaging systems. The resulting quality loss attributable to lateral chromatic aberration is becoming increasingly significant as sensor resolution increases; other classes of aberration are less significant with classical image enhancement (e.g.\ sharpening), whereas lateral chromatic aberration becomes more significant. The goals of higher-performance and lighter lens systems drive a recent need to find new ways to overcome resulting image quality limitations.

This work demonstrates the robust and automatic minimisation of lateral chromatic aberration, recovering the loss of image quality using both artificial and real-world images. A series of test images are used to validate the functioning of the algorithm, and changes across a series of real-world images are used to evaluate the performance of the approach.


The primary contribution of this work is introduced: a novel algorithm to robustly minimise lateral chromatic aberration in both calibration and real-world images. This is broken down into discrete steps and detailed. The basis of the algorithm uses chromatic correspondences to converge a set of distortion coefficients. Other contributions are then introduced: a second algorithm is developed to allow correction information to be correlated to the lens model and parameters. This information is subsequently stored in a database to allow offline correction of unseen images. Finally, an algorithm is developed to measure image fidelity in a way more relevant to how the Human Visual System processes information via spatial frequency analysis.

Algorithm validation is conducted through a series of steps to ensure correctness, then artificial and test images are analysed, and the quantification algorithm is applied to measure improvement. Lastly, the performance of this system is compared against prevailing methods and analysed.
\newline\newline
\textbf{Keywords:} chromatic aberration, aberrations, lens systems, image fidelity

\thispagestyle{empty}\newpage
\setcounter{page}{3}


\pdfbookmark[0]{Acknowledgements}{Acknowledgements}
\section*{Acknowledgements}
This work was enabled by the guidance of my supervisor, Professor Majid Mirmehdi, whom I extend my deepest gratitude to.

Additionally, I would like to thank all my friends and colleagues at the University of Bristol for their guidance and support in the Computer Graphics and Computer Vision labs, including Timo Kunkel. Thanks also to Sritrakool Waeladee for useful feedback from another discipline.

Finally, I am forever grateful to my family and friends for their moral support.
\thispagestyle{empty}\newpage

\section*{Declaration}
\pdfbookmark[0]{Declaration}{Declaration}
I declare that the work in this dissertation was carried out in accordance with the requirements of the University's Regulations and Code of Practice for Research Degree Programmes and that it has not been submitted for any other academic award. Except where indicated by specific reference in the text, the work is the candidate's own work. Work done in collaboration with or with the assistance of others is indicated as such. Any views expressed in the dissertation are those of the author.
\vspace{25mm}{\newline Daniel J Blueman}
\thispagestyle{empty}\newpage

\section*{License}
\pdfbookmark[0]{License}{License}
This thesis and associated software is licensed under the Creative Commons Attribution-NonCommercial-ShareAlike 3.0 Unported (CC BY-NC-SA 3.0) public license, available at \href{http://creativecommons.org/}{creativecommons.org}.
\thispagestyle{empty}\newpage

\pdfbookmark[0]{Table of Contents}{Table of Contents}
\tableofcontents
\pdfbookmark[0]{List of Figures}{List of Figures}
\listoffigures
\begingroup
\let\cleardoublepage\relax
\pdfbookmark[0]{List of Algorithms}{List of Algorithms}
\listofalgorithms
\pdfbookmark[0]{List of Tables}{List of Tables}
\listoftables
\endgroup

\newpage
\pdfbookmark[0]{Abbreviations}{Abbreviations}
\section*{Abbreviations}
\begin{description}[leftmargin=2.5cm,style=sameline]\itemsep0pt
\item[ACA]	Axial Chromatic Aberration
\item[AHD]	Adaptive Homogeneity-Directed demosaicing algorithm
\item[BFGS]	Broyden-Fletcher-Goldfarb-Shanno
\item[CA]	Chromatic Aberration
\item[CCD]	Charge Coupled Device imaging sensor
\item[CFA]	Colour Filter Array
\item[CIE]	Commission Internationale de l'Eclairage (International Lighting Commission)
\item[CMOS]	Complimentary Metal Oxide Semiconductor
\item[DFT]	Discrete Fourier Transform
\item[DMD]	Digital Micromirror Device
\item[EXIF]	EXchangeable Image File-format
\item[FPGA]	Field Programmable Gate Array
\item[GPU]	Graphics Processing Unit
\item[HVS]	Human Visual System
\item[L-BFGS-B]	Limited-memory Broyden-Fletcher-Goldfarb-Shanno-Bounded
\item[LCA]	Lateral Chromatic Aberration
\item[PPG]	Patterned Pixel Grouping demosaicing algorithm
\item[RGB]	Red-Green-Blue
\item[SIFT]	Scale-Invariant Feature Transform
\item[SLM]	Spatial Light Modulator
\item[SLR]	Single Lens Reflex camera system
\item[UD]	Ultra-low Dispersion
\item[VNG]	Variable Number of Gradients demosaicing algorithm
\end{description}

\onehalfspacing

\cleardoublepage
\chapter{Introduction}
\label{introduction}
\pagenumbering{arabic}
\section{Motivation}
Consumers, hobbyists and professionals are pushing photography into areas that were traditionally inaccessible. The expectation and latent demand for wide-aperture and wide-angle lenses, while retaining high-fidelity is at previously unprecedented levels.

Modern lens design devolves to a series of cost, size and performance tradeoffs, nominally leading to an intermediate balance of these factors. Fixed focal length prime lenses have movable elements only for focusing, however optical and design complexity increases substantially with variable focal length zoom lenses which are commonplace in consumer compact cameras. Before 2005, none of the high-zoom-ratio lenses available today were developed. Primarily for convenience purposes, these are used to replace a selection of prime\slash small-ratio lenses; a significant reduction to fidelity is commonly lost for this convenience.

There is a continual drive to reduce costs and increase margin; reduced complexity, material costs and weight are clear economic competitive advantages. Due to this, there has been a growing trend to move hardware camera functionality to in-camera software instead, reducing cost, increasing flexibility and marketable features. The most significant example of this would be contrast-based auto-focus; this is computed as a function over the image frame and allows mechanical focus seeking to maximise the subject's captured spatial frequency. Without this, specialised auto-focus sensor chips are needed along the optical path, increasing system complexity and costs, and attenuating available light arriving at the image sensor, e.g.~due to mirrors or prisms.

Software implementations clearly offer far more flexibility. For example, zone-based or face auto-focus and exposure are readily possible; these were traditionally infeasible due to the minimal processing power inside the camera. With these restrictions lifting, the wide scope for creating a more flexible and optimised imaging platform able to fully exploit both hardware and software is available.

As sensor resolution has been rapidly increasing, the optical performance has only marginally increased, thus detail attenuated in the lens system and thus the efficiency of the whole image-capture system is reduced, often significantly at more extreme parameters.

Lens performance is constrained by a number of well-understood areas of optical distortion~\cite{Jacobson00a}. Presently, barrel (geometrical) distortion is well understood and has robust solutions available; chromatic aberration previously manifested at the sub-pixel level --- consequently, there is far less literature and evolution of solutions. As sensor resolution increases, the relative impact of chromatic aberration also increases. In general, other types of aberration pose lower limitations to image quality and more importantly are less sensitive to sensor resolution. Consequently, this thesis will consider the problem of quality loss resulting from chromatic aberrations and will set out to address this area of loss.

In this work, the term \textit{aberration} is generally defined as the departure from a theoretically ideal lens system. By extension of this, \textit{image fidelity} is defined as an image with the minimum loss due to aberration, that is free of aberration to the fullest extent. \textit{Monochromatic} is defined as features existing at all visible wavelengths equally, such that they are invariant of chromaticity.

From the specific areas covered, there is a clear deficit that of no current solution which provides automatic, robust and adaptive correction for the penalty lateral chromatic aberration imposes on general lenses; this research aims to address this gap.

\section{Chromatic Aberration}
Loss in image fidelity is due to a known set of aberrations, classified as either a \textit{monochromatic} or \textit{chromatic} type of aberration. They occur due to the physical interaction of light with materials, lens design constraints and manufacture limitations. Von Seidel rigorously defined the monochromatic types of aberration to be astigmatic, comatic, field curvature, distortion and spherical \cite{Seidel56}. The chromatic type of aberration exists independently from monochromatic aberration, and is broken down into two further types: Axial or longitudinal Chromatic Aberration (ACA) and Lateral or transverse Chromatic Aberration (LCA). LCA can introduce pronounced undesirable image artifacts as shown later, partially since most lens systems correct for two spectra rather than the three sampled by the sensor, due to complexity and cost limitations~\cite{Kingslake78}. Since the effect of LCA cannot be minimised in practice unlike with ACA, this work will focus on addressing the intrinsic fidelity loss due to LCA, exclusive of other types of aberration.

In order to fully differentiate both LCA and ACA types of chromatic aberration, they are subsequently defined; other forms of aberration are monochromatic and outside the scope of this work and are not considered further. Chromatic aberration is the result of the refractive index of the lens medium (typically some form of glass) varying with the transmitted light wavelength, termed \textit{dispersion}~\cite{Newton04}; this is the definition of \textit{prismatic} behaviour.

With ACA, the focal point varies along the optical axis with the light's wavelength. As a consequence, the focal length of an object cannot be exactly coincident in all three image planes, but only approaching this crucial ideal. As light strikes the sensor plane, out of focus rays contribute to a \textit{circle of confusion} or \textit{bokeh}, which manifests as a subtle coloured halo around the boundary of an object in more extremal circumstances \textit{such as at the lens's widest aperture setting}. The resulting introduction of image artifacts is minimised as the lens aperture is \textit{stopped down} or reduced due to the increase in depth of field bringing the axially-misaligned focal points nearer. Many modern digital cameras when in fully-automatic mode, balance the aperture size preventing significant spatial frequency loss due to photon diffraction \cite{Mielenz99} and a shallow depth of field, increasing focus selectivity and focus error as a side-effect. Due to the camera's automatic behaviour, ACA is nominally minimised to imperceivable levels. Further, photographers generally reserve the use of large apertures for shooting portraiture and low-light pictures where maximum light flux is needed; lenses suitable for these situations with larger maximum apertures, typically have stronger compensation for ACA\@.

\begin{figure}\centering
\def\svgwidth{0.6\textwidth}\input{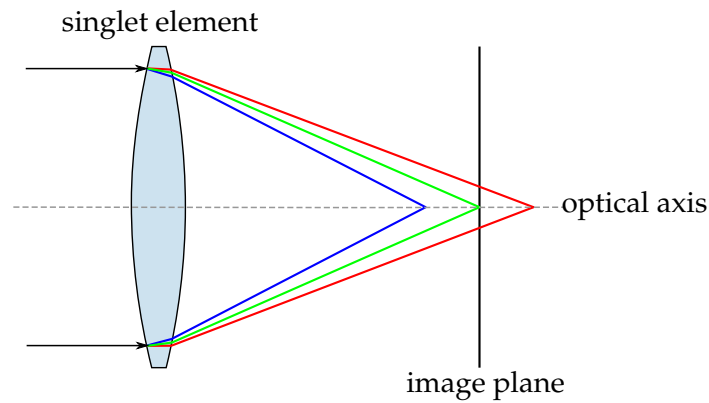}
\caption{Axial Chromatic Aberration}\label{fig:aca-diagram}\end{figure}

\begin{figure}\centering
\subfigure{\fbox{\includegraphics[width=0.2\textwidth,height=150pt,keepaspectratio=true]{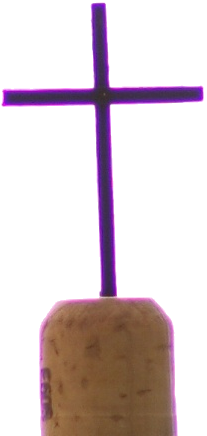}}}
\hspace{60pt}
\subfigure{\fbox{\includegraphics[width=0.2\textwidth,height=150pt,keepaspectratio=true]{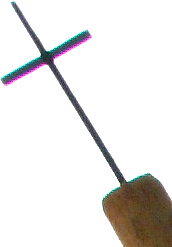}}}
\caption[Axial and Lateral Chromatic Aberration]{Axial (left) and Lateral (right) Chromatic Aberration, adapted from Walree~\cite{Walree97}}
\label{fig:ca-examples}
\end{figure}

Figure~\ref{fig:aca-diagram} shows the focal lengths of the primary wavelengths at different displacements along the optical axis due to ACA\@. Walree~\cite{Walree97} gives a clear example of this, showing the effects of ACA when increasing the focal depth sensitivity by means of opening the aperture, given in Figure~\ref{fig:ca-examples}. Due to the ability to minimise ACA with lens parameters at picture capture time, this typically does not pose a problem in practice.

For aforementioned reasons, ACA will not be considered further in this work. This research will address the more nascent issue of LCA within digital photography.

\section{Lateral Chromatic Aberration}
\begin{figure}\centering
\def\svgwidth{0.6\textwidth}\input{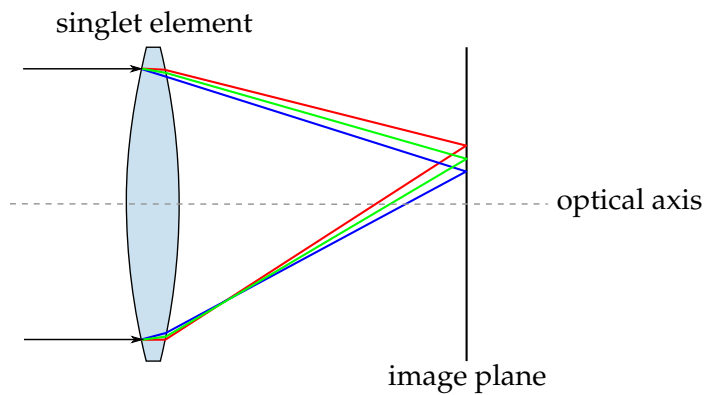}
\caption{Lateral Chromatic Aberration}\label{fig:lca-diagram}\end{figure}

LCA is the relative and non-linear displacement of the three colour planes, \textit{across the image plane}, conceptually shown in Figure~\ref{fig:lca-diagram}. The result of this chromatic distortion introduces undesired artifacts and thus information, which leads to a perceived detail loss in the image, due to the misaligning of coincident features throughout the image. Less perceptible impact in lower-contrast areas reduces texture detail and generally tends to reduce the perception that LCA compromises image quality.

LCA occurs along the \textit{sagittal} or \textit{oblique} direction to the optical axis. Figure~\ref{fig:ca-examples}, a top left image crop, illustrates that LCA only affects detail running orthogonal to the line through the centre of the image, due to the different distortion paths for the red, green and blue planes.

\begin{figure}\centering
\subfigure{\includegraphics[width=\textwidth]{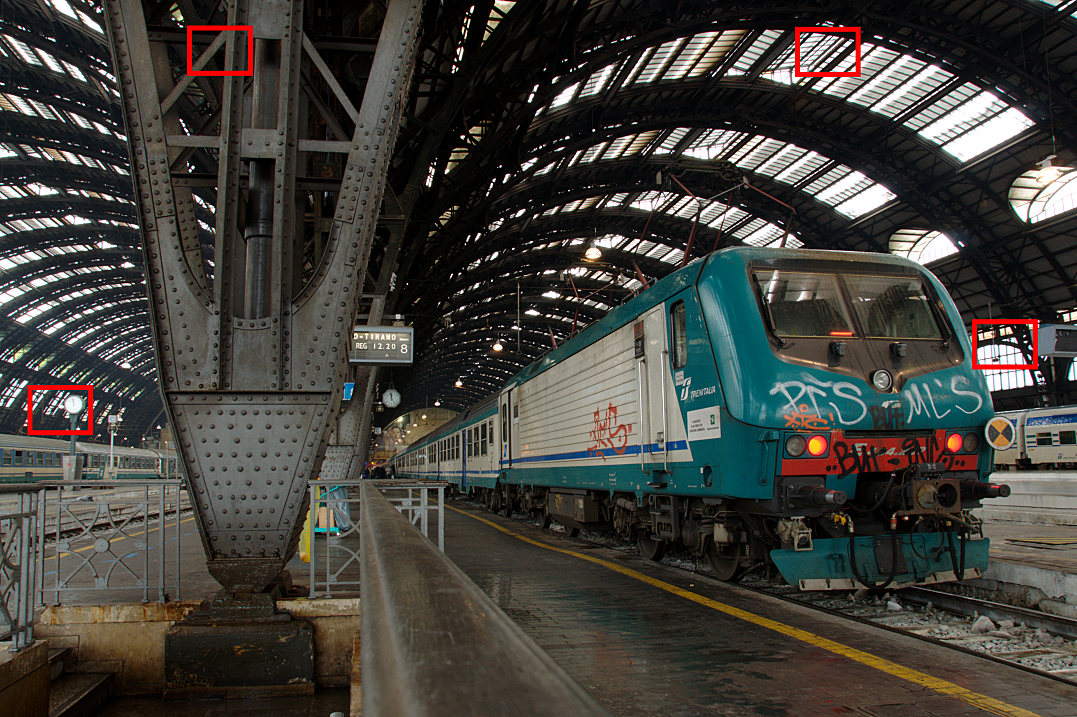}}\vspace{-7pt}
\subfigure{\includegraphics[width=0.4960\textwidth]{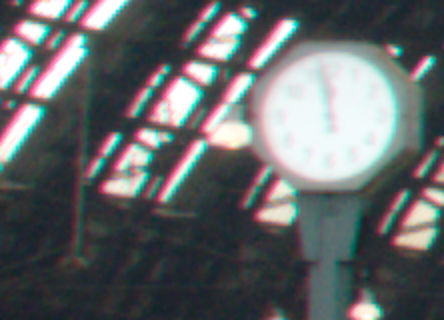}}
\vspace{-7pt}
\subfigure{\includegraphics[width=0.4960\textwidth]{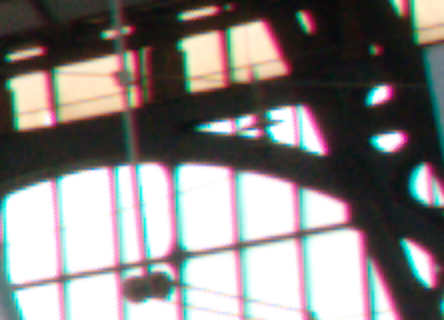}}
\subfigure{\includegraphics[width=0.4960\textwidth]{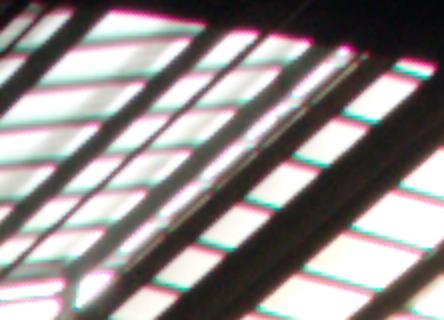}}
\vspace{-7pt}
\subfigure{\includegraphics[width=0.4960\textwidth]{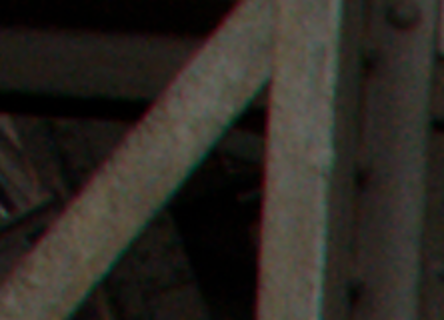}}
\caption{Cropped sections exhibiting LCA}
\label{fig:lca-example}
\end{figure}

Taking an example shown in Figure~\ref{fig:lca-example}, shot in near-ideal real world conditions with reduced aperture, no image shake and correct focus, we see crops from various zones of the image showing significant LCA, estimated at around an order of magnitude higher error than the Nyquist limits of the sensor, i.e.~up to around 10 pixels. The frame was taken at the optimal aperture of f/8.0 at 18mm focal length (equivalent to 27mm on a full-frame sensor) on a Nikon D90 digital SLR with a Nikon 18--200mm VR lens, considered as basic professional equipment. It is clear that chromatic aberration limits image detail, even without spatial frequencies approaching the Nyquist \cite{Shannon49} limits of the Bayer sensor array. Since the crops show the chromatic aberration being asymmetric, LCA is therefore the dominant aberration and attenuating quality. ACA occurs symmetrically to the image feature and is shown to be far less significant.

\section{Central argument and research objective}
The scope of this work is the correction of LCA using typical camera imaging systems, and thus the most common lens designs and optical geometries, such as large aperture, wide field-of-view angle and high zoom ratio goals; this can be achieved only in digital \textit{post-processing} steps. It is assumed that LCA can be sufficiently addressed by the generalised distortion model. Correction for \textit{optical decentering} is discussed later. Though this would allow analysis and correction with \textit{tilt-shift} lenses, this is outside the scope of this work, as these are comparatively infrequently used. Further, careful consideration will be employed to avoid introducing additional user or system constraints, or assumptions about the underlying system; this is a requirement for a robust system.

Consequently, the primary objective is to develop a \textit{robust}, \textit{generalised} algorithm able to perform across all real-world shooting conditions, and without manual adjustment or input. Secondary objectives are to develop a method to tangibly quantify image quality, and demonstrate the quality improvement due to CA correction. Subsequent to this, analysis will be conducted with synthetic images, and later with a typical imaging platform with real-world images. Lastly, this implementation will be compared to another widely-used solution of manual correction using Adobe Photoshop CS4. Algorithms presenting a new correction technique and an algorithm to analyse image quality improvement will be developed and evaluated throughout this work.

\section{Contributions}
In this work, the minimisation of LCA by means of development of a robust and automatic algorithm is assessed as a novel contribution in this field. An algorithm to quantify distortion due to LCA is proposed and developed, allowing determination of the optimal correction parameters and measurement of recovery of fidelity loss. This is applied and evaluated to understand algorithm robustness and correction performance. Methodology and workflow will be demonstrated showing the ability to automatically correct real-world pictures, allowing a closed-loop solution. Hence, these research findings are expected to directly benefit those who are interested in this area at large.

In the next chapter, the context for the work will be presented with concepts and definitions. Then, contributed work will be assessed relative to the objective of this work.

\section{Overview of methodology}
\begin{figure}\centering
\linespread{0.8}
\begin{tikzpicture}[
	node distance = 7em,auto,
	store/.style={cylinder, shape border rotate=90, draw=blue, very thick, text width=4em, text centered, inner sep=1pt, draw=green!50!black!50,top color=white,bottom color=green!50!black!20}, shape aspect=.35,
	data/.style={tape, tape bend top=none, minimum size=1em, very thick, draw=red!50!black!50, top color=white, bottom color=red!50!black!20, font=\itshape, text width=4.4em, text centered},
	line/.style={draw, thick, -latex', shorten >=2pt},
	process/.style={rectangle, minimum size=1em, rounded corners=3mm, very thick, draw=blue!30!black!50, top color=white, bottom color=blue!30!black!20, text width=4.4em, text centered},
	point/.style={coordinate}, >=stealth', thick, draw=black!50,
	tip/.style={->,shorten >=1pt}, every join/.style={rounded corners},
	dline/.style=thick, loosely dashed
	label/.style={}
]

\draw (5,-0.4) node[draw, thick, dashed, minimum height=2.7cm, minimum width=13cm, rounded corners] {};
\draw (5,-3.7) node[draw, thick, dashed, minimum height=3.2cm, minimum width=13cm, rounded corners] {};

\draw (9.0,-0.45) node {recovery phase};
\draw (0.7,-3.55) node {correction phase};

\node[data] (A) {source pictures};
\node[process, right of=A] (B) {coefficient recovery};
\node[store, below right of=B] (C) {parameter database};
\node[process, below right of=C] (D) {correction};
\node[data, right of=D] (E) {corrected pictures};
\node[data, left of=D] (F) {source pictures};

\node[left of=C, node distance=15em] (X) {};
\node[right of=C, node distance=15em] (Y) {};

\path[tip] (A) edge (B);
\path[tip] (B) edge (C);
\path[tip] (C) edge (D);
\path[tip] (D) edge (E);
\path[tip] (F) edge (D);

\end{tikzpicture}
\caption{Image correction method overview}\label{fig:architecture}\end{figure}
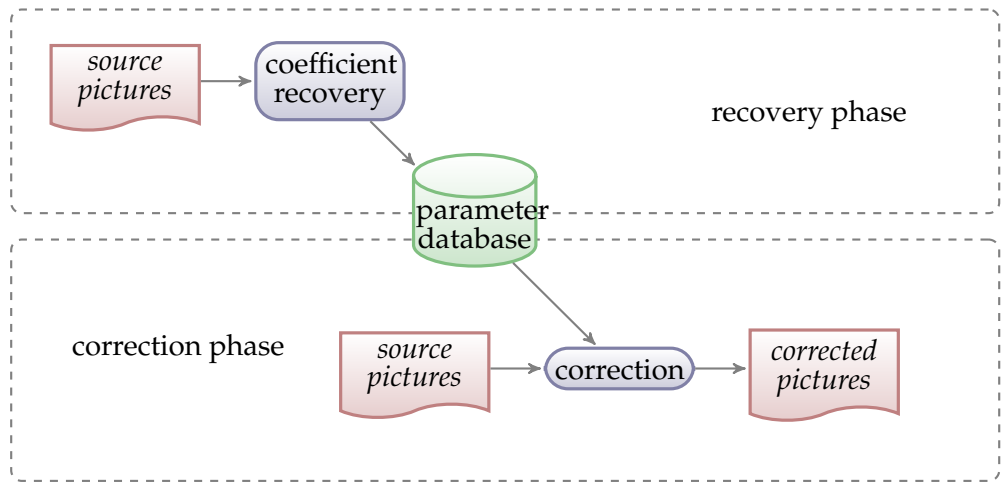

The methodology that will be presented is characterised by the scheme shown in Figure~\ref{fig:architecture}. Overall, there are two phases of operation: firstly, the coefficient recovery phase, where correction data is computed from images. Secondly, this correction data can be used to correct previously unseen images, using data extracted from the database, resulting in images corrected for LCA\@.

\section{Thesis layout}
Chapter~\ref{cha:background}, \textit{Background} introduces concepts and equations to be used in later chapters and looks at related constraints, presenting complete definitions. Related work in this field is grouped and analysed relative to the objective of this work --- robust correction of LCA\@. Existing works are revealed according to Chapter~\ref{cha:methodology}, \textit{Proposed method} which presents high and low-level details of the algorithms specialised to LCA correction, and breaks down each of the processing steps. Chapter~\ref{cha:evaluation}, \textit{Methodological evaluation} defines technique to show steps taken to validate increasing levels of capability. Finally, Chapter~\ref{cha:future}, \textit{Conclusion and further work} discusses research findings, along with areas for future consideration.

\cleardoublepage
\chapter{Background}
\label{cha:background}
\section{Introduction}
This chapter aims to present major concepts and information relevant to the understanding, measurement and correction of Lateral Chromatic Aberration. These concepts will be used in later chapters.

The chapter is split into two important intertwined parts. To create better understanding of the research topic, the first part will provide general knowledge relevant to LCA, which includes optics, paraxial optics, lens systems, optical properties, sensor properties, sensor blooming, image EXIF data and parameters, lens variance, correction overview and test images. To highlight the need of this research, the second part will review existing literature on LCA and areas related to this. There are two types of image correction which will be encountered: in-lens and post-processing. Post-processing correction will later be examined in more detail, since the former is only useful if control is available at the lens physical level. Post-processing affords the freedom to arbitrarily change characteristics, required for the intended use of this research. Subsequently, the knowledge gap found will be discussed.

\section{Optics}
The field of optics encompasses a broad group of fields based around transmission, modulation and sampling of electromagnetic spectra not limited to visible light. 

Concepts and equations directly relevant to this research are subsequently introduced, and later employed.

\subsection{Paraxial optics}
\begin{figure}\centering
\def\svgwidth{0.5\textwidth}\input{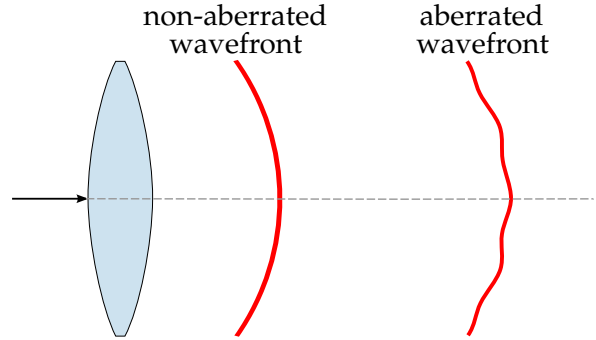}
\caption{Paraxial approximation}\label{fig:paraxial}\end{figure}

Paraxial (lit. \textit{parallel-axial}) optics model light as traversing optical elements within an optical system with an angle of incidence from the optical axis considered small \cite{Alda03}. Due to this low-angle approximation, wavefronts propagating through the optical system are modelled as spherical chords, thus radiating equally at all angles therefore free from \textit{aspherical} aberration \cite{Alda03}, as illustrated in Figure~\ref{fig:paraxial}.

As the wavefront is modelled as a two-dimensional \textit{chord} of a sphere, the standard trigonometric $\sin$, $\cos$ and $\tan$ functions are used. The known Taylor-series expansions for these are:

\begin{eqnarray}
\sin \theta &=& \theta - \frac{\theta^3}{3!} + \frac{\theta^5}{5!} - \frac{\theta^7}{7!} + \dots \nonumber \\
\cos \theta &=& 1 - \frac{\theta^2}{2!} + \frac{\theta^4}{4!} - \frac{\theta^6}{6!} + \dots \nonumber \\
\tan \theta &=& \theta + \frac{\theta^3}{3} + \frac{\theta^5}{5} + \frac{\theta^7}{7} + \dots
\label{eqn:trig-taylor}
\end{eqnarray}

Discarding orders higher than one in Equations~\ref{eqn:trig-taylor}, the approximation becomes:

\begin{eqnarray}
\sin \theta &\approx& \theta \nonumber \\
\tan \theta &\approx& \theta \nonumber \\
\cos \theta &\approx& 1
\end{eqnarray}

Applying this to the lens model, $\theta$ is the ray angle from the optical axis; lens systems with these assumptions are termed \textit{first order optics}, since they are valid up to only the first order approximation.

\subsection{Lens systems}
\begin{figure}\centering
\def\svgwidth{0.5\textwidth}\input{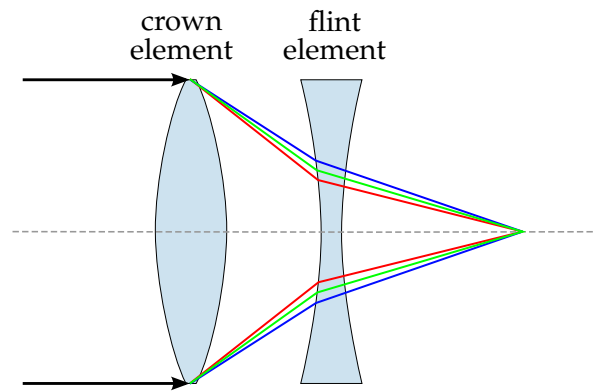}
\caption{Dual thin lens (``doublet'') with CA correction}\label{fig:doublet}\end{figure}

A lens is defined as one of more groups of lens elements through which light is focussed. Single Lens Reflex (SLR) camera lens systems typically have four to twenty lens elements in a number of moveable groups \cite{Jacobson00b}. Lens elements are grouped to minimise \textit{dispersion}, which occurs where different wavelengths of light are refracted through the medium at angles varying with distance from the optical centre. Since dispersion is a physical property of the material, it cannot be eliminated. Lens designers strategically employ lens elements with multiples types of glass, with intentionally differing \textit{Abbe} numbers \cite{Kingslake78}, also known as dispersion or refractive index. When elements are placed in a pair, cemented back-to-back, the number of medium transitions is reduced and thus light loss due to reflection; significant dispersion angles can be avoided. Such an arrangement is termed a \textit{doublet} and illustrated in Figure~\ref{fig:doublet}. The second element in the pair is often selected to disperse with the opposite sign of the first. \textit{Flint} and \textit{crown} glass types are most commonly used \cite{Jacobson00a} for this purpose. Techniques of applying an anti-reflective coating are used on selected elements to minimise additional light loss at medium transitions \cite{Peres07}.

From these implementation constraints, three resulting compromises occur: firstly, there are material choice constraints, secondly, at worst there are twice as many medium transitions for propagating rays with using doublets, and lastly, there are additional manufacturing requirements such as precision and surface geometry, where the pair of elements are interfaced together forming a doublet. An additional set of issues arise as a result: there is increased light loss from the more frequent medium transitions, secondly, there is also more light scattering from surface reflection \cite{Hallyn94}, and a single optimal material cannot be selected alone, due to uncontrolled dispersion.

In order to understand light loss due to surface reflection, the angle of refraction can be computed using Snell's law from Figure~\ref{fig:snell} and:

\begin{figure}\centering
\def\svgwidth{0.25\textwidth}\input{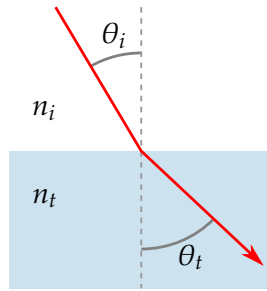}
\caption{Snell's law}\label{fig:snell}\end{figure}

\begin{equation}
\frac{n_i}{n_t} = \frac{\sin \theta_i}{\sin \theta_t}
\end{equation}

where $\theta_i$ is the angle from the surface normal of the incident ray, $\theta_t$ is the angle of the traversing ray, $n_i$ is the refractive index of the primary medium, and $n_t$ is the refractive index of the secondary medium.

Using fluorite lens elements was later found to minimise CA due to its lower refractive index than other types of glass \cite{Peres07}. Since fluorite is brittle and requires significantly more mechanical work to grind, it features in typically one lens element in high-end lenses due to the significant cost premium. `Ultra-low Dispersion' (UD) lens elements were developed as a way to get lower refraction by chemically doping normal glass to lower the refractive index, at a lower cost; this allowed entry into semi-professional lenses, where one or two UD lens elements are incorporated to reduce CA\@. In more recent times, special lens elements using diffraction grating on stock glass have been employed \cite{Tygier05}; however, they require premium and expensive manufacturing techniques only more recently available \cite{Canon00}. The cost of lenses incorporating such developed elements is beyond consumer reach at large.

Today, the use of diverging-converging doublets termed \textit{achromatic doublets}, shown in Figure~\ref{fig:doublet}, remains the most ubiquitous way of minimising CA \cite{Smith27}, yet this is known to be an imperfect solution due to three reasons: firstly, the doublet optimises correction for two primary wavelengths of red and blue, around the green wavelength, leaving an uncorrected \textit{secondary spectrum} \cite{Jacobson00a}. Secondly, cheaper and\slash or simpler lenses are modelled using paraxial and \textit{thin lens} approximations, termed Gaussian optics, which are suitable for applications where quality is not the primary concern \cite{Jacobson00a}. These modelling steps result in first-order residual error. More sophisticated lens designs employ computational ray-tracing through the lens system, such as offered in some commercial lens design software, such as Sinclair Optics' OSLO~\cite{Sinclair02}. Lastly, the number of interfacial surfaces is therefore greater (four, rather than three for a doublet) as previously mentioned, increasing light loss.

\subsection{Lens parameters}
With the construction of lenses, a number of element groups are designed to move with manual linkage or electrical control. Two parameters are observed from element movement, and a third invariant of element movement. Firstly, the lens \textit{focal length} relates to the magnification of the image. Secondly, the \textit{focussing distance} allows selection of the focus plane. Lastly, the \textit{aperture} size can be changed to vary the \textit{depth of field}. All three adjustments affect the path of light, so are recorded in the image metadata, thus available.

\subsection{Optical properties}
Due to the simplicity and consequent limitations of \textit{paraxial optics}, a more complete framework to model lens systems was sought. In 1956 Ludwig von Seidel developed a model defining various types of orthogonal aberrations \cite{Seidel56}, allowing aberrations to be decomposed into constituent types and expressed independently.

The complete Seidel equation \cite{Kidger01} using Zernike wave-front notation \cite{Welford86} is:

\begin{eqnarray}
W(r, \phi) &=& w_{020} \ r^2 \nonumber \\
	&+& w_{040} \ r^4 \nonumber \\
	&+& w_{131} \ \eta \ r^3 \cos \phi \nonumber \\
	&+& w_{222} \ \eta^2 \ r^2 \cos^2 \phi \nonumber \\
	&+& w_{220} \ \eta^2 \ r^2 \nonumber \\
	&+& w_{311} \ \eta^3 \ r \cos \phi \nonumber \\
\label{eqn:seidel-equation}
\end{eqnarray}

where $\eta$ is object size, $r$ is displacement from the optical centre, $\phi$ is the angle clockwise from vertical of the ray, and $w_{ijk}$ is the coefficient describing the magnitude of aberration. Where only distortion aberration is present, a number of terms become zero, leaving:

\begin{equation}
W = w_{311} \ \eta^3 \ r \cos \phi
\end{equation}

Following on, it is known that the cosine function can be written as an infinite Taylor series:

\begin{equation}
\cos \phi = 1 - \frac{\phi^2}{2!} + \frac{\phi^4}{4!} - \frac{\phi^6}{6!} + \frac{\phi^8}{8!} - \dots \\
\end{equation}

This becomes intractable, so a \textit{finite} number of terms in the Taylor series are used, approximating the cosine function, here up to the fourth order:

\begin{equation}
\cos \phi \approx \phi - \frac{\phi^2}{2!} + \frac{\phi^4}{4!} \\
\end{equation}

The \textit{Seidel approximation} proves the ability to mathematically model a set of lens aberrations with a Taylor series. Since this work applies to general, aberrated lenses, it is not possible to make assumptions about lack of particular aberrations, thus the only solution is to use the \textit{generalised Taylor series}:

\begin{eqnarray}
r_{dest} &=& a {r_{src}} + b {r_{src}}^2 + c {r_{src}}^3 + d {r_{src}}^4 + \dots \nonumber \\
\label{eqn:seidel-approx}
\end{eqnarray}

where $r_{src}$ is the image sensor plane, and $r_{dest}$ is the corrected image plane. $a$ is the overall scaling factor, and, $b$ and $c$ exert influence on the non-linear \textit{chromatic distortion} of the image plane. This can be used to non-linearly register one plane against another.

\subsection{Sensor properties}
In order to develop an imaging mechanism and thus a sensor that minimises information redundancy, Bayer~\cite{Bayer76} developed the concept of a Colour Filter Array (CFA) optimised to the spectral sensitivity of the Human Visual System (HVS). It is known that the eye has three photoreceptor \textit{cones} with peak sensitivity at three wavelengths, 564--580nm \textit{long}, 535--545nm \textit{medium} and 420--440nm \textit{short}, and twenty times the number of luminosity-sensitive receptors or \textit{rods} \cite{Wyszecki82}, sensitive to a wider spectral range and thus suited to low-light level vision, as they absorb wavelengths across a far greater band. Earlier research conducted by the \textit{Commission Internationale de l'Eclairage} (CIE) showed that the eye is more sensitive to the medium wavelength than the low or high, leading to the development of the \textit{photopic weighting function} $V(\lambda)$ \cite{CIE26,ITU02}:

\begin{eqnarray}
Y &=& V(\lambda) \nonumber \\
 &=& 0.299R + 0.587G + 0.114B
\label{eqn:photopic}
\end{eqnarray}

where $Y$ represents luminance of a pixel, $V$ is the photopic equation, giving the HVS response to light of frequency $\lambda$; $R$, $G$ and $B$ are the pixel values in the red, green and blue planes respectively. The Bayer array sensor thus maintains a quadrant grouping of elements, where two are sensitive to the medium wavelength, one to the short wavelength and one to the long wavelength. Specialising the elements to a particular light wavelength is achieved by a set of colour filters directly in front of the sensor. One negative side effect of filtering the light is that light flux illuminating the sensor elements is reduced by at minimum 66\% \cite{Maschal10}.

\begin{figure}\centering
\def\svgwidth{0.6\textwidth}\input{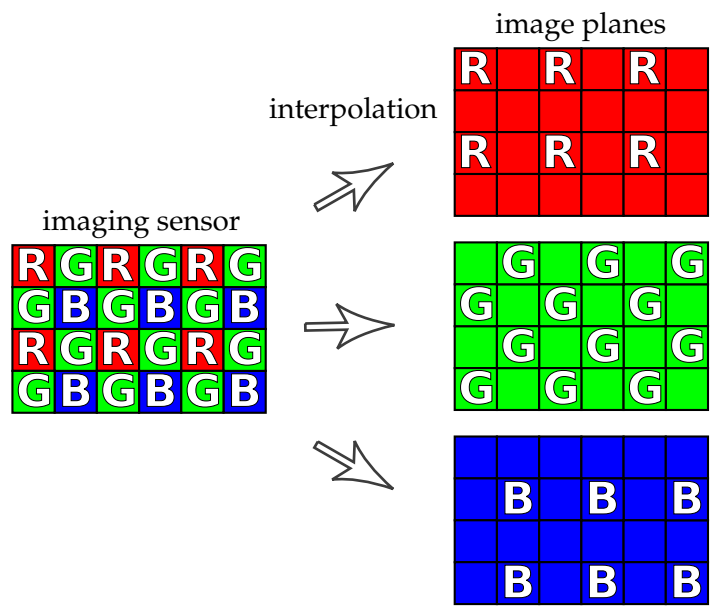}
\caption{Bayer demosaicing principle}\label{fig:bayer}\end{figure}

Processing steps are needed to reconstruct the CFA to the typical arrangement of stored image data of red, green and blue image values per pixel. Simple linear interpolation, shown in Figure~\ref{fig:bayer}, is known to leave chromatic edge artifacts \cite{Anagnostopoulos95}, so further algorithms have been developed such as Variable Number of Gradients (VNG) \cite{Cheung99}, Patterned Pixel Grouping (PPG) \cite{Hua04} and Adaptive Homogeneity-Directed (AHD) \cite{Hirakawa05} interpolation.

Other sensor types exist, such as the Foveon X3, which uses a stacked array of photodiodes \cite{Gamal02}. Since the CFA is ubiquitous, other sensor types are not considered in this work. The algorithms in this work equally apply and theoretically benefit from the higher chromatic spatial frequency information, allowing more accurate LCA recovery.

\subsection{Sensor blooming}
\begin{figure}\centering
\def\svgwidth{0.3\textwidth}\input{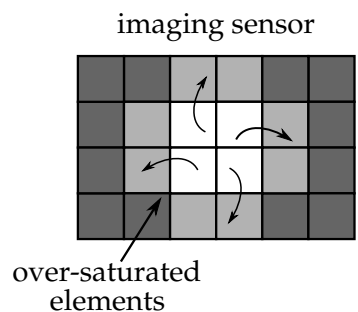}
\caption{CCD blooming principle}\label{fig:blooming}\end{figure}

The term \textit{blooming} applies exclusively to Charge Coupled Devices (CCDs) due to their implementation. This describes a specific situation that occurs: the charge caused by the potential stored at a CCD element overcomes the dielectric between one of more surrounding elements, and charge is transferred away or \textit{leaked}. The manifestation of this is typically that purple halos appear around areas which have experienced excessive saturation due to there being twice as many elements sensitive to green per unit area as blue or red. The mechanism is depicted in Figure~\ref{fig:blooming}.

Other types of imaging sensor, such as Complimentary Metal Oxide Semiconductor (CMOS) sensors are implemented in a different way and indirectly avoid this mechanism \cite{Fischer08}.

\section{Image EXIF data and parameters}
Image \textit{meta-parameters} are stored in the EXIF~\cite{JEITIA02} section in the image source file. This contains parameters which are not needed to correct the image but are essential to reusing the correction data later and secondarily storing useful state once correction is performed.

The camera body identifier and lens identifiers are used to select the right set of correction parameters; the focal length, aperture value and focus distance are affected by the position of the lens element groups, thus need to be known to find the nearest neighbour set of correction coefficients. The camera body orientation and image size information is needed to later store the lens decentering values, should this be calculated.

Subsequent to LCA correction analysis, a \textit{LCA corrected} flag can be set to indicate not to perform LCA correction on the image again, and the correction coefficients can be stored in the image file and database for later use.

\ctable[botcap,caption=Example stored EXIF tags and image parameters,label=image-params]{ll}{}{\FL
\multicolumn{2}{c}{From camera}					\ML
Lens identifier			& Nikon AF-S DX VR 18--200mm	\NN
Camera body identifier		& Nikon D90 SLR			\NN
Focal length			& 18.0mm			\NN
Subject (focus) distance	& 3100mm			\NN
Lens aperture value		& f/8.0				\NN
Image orientation		& Top-left			\NN
Image width			& 4352 pixels			\NN
Image height			& 2868 pixels			\ML
								\NN
\multicolumn{2}{c}{Calculated}					\ML
Computed image X-centre		& 2172 pixels			\NN
Computed image Y-centre		& 1432 pixels			\NN
LCA corrected			& False				\NN
LCA R-G correction coeff 1	& 0.99855155			\NN
LCA R-G correction coeff 2	& 0.0032236			\NN
LCA R-G correction coeff 3	& -0.00190334			\NN
LCA R-G correction coeff 4	& 0.00120401			\NN
LCA B-G correction coeff 1	& 1.00113412			\NN
LCA B-G correction coeff 2	& 3.02337291e-4			\NN
LCA B-G correction coeff 3	& -2.45039357e-3		\NN
LCA B-G correction coeff 4	& 1.76123214e-3			\LL}

\section{Lens variance}
Due to variance in manufacturing processes, there are sub-micron imperfections in lens element seating and alignment \cite{Smith08}. A tight Gaussian curve of variation is seen with quality control tests causing rework of lenses outside the tolerances. Once in consumer hands, wear in the guide grooves for the lens mechanics, particularly for zoom lenses, increases sensor data error. Ultimately, this causes lenses to slowly diverge from the original specification. Professional users have reported variation among particular lenses, with some units attenuating spatial frequency more than expected \cite{Cicala08}.

This field variance presents a key requirement: the correction data from one lens unit may not optimally correct another lens unit of the same model, so an adaptive technique which optimises for the behaviour of the lens from which images were taken is needed.

\section{Correction overview}
Correction of LCA with post-processing steps in a digital workflow is, at present, the only solution offering complete correction. Therefore, post-processing steps are slowly being incorporated to indirect (in-camera) processing \cite{Nikon08}, since it is simply not possible to design out LCA in a lens with more than one element. Various techniques have been developed from manual adjustment of dual-linear correction parameters using modern image processing software \cite{Bockaert06}, to more elaborate and calculated step-wise procedures \cite{Krause04}. Both approaches are error-prone due to the user's subjective observation feeding back into the adjustment process, resulting in a tendency to be inaccurate and slow. Moreover and critically, the work to correct a single image does not contribute useful correction values towards an accurate model of the lens due to lack of information re-use. Indeed, manual LCA correction by inspection relies on the user assessing the overall resulting picture and is inherently based on \textit{psychovisual} feedback, making this an unsuitable process for accurate and consistent data capture. Areas which are less perceptually apparent, e.g.~lower spatial-frequency textures, or lack of contrast \textit{edges} may be under-corrected.

The second key limitation post-processing introduces is that of complexity. A popular correction mechanism has two variables to manually adjust, which control linear scaling of the red plane relative to the green plane, and the blue plane relative to the green plane. Due to non-linearity in the lens system, this linear mapping is an approximation. Even with more complicated methods, accuracy is still limited and should be at the \textit{sub-pixel} level. Additional to this, if LCA correction is performed before demosaicing the sensor array, the advanced demosaicing algorithm will be able to use the \textit{true} edges and not edges introduced from LCA artifacts. Further, advanced demosaicing algorithms can introduce false colour aliasing \cite{Koren08}, which would change the true LCA-free minima. Non-linear interpolation, sub-sampling or sharpening may change spatial information in an undesirable way, such as emphasising high-frequency noise. To prevent this, the image should be converted from \textit{gamma corrected} to linear before interpolation and back after demosaicing.

A number of camera manufacturers (e.g.~Nikon \cite{Nikon07}) have recently introduced software LCA correction internal to some of their medium-level digital SLR cameras. This provides a practical integrated solution to tackle the issue of significant LCA on wide-angle lenses, and is a value-adding feature touted to improve image quality, giving an edge over competing products. Nevertheless, there are compromises which restrict the usefulness of this in practice. Firstly, when capturing maximum sensor detail for off-line processing, photographers shoot in raw image format, so in-camera LCA reduction is therefore not performed; it is only available when capturing in lossy JPEG format. Additionally, since processing is implemented on a minimal embedded platform, tradeoffs and approximations are employed to give reasonable processing time, and will be weighed against other tasks performed at capture-time, including picture analysis, white-balance correction, Bayer interpolation and JPEG compression. Most implementations use some form of Discrete Cosine Transform fast-integer approximation or other techniques to speed up compression encoding \cite{Zeng01}. Lastly, quantisation tables are used to truncate encode detail to the desired level.

\section{Test images}
\begin{figure}\centering
\subfigure{\includegraphics[width=.23\columnwidth]{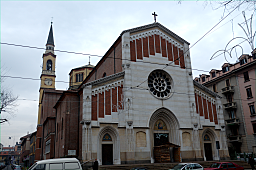}}\vspace{-1.7pt}
\subfigure{\includegraphics[width=.23\columnwidth]{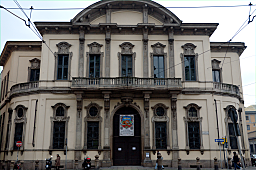}}\vspace{-1.7pt}
\subfigure{\includegraphics[width=.23\columnwidth]{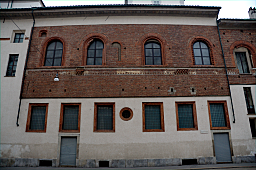}}\vspace{-1.7pt}
\subfigure{\includegraphics[width=.23\columnwidth]{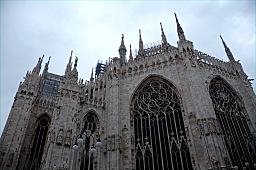}}\vspace{-1.7pt}
\subfigure{\includegraphics[width=.23\columnwidth]{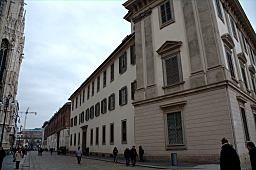}}\vspace{-1.7pt}
\subfigure{\includegraphics[width=.23\columnwidth]{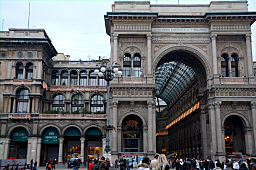}}\vspace{-1.7pt}
\subfigure{\includegraphics[width=.23\columnwidth]{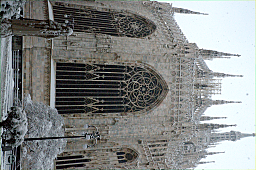}}\vspace{-1.7pt}
\subfigure{\includegraphics[width=.23\columnwidth]{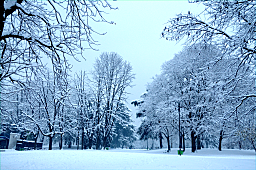}}\vspace{-1.7pt}
\subfigure{\includegraphics[width=.23\columnwidth]{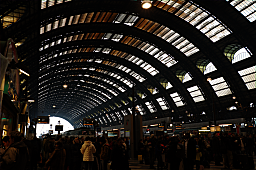}}\vspace{-1.7pt}
\subfigure{\includegraphics[width=.23\columnwidth]{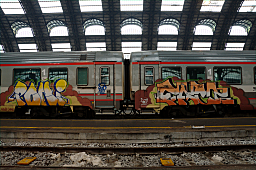}}\vspace{-1.7pt}
\subfigure{\includegraphics[width=.23\columnwidth]{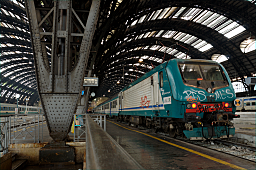}}\vspace{-1.7pt}
\subfigure{\includegraphics[width=.23\columnwidth]{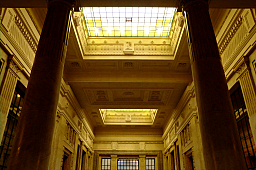}}\vspace{-1.7pt}
\subfigure{\includegraphics[width=.23\columnwidth]{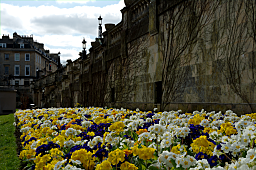}}\vspace{-1.7pt}
\subfigure{\includegraphics[width=.23\columnwidth]{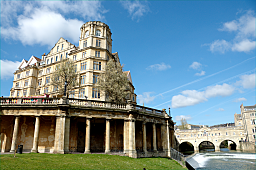}}\vspace{-1.7pt}
\subfigure{\includegraphics[width=.23\columnwidth]{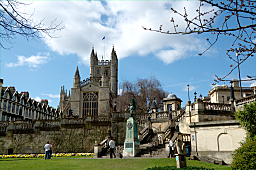}}\vspace{-1.7pt}
\subfigure{\includegraphics[width=.23\columnwidth]{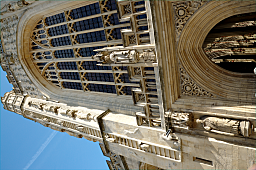}}\vspace{-1.7pt}
\subfigure{\includegraphics[width=.23\columnwidth]{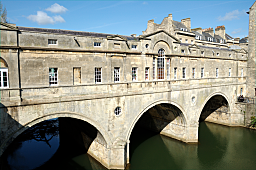}}\vspace{-1.7pt}
\subfigure{\includegraphics[width=.23\columnwidth]{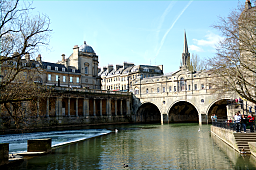}}\vspace{-1.7pt}
\subfigure{\includegraphics[width=.23\columnwidth]{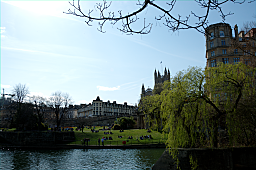}}\vspace{-1.7pt}
\subfigure{\includegraphics[width=.23\columnwidth]{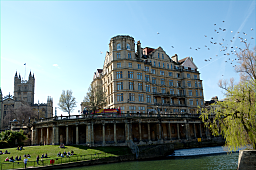}}\vspace{-1.7pt}
\subfigure{\includegraphics[width=.23\columnwidth]{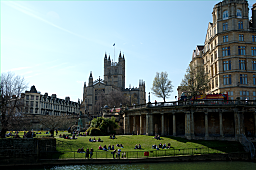}}\vspace{-1.7pt}
\subfigure{\includegraphics[width=.23\columnwidth]{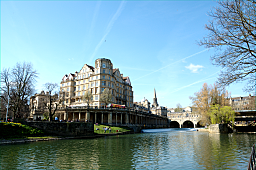}}\vspace{-1.7pt}
\subfigure{\includegraphics[width=.23\columnwidth]{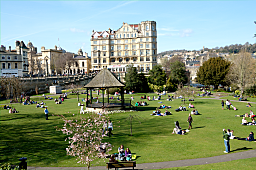}}\vspace{-1.7pt}
\subfigure{\includegraphics[width=.23\columnwidth]{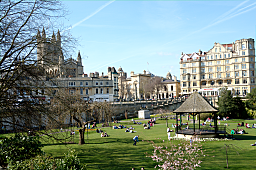}}\vspace{-1.7pt}
\subfigure{\includegraphics[width=.23\columnwidth]{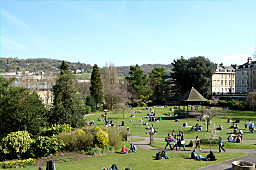}}\vspace{-1.7pt}
\subfigure{\includegraphics[width=.23\columnwidth]{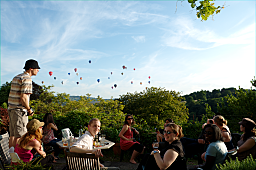}}\vspace{-1.7pt}
\caption{Real-world test images}
\label{fig:test-images}
\end{figure}

The input images used in this work are selected intentionally to present a number of typical shooting conditions. The input images are shown later in Figure~\ref{fig:test-images}. Images 1--8 are shot outside on an overcast low-brightness day, where a small aperture around f/8.0 would be used to minimise ACA\@. ISO sensitivity is higher at 400, and exposure time is longer at $5 \times 10^{-2}$ to $1 \times 10^{-2}$ seconds. Images 9--12 are typical light-constrained indoor shots with the aperture balanced around f/6.3; an increased ISO around 800 is used and much longer exposure time in the range $5\times 10^{-1}$ to $1$ second. Subsequent images 13--26 are shot in clear and bright conditions, allowing unrestricted use of shooting parameters, thus using an f/11.0 aperture for depth of field, higher corner detail and less ACA; ISO is set to 200 and exposure time between $1 \times 10^{-3}$ to $2 \times 10^{-2}$ seconds.

Additional to the real-world images, a CA-free picture is rendered using POVray with a high-complexity office scene \cite{Piqueres04}, shown later in Figure~\ref{fig:rendering}. Further, simple chequerboard patterns are rendered for basic validation, shown later in Figure~\ref{fig:pattern-comparison}.

From visual inspection across a number of the test images, the most visible LCA artifacts (i.e.~at high-contrast edges) are seen to be around 3--4 pixels in displacement. Since the red and blue planes were linearly interpolated from a Bayer array (due to simple demosaicing), there is around one pixel of uncertainty.

\section{Literature review}
Available research relating to LCA correction covers a wide spectrum of themes across varying application environments from scientific measurement to practical areas applicable to photography. Directly and indirectly relevant works are identified and grouped thematically into \textit{optical correction}, typically achieved through lens design enhancements, and \textit{computational correction} (typically achieved through software post-processing) areas. The contribution of the literature is then assessed relative to the goal of this research. For completeness, CA is briefly addressed in the wider field of non-camera optics relating to other imaging systems, including in the fields of microscopy, diffractive systems and contact lenses. Finally, the knowledge gap is identified and discussed.

\subsection{CA and non-lens systems}

Since CA affects waves of differing wavelengths propagating through media in general, the effects of this are found in many non-lens systems (though in principle, it affects \textit{all} such systems). Much of the research in these areas shows CA as having profound effects in three major fields outside the classical use of optics: ocular CA affecting the HVS, electron microscopy and diffractive systems.

Firstly, the study of LCA in the human eye is a buoyant field, driving advances in laser surgery and other treatments. Hay et al.\ shows the HVS to have intrinsic adaptation to CA \cite{Hay63}, though LCA alone is shown to contribute significantly in vision acuity loss in humans despite the fovea covering only an inner 3 degree arc \cite{Thibos87,Thibos92}. CA introduced from contact lenses or eyeglasses has been shown of higher significance, and a wide body of literature is available on this. Powell introduces a contact lens design that corrects ACA without increasing LCA \cite{Powell81}. Likewise, Zhang et al.\ finds the use of dual lens elements has proven to address LCA without the previous increases of ACA \cite{Zhang10}.

Secondly, correction for chromatic aberration in scanning electron microscopes involves third order and greater wavefront correction, based largely on the emergent work by Scherzer \cite{Scherzer47}. Later research by Tucker et al.\ used cubic phase plates for wave-front coding to increase the depth of field in microscopy while addressing ACA \cite{Tucker98}.

Lastly, the ability to use Fresnel and diffraction gratings with optics to compute Fourier transforms is limited in polychromatic light sources due to frequency-dependent variation; Ferriere and Goedgebuer developed a lens configuration which eliminated chromatic aberration by using an exactly symmetrical selection and placement of lens elements along the lens \cite{Ferriere82}.

\subsection{Lens-system optical correction}
A group of the literature addresses correction or compensation at some level of chromatic aberration by means of modifying the optical path. Since this occurs physically prior to image capture, it requires \textit{in-loop} design knowledge and modification typically via hardware or design changes. Due to the less flexible nature and the expense of implementing changes in this way, this area lacks the accessibility that alternative correction mechanisms offer.

Grey \cite{Grey63} conducted analysis on the inter-dependency of the various types of aberration within aberration theory, finding that adjustments to correct for one type of aberration affected other types, due to the non-orthogonal nature of optical aberrations. This work presented early conceptual methodology to computationally refine lens design using error minimisation.

The earliest literature to directly address CA introduced the minimisation of CA via active optical correction by Willson and Shafer \cite{Willson91}; this was achieved by using three independent optical paths for the three primary wavelengths through zoom lens motor micro-adjustment. This allowed an overall reduction in LCA for a specific focal length. Since three exposures are needed, temporal coherence is lost; this presents real-world impracticality with the motorised zoom lens constraint. Due to three discrete image samples, this method is thus unsuitable for scenes with movement.

In optimising lens design for lower CA (and other aberrations), Fang introduced the use of genetic algorithms \cite{Fang07}. This addressed the complexity associated with lens element selection from the wide variety of lens elements available, and from a range of sources. He demonstrated a measurable reduction in LCA and other optical aberrations.

Spatial Light Modulators (SLMs) have been developed in certain application fields where cost is not sensitive, e.g.\ microscopy. Millan et al showed application of this in diffractive optics and modifications to the light exit aperture to reduce LCA \cite{Millan06}.

More recent research has emerged showing ray-tracing through optical lenses in forward and backward directions using lasers and CCD sensors to obtain empirical chromatic aberration data \cite{Seong08}. This was shown to address the complexity and assumptions in software lens models though iteratively optimising the design of the lens system.

Indirectly related at the sensor capture level, CCD blooming presents significant issues and is often misinterpreted as CA due to the chromatic changes it causes. This is crucial to particular fields outside general photography, such as in astrophotography: the contrast of planets illuminated by the sun against free space presents very high contrast achievable in adjacent CCD cells, saturating the charge buckets and causing charge leakage to adjacent cells. At the CCD cell level, the introduction of additional circuitry to mitigate this has been shown to reduce the impact of this issue, at the expense of increased noise and complexity \cite{Ohba80}. Notably, CMOS systems are not fallible to this issue, so they remain a superior choice for high-contrast optical systems.

In a wider field outside correction of CA, \textit{active optics} have been in use for some time. One of the most significant and earliest fields was to use the sub-field of \textit{adaptive optics} in connection with astrophotography. This is employed to adapt the surface of a telescope mirror to approximately rectify the incoming wavefront as far as possible. To achieve this, Hartmann-Shack wavefront sensors and a tight control loop to update an array of actuators are used \cite{Shack71}. Related to this technology, Digital Micromirror Devices (DMDs) allow high-speed deflection (and therefore switching) of the optical light path \cite{Hornbeck87} within discrete pixels. DMDs have been widely deployed in consumer image projectors.

Although not directly relevant to CA reduction, Willson addressed the calculation of lens intrinsic parameters in motorised zoom lenses over a wide range of lens parameters, identifying additional problems \cite{Willson93}. Later, he went on to model the relationship between various extrinsic camera parameters for a motorised zoom-lens system \cite{Willson94}, based around Tsai's model \cite{Tsai87}. This work proved that interpolation is successful in obtaining values at intermediate lens parameters from a discrete set of parameter. Chen et al.\ went further to show bilinear interpolation on lens parameters holds for manual (i.e.\ non-motorised) lenses \cite{Chen00}. This allows measurements to be taken for a number of discreet lens parameters, interpolating between available parameters to correct for a tuple of parameters previously not seen.

\subsection{Computational correction}
While design changes, introduction or modification of hardware present cost and engineering constraints, the modification of image data \textit{post-capture} is attractive due to the accessibility and freedom it brings. Correcting images computationally is inherently in the \textit{post-processing} phase, as the image data is digitised after being affected by the analogue optics.

Relative to the optical correction group of works which can be considered comparatively mature, this area is rapidly developing. Some methods within software correction, including forward and inverse Fast Fourier Transforms and deconvolution, introduce significant error at the typical quantum depths of 32 or 64 bits per pixel. Future image processing is thus reliant on the introduction of GPU and processor architecture support for 128 and 256 bits per pixel quantum depths to minimise introduced noise, which will allow far more creative license, such as refocusing and near-total aberration correction via high-precision iterative deconvolution.

Overall, the majority of software techniques found present quantification, analysis and algorithms centred around fixed lens systems, i.e.\ for a tuple of given lens settings. This is traditionally sufficient for most applications, as for example web cameras commonly used in vision systems have a fixed focal plane and aperture, however with increasing imaging requirements and applications partially driven by continuously increasing specifications and costs, this area has steeper requirements and expectations.

Early work in the field suggested the use of image warping to correct LCA \cite{Green89,Rifman74}. Boult understood solving this issue numerically had advantages and introduced LCA correction via digital imaging with cubic splines as a function of radial distortion \cite{Boult92}; this resulted in a similar error reduction as with Willson's active optics. An achromatic chess board target was used to measure improvement.

Dersch started a project in 1998 to develop a toolkit to allow manipulation of images for the correction of distortion aberration \cite{Dersch03}; this relates to LCA correction through the use of warping for aberration correction. This was later extended for the purpose of automating image identification and registration for panoramic photography and used Lowe's SIFT algorithm \cite{Lowe04} to extract stable image features \cite{Brown03}. A similar and early approach in this work used the SIFT algorithm to identify image feature points on each plane for converging via non-linear registration, however instability and inaccuracy were found due to lack of information from reliance on discrete image features, which also limited radial coverage and resulting in under-correction at image extremities.

Later, Dersch's project was forked and openly developed. Watters identified it as being useful for the correction of LCA, presenting a manual technique for finding approximate LCA correction coefficients \cite{Watters04}. The steps consisted of: the red and blue channels were manually aligned to minimise visible LCA at an initially chosen radius, recording the pixel displacement and radius into a spreadsheet. Next, this was repeated for four further radius intervals. Lastly, the number of data points was selected, and the curve fitting function in Microsoft Excel was used to generate correction coefficients. This technique was subsequently refined by Krause to simplify the linear scaling of channels by the development of a software tool \cite{Krause04}, though the red and blue channel correction coefficients were still curve-fitted via spreadsheet software as before.

Notably, Park et al.\ used feature points for measurement and subsequent elimination of barrel distortion on zoomable lens systems \cite{Park00}. Benhimane developed a similar approach to correct for distortion aberration via corner detection against specific calibration patterns \cite{Benhimane04}.

Remondino concluded that LCA was the dominant form of CA and thus needed a robust correction mechanism \cite{Remondino06}, contextualising the basis for this work.

Kaufmann used EOS Systems' PhotoModeler software to obtain a simple translation and linear scaling for the red and blue channels relative to green \cite{Kaufmann05}, citing the translation was necessary to compensate for slight sensor and\slash or lens element misalignment. Difficulty was reported in detection of false-positive features when calibration was attempted in real-world scenes, concluding in the suggested use of bespoke achromatic calibration targets.

The work of Heuvel~and~Verwaaal showed that application of available camera barrel distortion correction software---in this case EOS Systems' PhotoModeler on a per-plane basis---was largely unsuccessful with their fixed fish-eye lens application \cite{Heuvel06}. Least squares minimisation was performed to reduce registration standard deviation of the 90 targets they measured down to 0.15 pixels. This left large registration error cited as up to 6.0 pixels, which was attributed to other issues, such as CCD blooming despite using linear plane scaling approximation. Further tests were conducted using manual linear scaling to reduce image artifacts observed, using Picture Window Pro 4.0, citing ``significant visual improvement'', but it still proved an incomplete solution \cite{Heuvel06}.

A self-calibration model proposed by Cronk used coloured circular features to calculate intrinsic parameters \cite{Cronk06}. Rudimentary blob-detection was used with five differently coloured features and multiple views. Quantification of the calibration results showed sub-pixel registration across much of the image, however accuracy was proven problematic citing a worst-case error of ``several pixels''.

Mallon et al.\ introduced fully automated correction of LCA through post-process warping the three colour planes from images of classic monochromatic chess-board calibration targets \cite{Mallon07}. Iterative minimisation was achieved through analysis of the colour histogram from the calibration target.

It has been found that variation among zoom lenses due to wear and hysteresis is ultimately significant for accurate correction mechanisms \cite{Cicala08}, challenging these earlier models.

Luhmann \cite{Luhmann06} used the method of self-calibrating \textit{bundle adjustment} on features from a calibration pattern, using interior and exterior parameter correlation; correspondence was introduced between channels correcting for LCA\@. This was achieved with per-channel feature measurement with opaque software \textit{Ax Ori Axios 3D}. Fryer and Brown \cite{Fryer86} extended this to use \textit{plumbline calibration} for calibration without exterior orientation. An improvement to feature-point accuracy of factor 1.6 was shown. Later, he demonstrated the effect of LCA via channel subtraction \cite{Luhmann06}, showing that it can be measured in a continuous way to produce an \textit{edge-map}.

Remondino~\cite{Remondino06} et al.\ proposed a method employing the use of colour filters to selectively calibrate individual colour planes with an achromatic calibration target and suggested the use of capturing the raw sensor data. They were able to show the significance of LCA with a maximum registration error of 10 pixels with a second-order radial distortion model based around earlier work \cite{Tsai87}. This was shown for four different focal lengths and varied with changes in focal length.

Taehee used a technique of edge refinement to minimise LCA \cite{Taehee07}; a filter was applied in both directions to adjust pixel colour when meeting various colour saturation conditions. The behaviour of this approach was not analysed in pictures where LCA may be falsely be measured, leading to the conclusion that it may not be robust. Moreover, this technique does not account for geometrical changes introduced by LCA, and adapts local regions.

The recent work of Cecchetto \cite{Cecchetto09} used inter-channel feature point correspondence, with statistical outlier filtering to correct LCA\@. As previously found, this technique highlights non-robust behaviour due to relatively lower-density of discrete feature points.

Pomaska superficially  showed that subtracting per-channel pixel values in Photoshop could be used to visualise changes in distortion (albeit for barrel distortion) \cite{Pomaska01}. This proves the mechanism of allowing a continuous approach, as opposed to using discrete image features for manual correction of barrel distortion. This allows the key technique of continuous correction to maximise information utilisation.

\section{Existing approaches}
The overriding advantage of the existing approaches grouped into the design and manufacture improvements is that the solution becomes intrinsic to the lens. Therefore, correction cannot be omitted later, and there is no potential for spatial frequency loss due to aliasing. Further, once the changes are incorporated into the lens design, there is no further time or production costs, and material costs from selection of more exotic glass materials with lower dispersion are known up-front.

In research grouped into image post-processing, the primary advantage is compelling: end-users or consumers of the optical hardware would be able to perform this or similar calibration and\slash or correction in order to minimise the effect of LCA and using arbitrary hardware. This has two issues though, firstly digital correction has a steep requirement of potentially a complex number of steps, and secondly that developed computational models and algorithms are needed. Optical or design techniques intrinsically avoid these.

\section{Summary}


Evaluating literature grouped into computational correction, a number of research works perform prior calibration against either the classic chess board or a target bespoke to the application. This introduces challenges in designing and fabricating a suitable target, error introduction from target mishandling (e.g.\ a printed target being imperfectly planar), and the need to perform calibration for each individual lens and parameter set.

The intersection of LCA correction and camera optics shows a narrow field, mainly understood due to LCA being an increasing concern with the availability of high-resolution sensors and reasonable fidelity optics. The application of a number techniques in other imaging areas has been tested against the application of LCA correction and highlights some areas of non-robust behaviour. This is particularly the case where post-capture techniques for LCA correction are manual and based on visual judgement; this demonstrates a wide knowledge gap with automatic or closed-loop correction.

Consequently there is a clear opportunity to develop techniques tailored for the micro and interrelated adjustments needed for stable LCA detection and correction to a sub-pixel level. A model that takes advantage of the full continuity of all the pixels on the image is proposed, rather than using a comparatively small subset of the information available from discrete feature points. 



The next chapter will deal with recovering longitudinal chromatic aberration distortion information from a collection of images, via a completely automated workflow. Using appropriate image decoding and analysis, accurate chromatic aberration correction can then be applied, exceeding the resulting quality of most hardware and solutions and with no requirement for user-expertise and allowing totally automatic correction.

\cleardoublepage
\chapter{Proposed method}
\label{cha:methodology}
\section{Introduction}
In this work, an algorithm that eliminates LCA is proposed. This is achieved by compensating for the prism-like deflection of light through the lens elements used in capturing a photograph. Since the camera's sensor filters light down to three discrete wavelengths rather than a continuum of wavelengths, it is proposed that the two sets of image data (planes) reconstructed from red and blue elements of the Bayer CFA, will undergo non-linear radial warping to map them onto the third (green) plane.

As the camera's sensor measures the luminance of light at these wavelengths across the same plane, separation of the sensor cells with the same wavelengths into different planes gives three incomplete arrays; typically cells are spatially divided into groups of four as shown earlier in Figure~\ref{fig:bayer}: two with a green cell-filter and the other two with red and blue filters. This provides the most cost and complexity-efficient implementation due to the eye's peak sensitivity near the 555nm (green appearance) wavelength \cite{Fairchild05}, thereby giving increased spatial resolution in the green plane. The CIE 1924 colour appearance model \cite{CIE26} uses this basis. To derive an image plane with the quarter or half per-colour pixel occupancy from the sensor, linear interpolation is used and results in a continuous plane. Since the green plane has twice as many samples per unit area than red or blue, the red and blue planes are selected for later distortion to register against the green plane; this minimises error induced from pixel aliasing which would reduce the maximum spatial frequency of the resulting image.

\section{Approach}
The proposed algorithm remaps pixels according to a radial distortion model from the source plane; this is done to align luminance features and detail with that of the pixels in the green plane, removing LCA\@. This is feasible through exploiting inter-planar luminance correspondence, for example due to objects commonly contributing luminance information over multiple RGB planes; even green leaves on a tree have some luminance detail in the red and green planes, but moreover their background will be visible and give strong edge detail. Geometric differences are thus observable \cite{Luhmann06}, allowing an error function to be derived and subsequently minimised to give inter-plane congruency.

From Snell's law, differing wavelengths will refract at different angles through a medium, thus the resulting geometrical error is radial from the optical centre. As shown earlier, modelling this radial distortion using a Taylor series as shown in Equation~\ref{eqn:seidel-approx} and up to the fourth order is suitable. The resulting polynomial allows a remapping function to be generated, used to compute radial pixel displacement from the optical centre; this is used to reverse the chromatic distortion due to LCA.

\begin{figure}\centering
\subfigure{\includegraphics[width=0.2\textwidth]{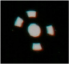}}
\subfigure{\includegraphics[width=0.2\textwidth]{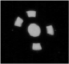}}
\subfigure{\includegraphics[width=0.2\textwidth]{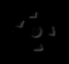}}
\caption{Plane subtraction first observed by Luhmann}
\label{fig:luhmann}
\end{figure}

\begin{figure}\centering
\subfigure{\includegraphics[width=\textwidth]{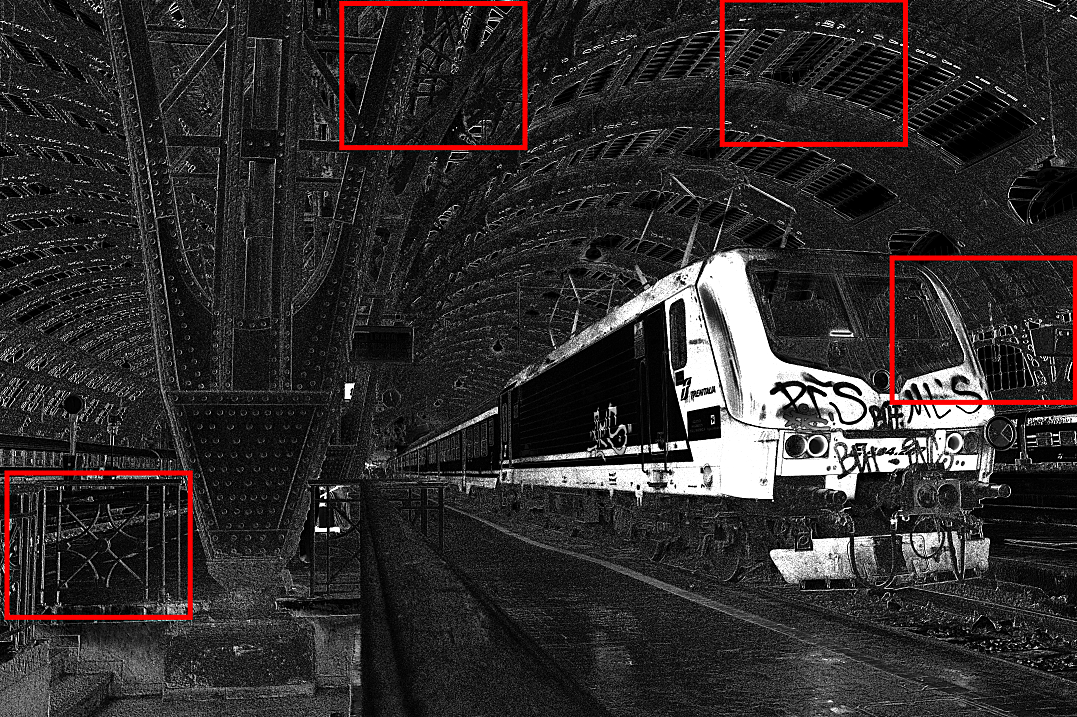}}\vspace{-7pt}
\subfigure{\includegraphics[width=0.4960\textwidth]{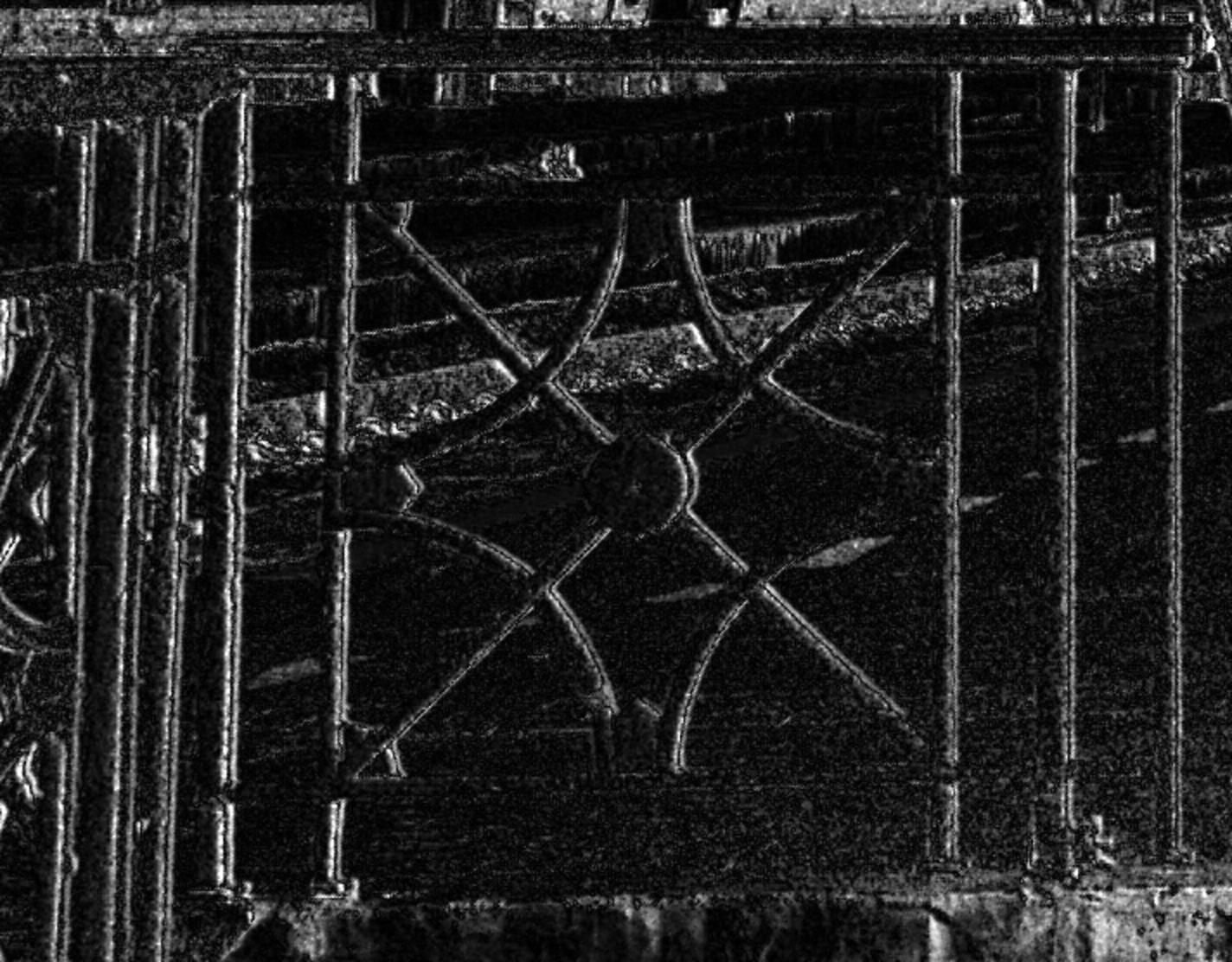}}
\vspace{-7pt}
\subfigure{\includegraphics[width=0.4960\textwidth]{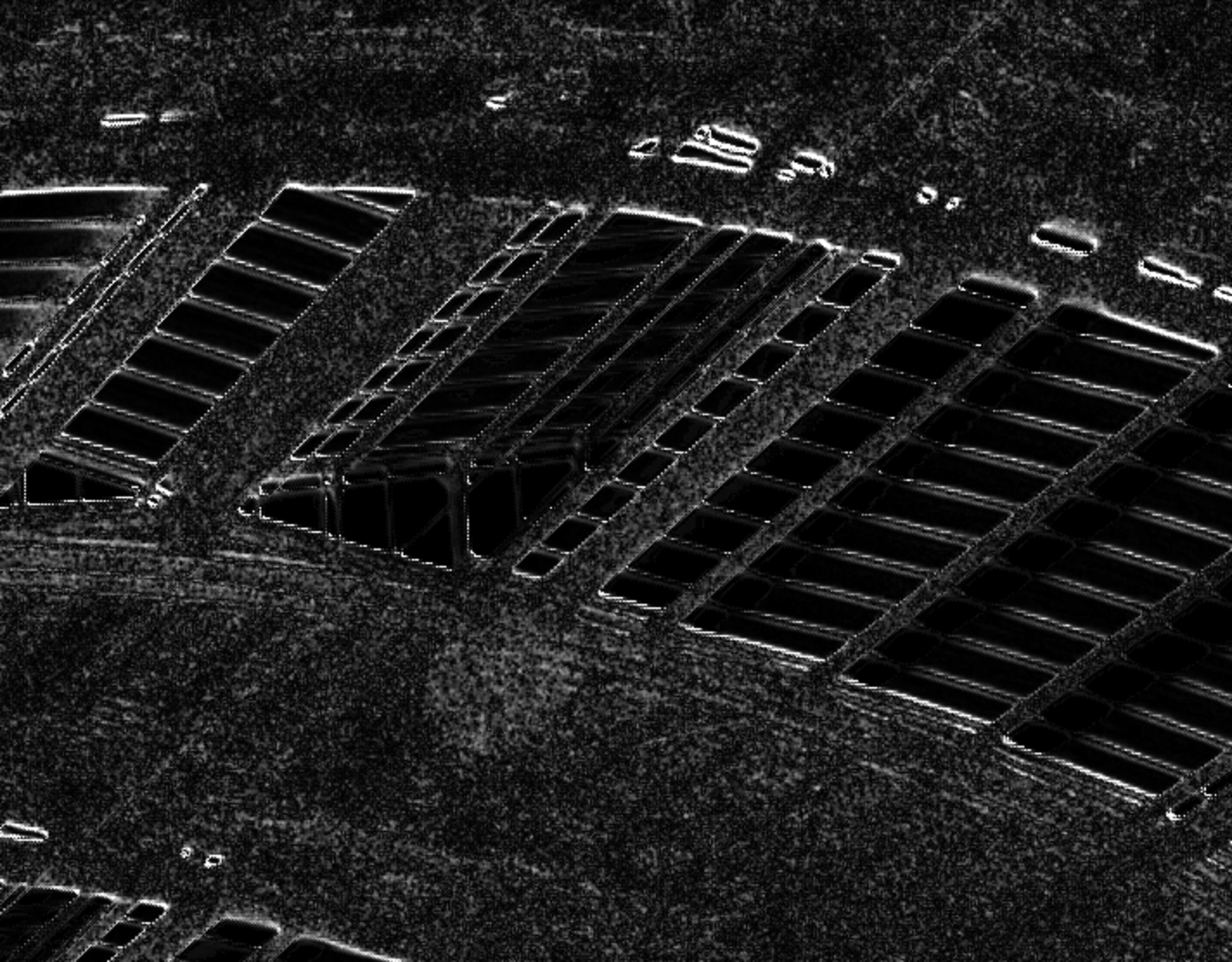}}
\subfigure{\includegraphics[width=0.4960\textwidth]{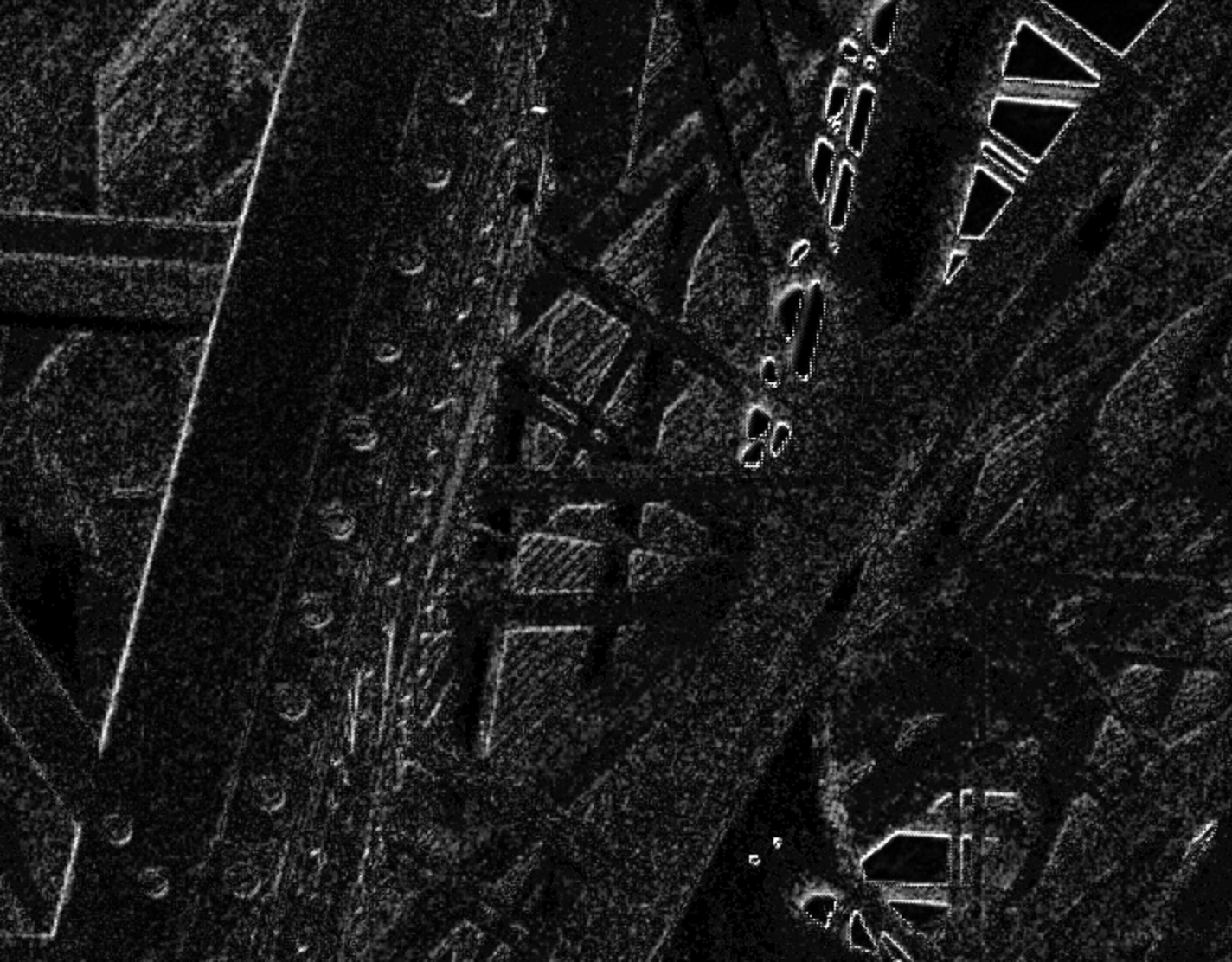}}
\vspace{-7pt}
\subfigure{\includegraphics[width=0.4960\textwidth]{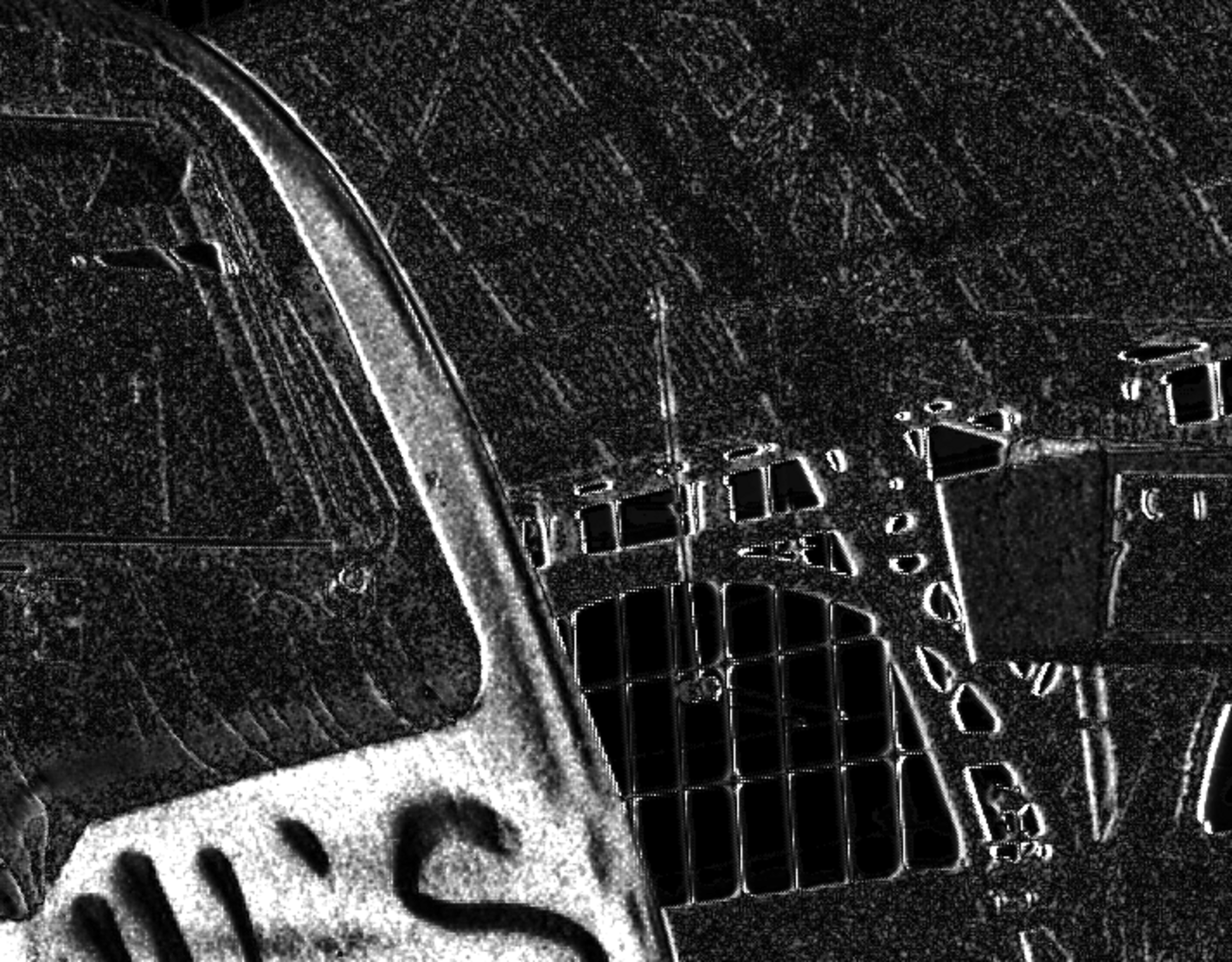}}
\caption[Cropped sections from uncorrected difference map]{Cropped sections from red-green uncorrected absolute difference map}
\label{fig:difference-uncorrected}
\end{figure}

Since the internal lens geometry and precise positioning of lens elements and aperture are abstracted somewhat by imprecise sensor measurement and thus unknown, the function of how light will distort cannot be known or computed \textit{a priori}; manufacturers typically ray-trace through theoretical lens models, and equations used do not apply to end-use of the system \cite{Laikin91}. Lack of insight and complete specification of internal lens geometry and construction prevents direct mathematical or computational methods of finding the radial distortion model coefficients providing least error. Instead, an error function is defined and minimised iteratively. A novel technique is employed for the error function: for each R-G and B-G pair, iterate over all the pixels in the image planes, summing the absolute difference between each plane; a similar technique was first observed by Luhmann, though using subtraction \cite{Luhmann06}, shown in Figure~\ref{fig:luhmann}. As LCA introduces spatial incongruities between an object's pixels in different planes, the absolute difference is numerically lower when there is complete alignment. Typical visible incongruities are seen in images as colour fringes around high-contrast features, such as dark tree branches against the light sky, and particularly towards the image extremities where the aberration error is typically maximal, and zero at the optical centre from Seidel's aberration theory, seen in Equation~\ref{eqn:seidel-equation}. An appropriate non-linear error minimisation function is employed to adjust the Taylor series coefficients; the distortion is applied against the image plane, and the error function executed and further iterations performed until sufficient accuracy is met. This technique is only valid in the untransformed RGB colourspace; conversion to other colourspaces samples values from multiple channels at coincident points, defeating the separation needed to preserve the discrete chromatic information. The application of this method is shown in Figure~\ref{fig:difference-uncorrected}.

Since the optical path through the lens system varies with focussing distance, focal length and aperture size, the corresponding tuple of correction coefficients as applied to the polynomial, cannot be applied to pictures with non-identical lens parameters. For practical use, a database of the lens parameters is constructed allowing applicable correction coefficients to be found for any given set of lens parameters; this is chosen since correction parameters for a given tuple of lens parameters can be held as state for rapid later use. It is hypothesised that a significant number of pictures across numerous lens settings need to be optimised and correction parameters added to the lens database, to give sufficient data for later \textit{high-fidelity} offline correction on unseen images.

Since the computational cost of recovering the LCA correction coefficients is high, and a single remapping correction pass is comparatively low-cost, the overall procedure is performed in two stages: \textbf{coefficient recovery} and \textbf{image correction}, shown previously in Figure~\ref{fig:architecture}. This allows efficient re-use of the correction coefficients.

\subsection{Error function}
The function which is minimised to undo the distortion that LCA has introduced is core to the algorithm. It operates across all pixels in the source plane buffer and distorted plane buffer, returning a scalar value; lower values indicate less LCA incongruity. The function is:

\begin{eqnarray}
e(P, M) = \frac{\sum_{P \in M} \mid P_g - P_w \mid}{n}
\label{eqn:converge}
\end{eqnarray}

where $P$ is a pixel in the set of pixels in mask $M$, of which $P_g$ is pixels value in the green plane and $P_w$ is the value in the working plane (i.e. the red or blue plane) and $n$ is cardinality of set $P$.

\subsection{Decentering calibration}
Lens decentering relative to the image centre is useful for the benefit of completeness according to Brown's method \cite{Brown66} and is not used as part of the algorithm presented; it is simply integrated by changing the distortion centre offsets, thus it is intrinsic to the methodology.

The offsets due to optical decentering are relatively small with respect to the magnitude of LCA aberrations seen with production SLR lens hardware, itself by definition non-existent at the optical centre. Any decentering offset outside the central image zone would have a diminishing effect as the distance from the optical centre increases \cite{Karras98}, therefore the impact optical decentering has on image quality is negligible and is not considered further. Existing works using manual linear adjustment omit this, though it could be considered in a future step.

\section{Coefficient recovery}
\begin{algorithm}
\SetKwFunction{LoadImage}{LoadImage}
\SetKwFunction{PlanarSplitInterpolate}{PlanarSplitInterpolate}
\SetKwFunction{EqualiseHistogram}{EqualiseHistogram}
\SetKwFunction{RadialDistort}{RadialDistort}
\SetKwFunction{AbsoluteDifference}{AbsoluteDifference}
\SetKwFunction{MaskedAverage}{MaskedAverage}
\SetKwFunction{BoundedBFGS}{BoundedBFGS}
\SetKwFunction{StoppingCriteria}{StoppingCriteria}
\SetKwFunction{WriteDatabase}{WriteDatabase}
\SetKwFunction{WriteImage}{WriteImage}
\SetKwFunction{IntrinsicParams}{IntrinsicParams}
\SetKwFunction{Blank}{Blank}

\SetKwData{Database}{Database}
\SetKwData{Image}{Image}
\SetKwData{Metadata}{Metadata}
\SetKwData{Intrinsics}{Intrinsics}
\SetKwData{Planes}{Planes}
\SetKwData{RefPlane}{RefPlane}
\SetKwData{WorkingPlanes}{WorkingPlanes}
\SetKwData{plane}{plane}
\SetKwData{distorted}{distorted}
\SetKwData{error}{error}
\SetKwData{bounds}{bounds}
\SetKwData{coeffs}{coeffs}
\SetKwData{mask}{mask}
\SetKwData{diff}{diff}
\SetKwData{Immediate}{Immediate}
\SetKwData{Lens}{Lens}

\SetKwInOut{Input}{input}
\SetKwInOut{Output}{output}

\Input{Image to be corrected complete with metadata}
\Input{Database of lens intrinsic parameters}
\Output{Database entry for correction coefficients at given lens parameters}
\Output{Optionally, corrected image with updated metadata}

$\Image\leftarrow$ \LoadImage{}\;
$\Metadata\leftarrow$ \Image.\Metadata\;
$\Intrinsics\leftarrow \Database[\Metadata.\Lens].\Intrinsics$\;
$\Planes\leftarrow$ \PlanarSplitInterpolate{\Image}\;
$\RefPlane\leftarrow \Planes[G]$\;
$\WorkingPlanes\leftarrow \Planes[R, B]$\;
\BlankLine
\ForEach{$\plane \in \Planes$} {
	\EqualiseHistogram{\plane}\;
}
\BlankLine
\ForEach{$\plane \in \WorkingPlanes$} {
	$\coeffs\leftarrow$ ($0.99$, $0.01$, $0.0$, $0.0$)\;
	$\bounds\leftarrow$ ($1.0\pm0.05$, $0.0\pm0.05$, $0.0\pm0.05$, $0.0\pm0.05$)\;

	\Repeat{\StoppingCriteria{\error}} {
		$\distorted\leftarrow$ \RadialDistort{\plane, \coeffs, \Intrinsics}\;
		$\mask\leftarrow$ \RadialDistort{\Blank{}, \coeffs, \Intrinsics}\;
		$\diff\leftarrow$ \AbsoluteDifference{\distorted, \RefPlane}\;
		$\error\leftarrow$ \MaskedAverage{\diff, \mask}\;
		$\coeffs\leftarrow$ \BoundedBFGS{\coeffs, \error, \bounds}\;
	}
}
\BlankLine
\WriteDatabase{\coeffs}\;

\If{\Immediate}{
	\WriteImage{\distorted, \Metadata}\;
}
\label{alg:flow-coefficient-recovery}
\caption{Correction coefficients recovery algorithm}
\end{algorithm}

\begin{figure}\centering
\linespread{0.8}
\begin{tikzpicture}[
	node distance = 3.6em,auto,
	decision/.style={diamond, draw=blue, very thick, text width=4em, text centered, inner sep=1pt, draw=green!50!black!50,top color=white,bottom color=green!50!black!20},
	cloud/.style={rectangle,minimum size=1em,very thick,draw=red!50!black!50,top color=white,bottom color=red!50!black!20,font=\itshape},
	line/.style={draw, thick, -latex',shorten >=2pt},
	block/.style={rectangle,minimum size=1em,rounded corners=3mm,very thick,draw=blue!30!black!50,top color=white,bottom color=blue!30!black!20, text width=8em, text centered},
	point/.style={coordinate},>=stealth',thick,draw=black!50,
	tip/.style={->,shorten >=1pt},every join/.style={rounded corners},
]
	\node[cloud] (X) {start};
	\node[block, left of=X, node distance=8em] (A) {acquire source image}; 
	\node[block, below of=A] (B) {decode image metadata};
	\node[block, below of=B] (C) {load intrinsic lens parameters};
	\node[block, below of=C] (F) {histogram equalisation};
	\node[block, below of=F] (G) {distort working plane};
	\node[block, above right of=G, node distance=16em] (H) {compute mask};
	\node[block, below of=H] (I) {compute absolute difference};
	\node[block, below of=I] (J) {compute average pixel value};
	\node[block, below of=G] (L) {refine distortion coefficients};
	\node[decision, below of=J, node distance=7em] (K) {evaluate stopping criteria};
	\node[block, above right of=K,node distance=16em] (M) {save data}; 
	\node[cloud, below of=M] (Y) {finish};

	\path[tip] (X) edge (A);
	\path[tip] (A) edge (B);
	\path[tip] (B) edge (C);
	\path[tip] (C) edge (F);
	\path[tip] (F) edge (G);
	\path[tip] (G) edge [out=0, in=180] (H);
	\path[tip] (H) edge (I);
	\path[tip] (I) edge (J);
	\path[tip] (J) edge (K);
	\path[tip] (K) edge [out=180,in=0] node [near end] {descent} (L);
	\path[tip] (L) edge (G);
	\path[tip] (K) edge [out=0,in=180,below] node [near start,auto,swap] {converged} (M);
	\path[tip] (M) edge (Y);
\end{tikzpicture}
\caption{Coefficient recovery steps}\label{fig:flow-coefficient-recovery}\end{figure}
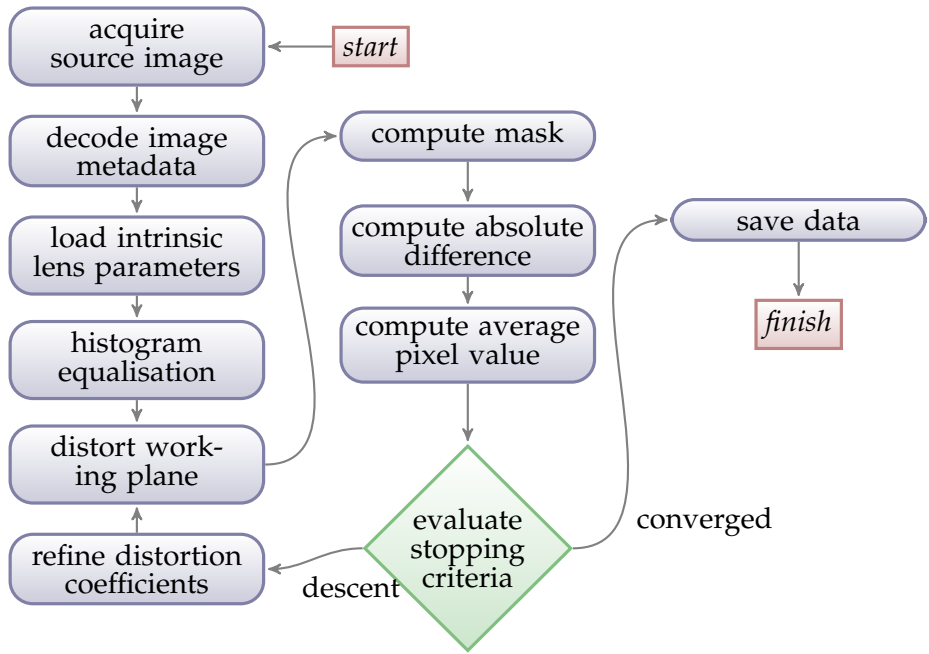

The coefficient recovery algorithm estimates the correction coefficients which give the minimum LCA\@, thus either artificial, e.g.\ calibration target or real world images are suitable input data for this algorithm, given in Algorithm~\ref{alg:flow-coefficient-recovery}. This is broken down into discrete functional steps, shown in Figure~\ref{fig:flow-coefficient-recovery}.

\textit{Acquire source image}: An image acquired in raw format is selected and the data loaded into a buffer. Lossy-compressed images (e.g. JFIF's JPEG \cite{JPEG92}) are typically not used, due to two main reasons: primarily, non-raw formats introduce colour aliasing during CFA demosaicing, due to adaptive algorithms which adjust the local pixel colour due to directional gradients, or \textit{misguidance} artifacts, such as in AHD \cite{Hirakawa05} --- this prevents correct micro-adjustment with the error minimisation step later. Secondly, compression artifacts and chrominance subsampling will suppress high-frequency spectral detail and attenuate low-amplitude edges, needed for optimal convergence. All digital SLRs and most bridge and some high-end point and shoot cameras can capture images in raw format, thus this is not restrictive for the scope of this work (line 1).

Simple demosaicing is performed during this step to extract pixel value information for each group of cells, performed using DCRAW\footnote{DCRAW is ubiquitous for raw conversion integrated in a wide variety of open and closed-source projects} \cite{Coffin06} to three separate R, G and B plane buffers as \textit{brightness} values, thus not scaled as luminance values according to the photopic equation \cite{CIE26} (line 4).

Lastly in this step, linear interpolation is performed. Since the sampling points at the three primary wavelengths are not co-incident but adjacent, linear interpolation must be used to give continuous colour planes with the same spatial size and position. CFA sensors have twice as many green-filtered photosensitive cells than red or blue, thus the interpolated green plane can have twice the spatial frequency than the red or blue planes. Modern algorithms used to avoid colour `zipper' effects at colour boundaries, such as Hirakawa's popular Adaptive Homogeneity-Directed demosaicing algorithm \cite{Hirakawa05} must be avoided, as these use colour selectivity to give homogeneous edge features by detecting and minimising differences across colour planes (line 4).

\textit{Decode image metadata}: Particular parameters specific to the lens settings at the time the image are captured and encoded in EXIF tags. These are extracted, and include the lens descriptor, focal length, aperture value and image sequence number and are needed for associating the correction coefficients with lens parameters for later use (line 2).

\textit{Load intrinsic lens parameters}: Optionally if decentering has been computed or is known, the camera lens decentering values, as part of the precomputed lens intrinsic parameters are loaded from the database using the lens descriptor (line 3).

\textit{Histogram equalisation}: Camera systems seek to use the correct exposure to achieve a wide global tonal range at image capture time; this is done in a way to preserve the relationship between channels. Since the algorithm uses inter-channel contrast, \textit{per-channel} histogram equalisation is performed to ensure convergence is stable on pictures with heavy colour casts, such as shot with a non-white primary light source, e.g.\ under a sodium street lamp (line 8).

\textit{Distort working plane}: Distortion of the working plane to the distortion buffer is performed, using the current set of distortion coefficients. Bilinear interpolation is used to minimise aliasing from non-integer pixel displacements (e.g.\ that nearest neighbour would cause), and is selected as a tradeoff of \textit{compute time} and \textit{aliasing error} (line 14).

\textit{Compute mask}: A mask is computed to select only the subset of pixels common to the reference plane and distorted plane; this is achieved by testing each pixel in the distorted plane and setting the pixel in the mask if it is interior to the distortion, otherwise clearing it. This later ensures that pixels outside the resulting distorted frame do not influence the computed pixel value average, ensuring the convergence will be the true minimum error (line 15).

\textit{Compute absolute difference}: For each pixel, the absolute difference between the reference plane and distorted plane is calculated and stored (line 16).

\textit{Compute average pixel value}: The non-masked absolute difference pixels are summed and divided by the total non-masked pixel count in the distortion buffer, giving the average per-pixel difference. This serves as a size-independent measure of how \textit{homogeneous} the planes are, required for convergence (line 17).

\textit{Evaluate stopping criteria}: The stopping criteria is evaluated to determine if the rate of change of error is sufficiently low from the minimum observed, or a reasonable limit on iteration count is reached. This is implemented as part of the error minimisation algorithm selected, in the next step (line 19).

\textit{Refine distortion coefficients}: The distortion coefficients are modified based on feedback from the error function, via the L-BFGS-B method \cite{Byrd95}. L-BFGS-B seeks to minimise the value of the error function by adjusting the distortion coefficients which are passed to the error function, which performs the per-plane distortion and averaging of the inter-plane absolute difference to compute the resulting error. L-BFGS-B was selected for faster and more robust convergence than the Modified Powell method \cite{Brent73} and Nelder and Mead's downhill simplex method \cite{Nelder65}, due to the function being minimised having a low rate of change over small changes in area \cite{Melchior07}. Other gradient descent algorithms are unsuitable due to inability to compute the function gradient or Hessian matrix, which is inherently not possible from using an abstract error function (line 18).
         
\textit{Save data}: After convergence, selected image metadata, extracted from EXIF fields in the image headers is stored along with the converged correction coefficients for later use (line 21).

The coefficients given by minimisation are used to distort the original image and written to a file, thus without the intermediate steps of \textit{histogram equalisation}. This part of the step is optional, and omitted when just generating correction data for the database (line 23).

\section{Image correction}
\begin{figure}\centering
\linespread{0.8}
\begin{tikzpicture}[
	node distance = 4.5em,auto,
	decision/.style={diamond, draw=blue, very thick, text width=4em, text centered, inner sep=1pt, draw=green!50!black!50,top color=white,bottom color=green!50!black!20},
	cloud/.style={rectangle,minimum size=1em,very thick,draw=red!50!black!50,top color=white,bottom color=red!50!black!20,font=\itshape},
	line/.style={draw, thick, -latex',shorten >=2pt},
	block/.style={rectangle,minimum size=1em,rounded corners=3mm,very thick,draw=blue!30!black!50,top color=white,bottom color=blue!30!black!20, text width=8em, text centered},
	point/.style={coordinate},>=stealth',thick,draw=black!50,
	tip/.style={->,shorten >=1pt},every join/.style={rounded corners},
]

\node[cloud] (X) {start};
\node[block, left of=X, node distance=8em] (A) {decode lens parameters};
\node[block, below of=A] (C) {search for nearest neighbour} edge [loop right] node {for all entries} (C);
\node[block, below of=C] (E) {decode image content to RGB planes};
\node[block, below of=E] (F) {distort plane} edge [loop right] node {for R,B planes} (F);
\node[block, below of=F] (G) {encode image data and tags};
\node[cloud, below of=G] (Y) {finish};

\path[tip] (X) edge (A);
 \path[tip] (A) edge (C);
\path[tip] (C) edge (E);
\path[tip] (E) edge (F);
\path[tip] (F) edge (G);
\path[tip] (G) edge (Y);
\end{tikzpicture}
\caption{Image correction with model database}\label{fig:flow-correction}\end{figure}
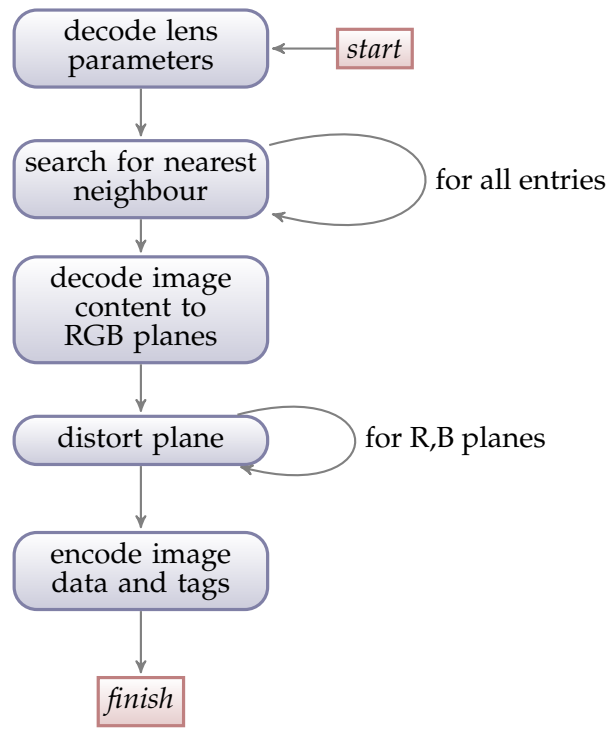

The workflow of leveraging the correction database to correct pictures is given in Figure~\ref{fig:flow-correction} and broken down into discrete steps:

\textit{Decode lens parameters}: Parameters specific to the image are decoded from the image's EXIF tags. This includes the lens descriptor, focal length and aperture value. The database records matching the lens descriptor are selected.

\textit{Search for nearest neighbour}: For each record, a distance vector is computed based on the difference between focal length, focus distance and discrete aperture setting. This is part of the Nearest Neighbour Search \cite{Clark54} in the next step, and can use cost vectors to optimise selection.

After iterating through all database records, the nearest neighbour is known, and the correction coefficients are selected from this record.

\textit{Decode image content to RGB planes}: The luminance values at each CFA cell are separated into image planes by the colour filter pattern. Simple bilinear interpolation is used to generate continuous colour planes with the same spatial size and position. This step is achieved using DCRAW \cite{Coffin06}.

\textit{Distort plane}: For each of the red and blue planes, the plane is resampled into a new planar buffer using the selected correction coefficients. Accurate resampling is achieved via a windowed approximation of the classic sinc function \cite{Shannon49}, such as the Lanczos function \cite{Lanczos50}; a higher node count in the windowed section can be used for more accuracy. Since the green plane acts as a reference for other planes, and has a significantly higher spatial frequency limit than the other planes, it does not undergo any transformation, thus minimising loss due to aliasing and rounding to the colour depth.

\textit{Encode image data and tags}: The planar image data is assembled into an RGB triplet buffer and encoded into the preferred intermediate and lossless image format. A suffix is added to the filename to associate the correct image with the original and the image data is written. Additionally, the EXIF tags are written and updated with information that this image has undergone LCA correction applied, thus should not be reapplied.

\section{Novel approach to correction quantification}
Where an image is presented with a known (computed, predefined or otherwise) distortion, the most direct measure of success after application of the correction algorithm, is some quantified difference or \emph{distance vector} between the known distortion and the converged correction. A sensible metric for this is the area of difference of the two polynomials representing the actual distortion and the correction from the selected distortion coefficients. In real-world situations, a more useful metric is the \emph{perception} of chromatic aberration, since the \textit{ground-truth} correction coefficients are not known. Though perception cannot be accurately quantified, the changes to image spatial frequency can be accurately quantified. Since the HVS is sensitive to high spatial frequency in image luminosity, image \textit{clarity} or \textit{acuity} is strongly linked to changes in spatial frequency \cite{Bovik05}. If chromatic distortion is introduced, per-plane spatial frequency will remain the same, though the spatial frequency of the image's luminosity will increase, introducing false detail and reducing desired detail contrast.

Methods similar to this have not been encountered in this field, so a method to quantify this change has been developed, optimised to this application. Firstly, the image is transformed from a triple-plane RGB image into single-plane luminance image via standard weightings from the CIE 1924 day photopic equation, from Equation~\ref{eqn:photopic} \cite{CIE26}. The luminance plane is transformed into the frequency domain through applying a Discrete Fourier Transform (DFT), giving an array of complex numbers in the frequency domain:

\begin{eqnarray}
F_1(u,v) &=& \frac{1}{MN} \sum_{x=0}^{M-1} \sum_{y=0}^{N-1} f(x,y) e^{-j2\pi (\frac{ux}{M} + \frac{vy}{N})}
\end{eqnarray}

where $f(x,y)$ is the source image in $x$ and $y$ spatial domain with size $M$ and $N$ elements respectively. $F_1(u,y)$ is the destination image in $u$ and $v$ frequency domain. A second DFT is produced from the image being compared against.

In order to generate a useful spectrogram, the algorithm proceeds to iterate through the $u$ and $v$ dimensions in frequency space; it computes the spatial frequency as the Euclidean distance from the 0Hz point and adds the magnitude of the complex value at that point to the bin covering that spatial frequency, producing a spectrogram $S$. Represented as a function of spatial frequency $f$:

\begin{eqnarray}
S(f) &=& \sum_{\theta = 0}^{\pi} \mid F_2(f \cos \theta, f \sin \theta) \mid - \mid F_1(f \cos \theta, f \sin \theta) \mid
\end{eqnarray}

where $S(f)$ is the spectrogram of spatial frequency $f$, and $F$ is the DFT image.

Lastly, a windowed \textit{Hann} filter is employed over the resulting one-dimensional vector to attenuate undesired high-frequency components for the sake of visualisation \cite{Hamming98} and the result is plotted against radius, where the centre is 0Hz, i.e.~the whole-image average and up to the Nyquist frequency, thus one pixel spatial resolution \cite{Cochran67}. The resulting graph plots luminance spatial frequency from the given input image. Where the input image is the subtraction of two DFTs, this allows inspection of frequency changes introduced by LCA occurring in the lens system, along with magnitude information. This method for quantifying the loss in image \textit{acuity} due to LCA is represented in Algorithm~\ref{alg:quantification}.


\begin{algorithm}
\SetKwFunction{LoadImage}{LoadImage}
\SetKwFunction{Luminosity}{Luminosity}
\SetKwFunction{Pad}{Pad}
\SetKwFunction{DFT}{DFT}
\SetKwFunction{sqrt}{sqrt}
\SetKwFunction{centreDFT}{centreDFT}
\SetKwFunction{polarToCartesian}{polarToCartesian}
\SetKwFunction{filterHanning}{filterHanning}
\SetKwFunction{sumColumn}{sumColumn}
\SetKwFunction{Log}{log}
\SetKwFunction{graph}{graph}
\SetKwFunction{Max}{max}

\SetKwData{Ia}{Ia}
\SetKwData{La}{La}
\SetKwData{Ib}{Ib}
\SetKwData{Lb}{Lb}
\SetKwData{Image}{image}
\SetKwData{complexMap}{complexMap}
\SetKwData{outputMap}{outputMap}
\SetKwData{histogramMap}{histogramMap}
\SetKwData{histogram}{histogram}
\SetKwData{filtered}{filtered}
\SetKwData{magnitude}{magnitude}
\SetKwData{Elem}{elem}
\SetKwData{diff}{diff}
\SetKwData{Width}{width}
\SetKwData{Height}{height}

\SetKwInOut{Input}{input}
\SetKwInOut{Output}{output}

\Input{Images to be analysed}
\Output{Graph showing change in spatial frequency}

\Ia$\leftarrow$ \LoadImage\;
\Ib$\leftarrow$ \LoadImage\;
\La$\leftarrow$ \Luminosity{\Ia}\;
\Lb$\leftarrow$ \Luminosity{\Ib}\;
\ForEach{\Image $\in \La, \Lb$} {
	\Image$\leftarrow$ \Pad{\Image\, \Max(\Image.\Width, \Image.\Height)}\;
	\complexMap$\leftarrow$ \DFT{\Image}\;
	\ForEach{$x, y \in$ \complexMap} {
		\Elem$\leftarrow$ \complexMap$[x, y]$\;
		\magnitude$\leftarrow$ \sqrt{${\Elem_i}^2 + {\Elem_j}^2$}\;
		$\outputMap[\Image, x, y]\leftarrow$ \Log{$\frac{1 + \magnitude}{\Image.\Height}$}\;
	}

	\centreDFT{\outputMap$[\Image]$}\;
}
\diff$\leftarrow \outputMap[A] - \outputMap[B]$\;
\histogramMap$\leftarrow$ \polarToCartesian{\diff}\;
\ForEach{$x \in$ \histogramMap} {
	$\histogram[x]\leftarrow$ \sumColumn{\histogramMap, $x$}\;
}
\filtered$\leftarrow$ \filterHanning{\histogram}\;
\graph{\filtered}\;
\label{alg:quantification}
\caption{Quantification algorithm}
\end{algorithm}

\section{Minimiser selection}
\begin{figure}\centering
\subfigure[red to green displacement]{\includegraphics[width=0.8\textwidth,trim=.3cm 1.15cm 1.6cm 0cm,clip]{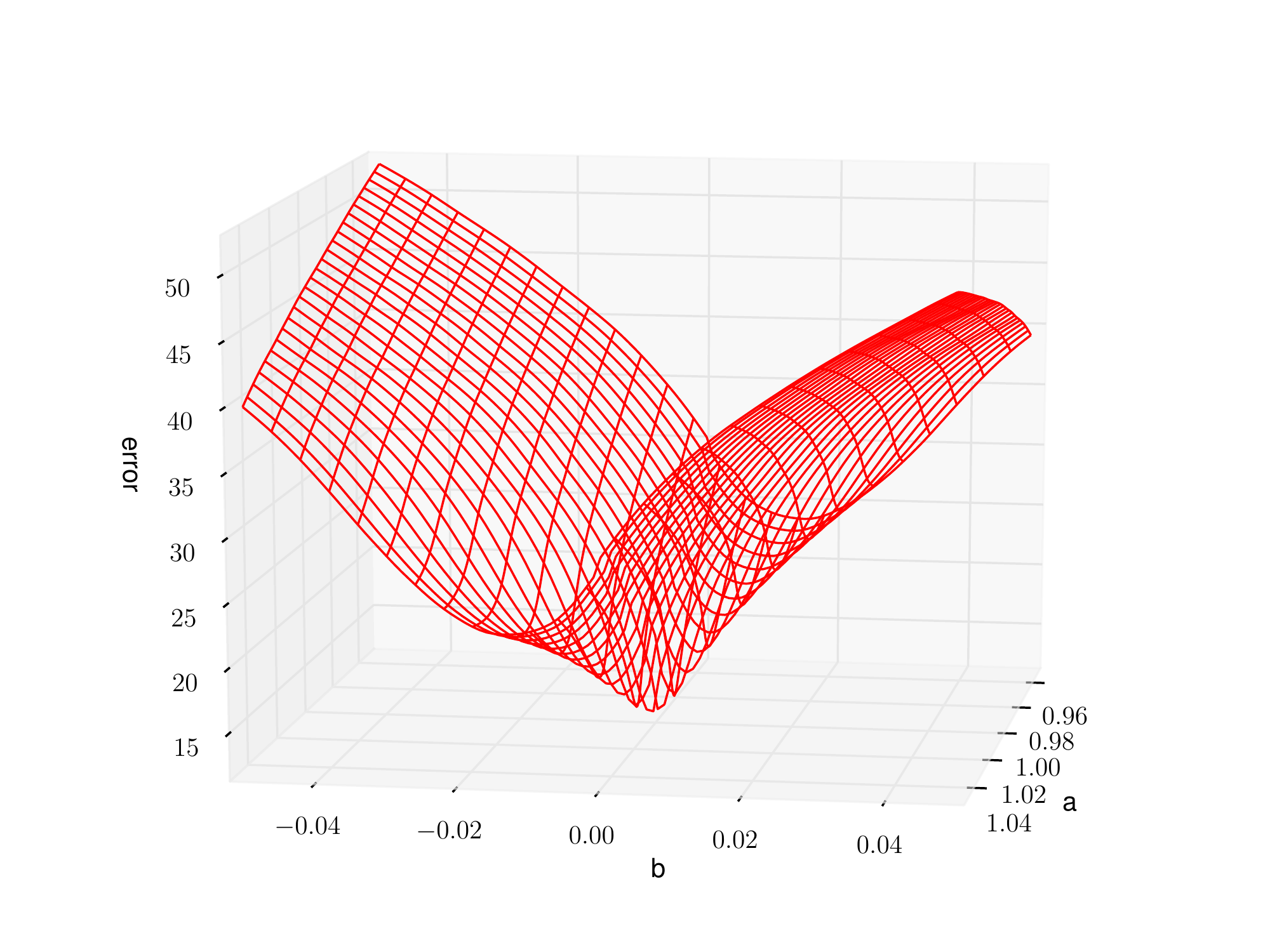}}
\subfigure[blue to green displacement]{\includegraphics[width=0.8\textwidth,trim=.3cm 1.15cm 1.6cm 0cm,clip]{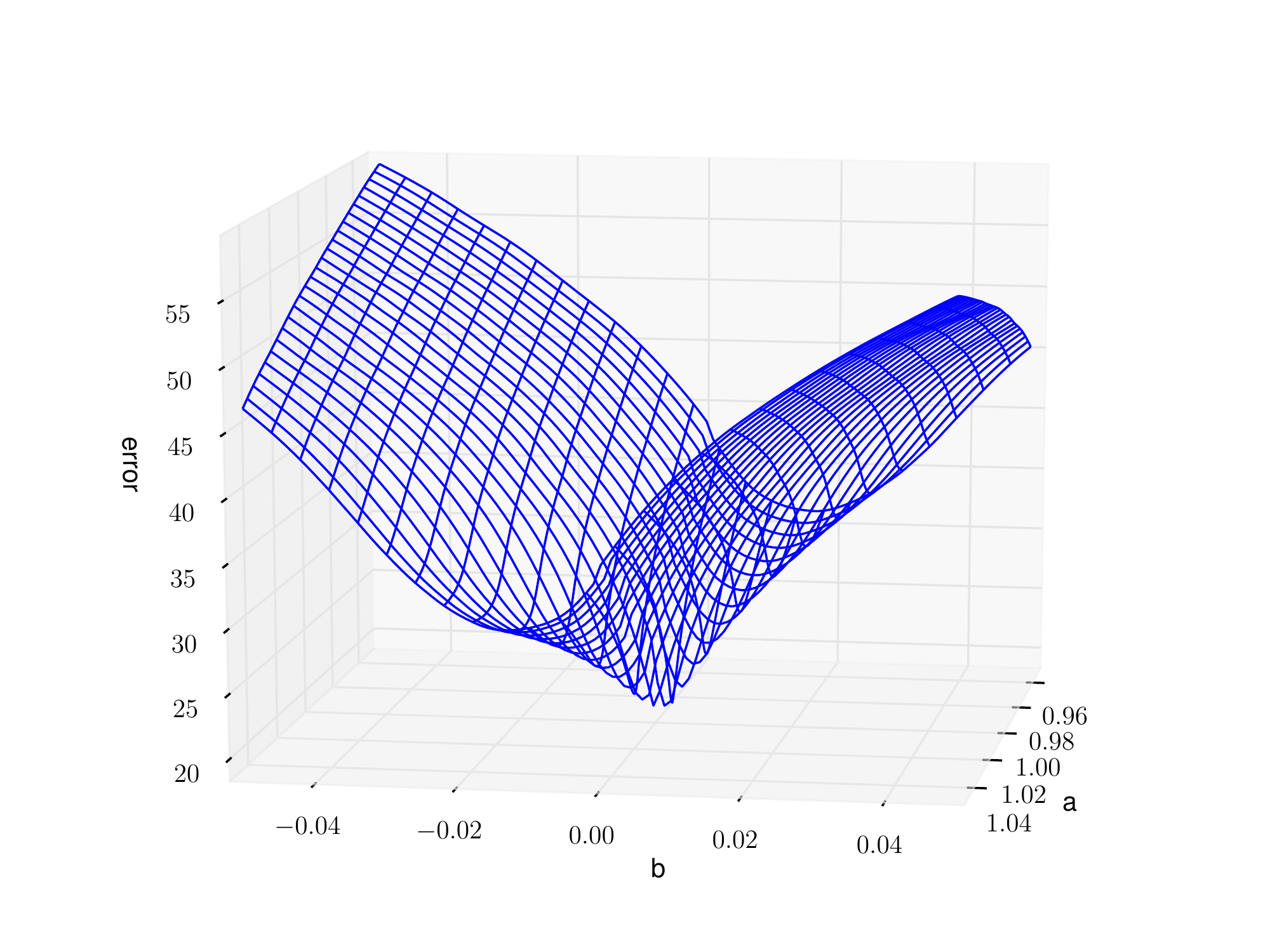}}
\caption{Error varying the first and second correction coefficients}
\label{fig:energy}
\end{figure}

The function measuring the difference between image planes for a given set of distortion coefficients, shown in Equation~\ref{eqn:converge} is minimised by varying the coefficients used for distorting the image and measuring the error. Since the function operates by taking the absolute difference of the image planes, it is not possible to compute the function derivative or Hessian matrix, limiting the choice of minimisation algorithms. Understanding the surface geometry of this function gives insight into the complexity of the problem space and if convergence is expected to be robust. To show the surface, brute-force sweeping is conducted in the first and second order coefficients for the Red-Green and Blue-Green plane pairs, shown in Figure~\ref{fig:energy}. It can be seen that there is a clear global minimum, thus robustness is expected within the limits.

A number of optimisation methods were evaluated for convergence performance and stability. The simplex downhill algorithm \cite{Nelder65} was found to not stabilise on the global minima, but diverge on particular images. This was found to be in part from the gradient descent also being designed for minimising linear problems, whereas Figure~\ref{fig:energy} shows the non-linearity. Powell's \cite{Powell64} alternative to the simplex algorithm showed fewer iterations were needed for convergence, however it exhibited stability issues with the noisier blue channel with three pictures from the test set. The Broyden-Fletcher-Goldfarb-Shanno (BFGS) algorithm \cite{Shanno70,Goldfarb70,Fletcher70,Broyden70} was found to address these stability issues due to taking smaller steps, limiting overshoot and non-linearity error. A final refinement was to use the Limited-memory BFGS-Bounded (L-BFGS-B) constrained optimiser \cite{Byrd95}.

Whilst careful selection of constraint values is important for consistent stability, it was found that limiting the polynomial coefficients to a level that would give in excess of maximum possible error (i.e.~more than 30 pixels) still maintained the desired stability; thus $1\pm0.05$ was chosen for the zeroth order coefficient and $\vert0.05\vert$ was chosen for higher order coefficients; even the most basic of lens designs would have polynomial coefficients within these bounds.

	

\section{Summary}
The methodology was broken into two phases, \textit{coefficient recovery} and \textit{image correction}. Offline image analysis is performed in the coefficient recovery stage; lens setting parameters are extracted and correction coefficients are found and stored in the database. Image correction takes sets of image parameters and selects the nearest match for the correction of unseen images without incurring the coefficient recovery overheads, and runs the correction algorithm.

Alternative approaches may involve discrete feature points, yielding far less information and thus correction accuracy or stability, or may examine gray areas. This approach performs no segmentation, using all areas, since non-gray areas typically have colour components in two of the three red, green and blue planes.

\cleardoublepage
\chapter{Methodological evaluation}
\label{cha:evaluation}
\section{Validation}
This chapter introduces a number of steps taken to validate the proposed algorithm at levels of increasing complexity. 

\subsection{Chequerboard test image}
Initially, a classic chequerboard image is rendered; this is distorted via radial pixel remapping from Equation~\ref{eqn:seidel-approx}. Coefficients presenting a significant level of LCA distortion are selected:

\begin{eqnarray}
r_{dest} &=& a \times {r_{src}} + b \times {r_{src}}^2 + c \times {r_{src}}^3 + d \times {r_{src}}^4 \nonumber \\
a &=& 0.98 \nonumber \\
b &=& 0.01 \nonumber \\
c &=& -0.01 \nonumber \\
d &=& 0 \nonumber \\
\therefore \ \ r_{dest} &=& 0.98 r_{src} + 0.01 {r_{src}}^2 -0.01 {r_{src}}^3
\end{eqnarray}

The correction algorithm is executed on the resulting distorted image; it is expected that it will converge on the minima at which there is least LCA distortion, i.e.~back to LCA-free. This can only be tested with no uncertainty because the pattern is \textit{achromatic} and by definition, LCA-free to start with.

\begin{figure}\centering
\subfigure{\includegraphics[width=\textwidth]{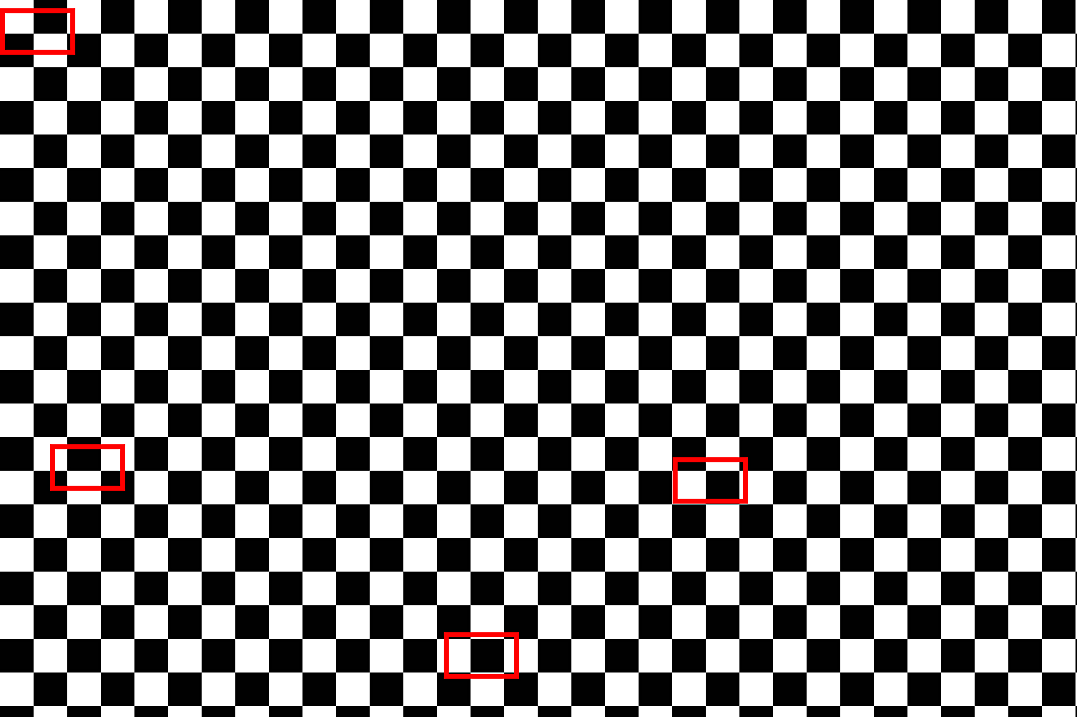}}\vspace{-7pt}
\subfigure{\fbox{\includegraphics[width=0.3113\textwidth]{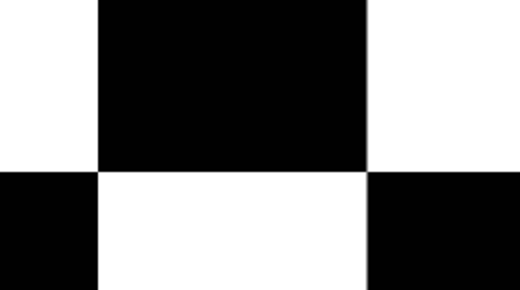}}}
\subfigure{\fbox{\includegraphics[width=0.3113\textwidth]{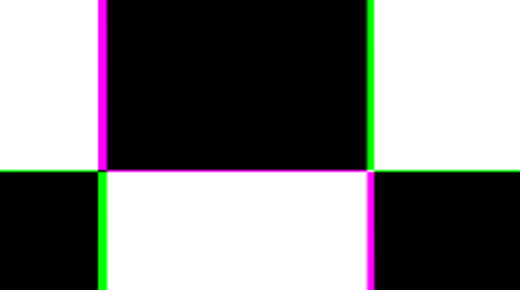}}}
\vspace{-7pt}
\subfigure{\fbox{\includegraphics[width=0.3113\textwidth]{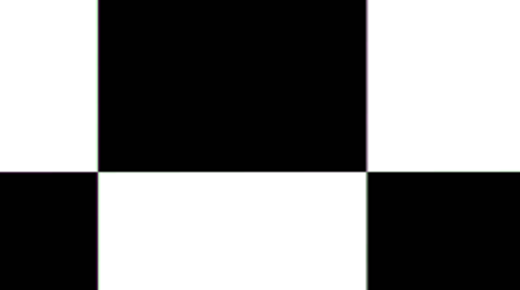}}}
\subfigure{\fbox{\includegraphics[width=0.3113\textwidth]{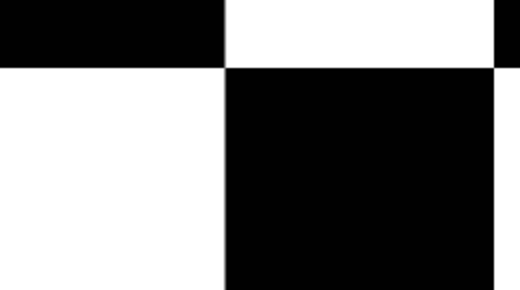}}}
\subfigure{\fbox{\includegraphics[width=0.3113\textwidth]{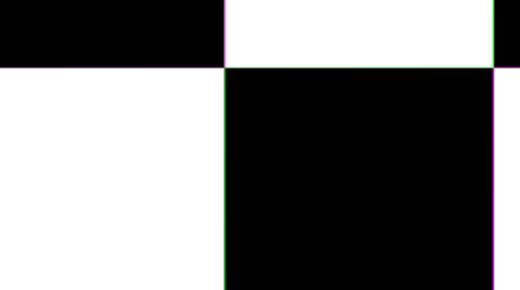}}}
\vspace{-7pt}
\subfigure{\fbox{\includegraphics[width=0.3113\textwidth]{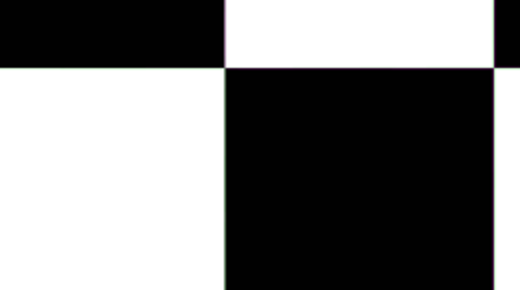}}}
\subfigure{\fbox{\includegraphics[width=0.3113\textwidth]{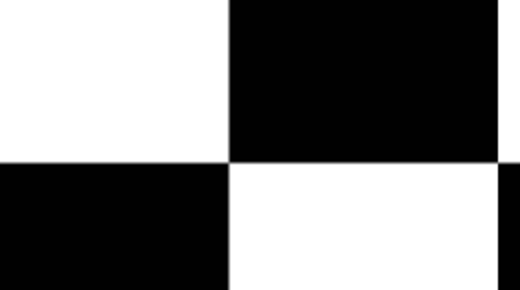}}}
\subfigure{\fbox{\includegraphics[width=0.3113\textwidth]{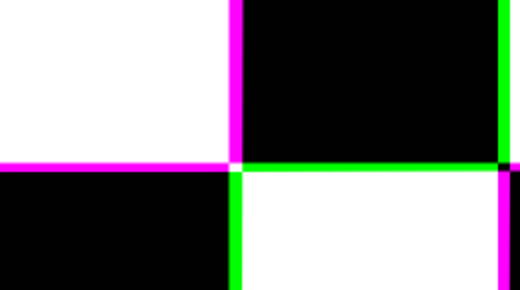}}}
\vspace{-7pt}
\subfigure{\fbox{\includegraphics[width=0.3113\textwidth]{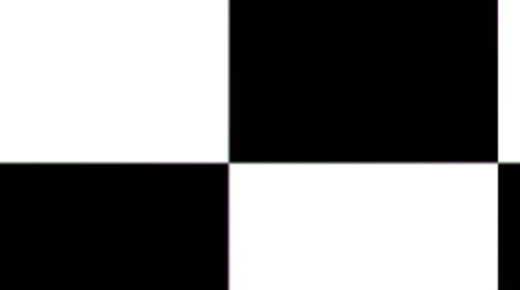}}}
\subfigure{\fbox{\includegraphics[width=0.3113\textwidth]{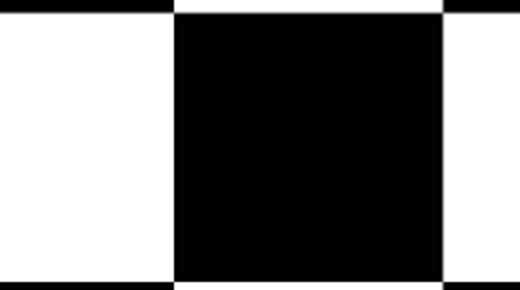}}}
\subfigure{\fbox{\includegraphics[width=0.3113\textwidth]{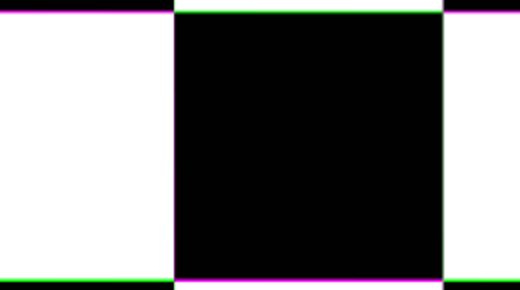}}}
\vspace{-7pt}
\subfigure{\fbox{\includegraphics[width=0.3113\textwidth]{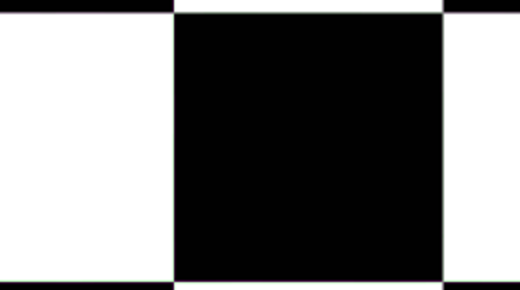}}}
\caption[Cropped sections from test image]{Cropped sections from original (left) distorted (center) and corrected (right) test image}
\label{fig:pattern-comparison}
\end{figure}

Crops of the test image are shown in Figure~\ref{fig:pattern-comparison}. By inspection of the righthand column of crops, no residual LCA artifacts, which would appear as linear colour fringes, are visible down to sub-pixel level. This shows three results; firstly the algorithm has converged on the global minima and not at any local minima. Secondly, convergence is accurate due to correction at the sub-pixel. Lastly, convergence is reliable across various radii as no LCA is detectable in any of the resulting crops.

Visual inspection is not reliable, due to a number of factors: firstly, subjectivity and perception may label measured error insignificant. Secondly for the error to be observed, it needs to be perceptually and psychovisually significant, i.e.\ colour offset within a noisy texture generates a low HVS response compared to high-contrast edges, thus is harder to visually flag. Lastly, quantifying the magnitude of the change is important to understand and \textit{measure} the performance of the algorithm. To address these concerns, Algorithm~\ref{alg:quantification} is employed, measuring the change in spatial frequency from the original aberration-free pattern relative to the distorted pattern. The quantification algorithm produces a graph with the magnitude of the change in spatial frequency plotted against spatial frequency.

\begin{figure}\centering\includegraphics[width=0.7\textwidth,trim=.3cm 0cm 1.6cm 1.0cm,clip]{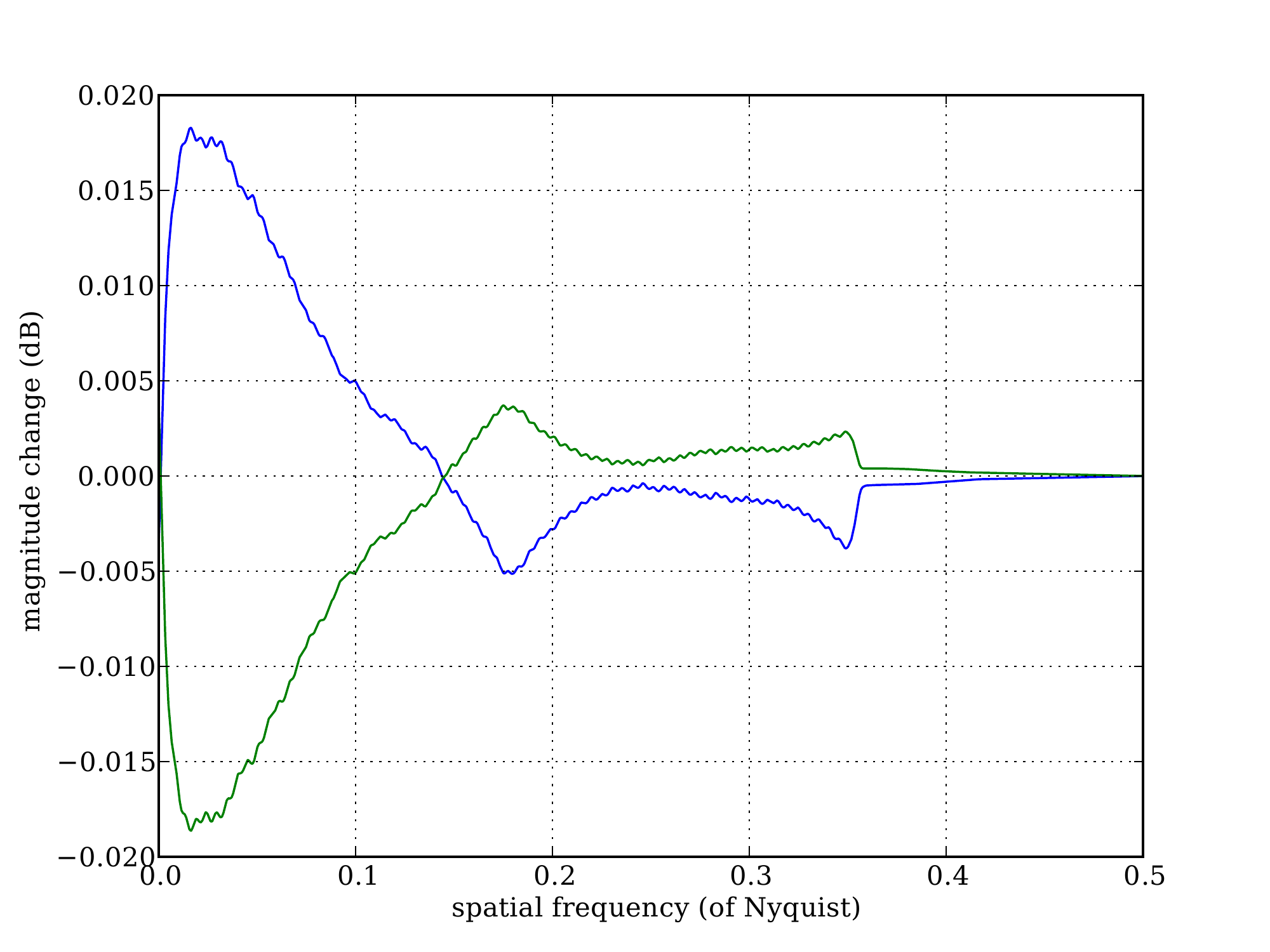}\caption[Change in spatial frequency in pattern images]{Change in spatial frequency from undistorted to distorted (blue) and distorted to corrected (green) pattern images}\label{gph:pattern-comparison}\end{figure}

The process of artificially introducing \textit{effectively}-LCA distortion, seen as the green line in Figure~\ref{gph:pattern-comparison}, increases low spatial frequency up to around 0.15~Nyquist, and decreases the higher frequency magnitude significantly up to around 0.36~Nyquist. Since the chequerboard image has regular intervals and maximal contrast at transitions, there is a very strong response at harmonic frequencies of the block size, due to the periodic nature of DFTs. In essentially damping the luminance contrast and geometrically spreading the contrast transitions, attenuation measured in the DFT at higher frequencies would be expected and is observed. The distortion recovery algorithm, shown as the blue line, effectively reverses the spatial frequency changes, including restoring the higher frequencies that were attenuated in order to converge on minimal distortion. From the graph, at $f = 0.18 N$ where $N$ is the Nyquist frequency of the image, we see an around 20\% reduction in frequency restoration. This is attributed to aliasing loss, since the pattern observed is symmetrical with respect to the introduced distortion.

\subsection{Ray-traced scene}
\begin{figure}\centering
\subfigure{\includegraphics[width=\textwidth]{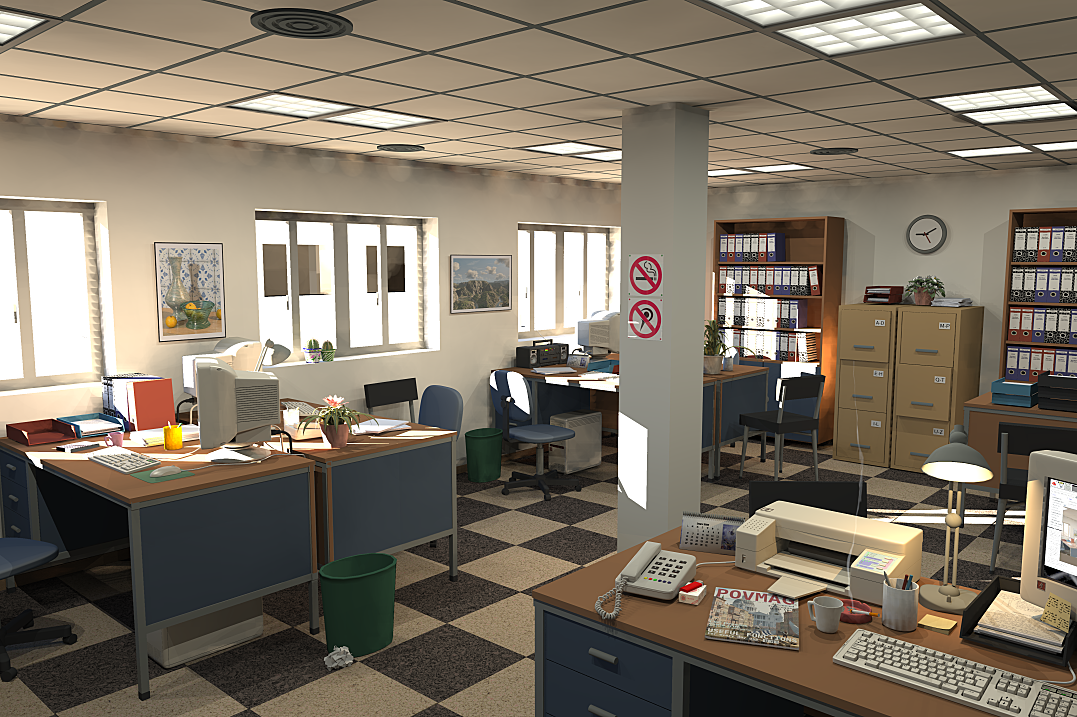}}\vspace{-7pt}
\caption{Rendered test image}
\label{fig:rendering}
\end{figure}

As a second validation step, a ray-traced scene representative of real-world luminance, spatial detail and colour complexity is selected and rendered, shown in Figure~\ref{fig:rendering}. This provides a chromatic aberration-free source image, which is consequently distorted in the same manner with known coefficients to test the performance of the algorithm with typical spatial frequencies present. Since the image is not acquired from a Bayer array, it has higher information density and approaches the Nyquist limit. As a result, more detail loss due to interpolator quality, aliasing, and approximation or accuracy limits in the image warping is expected.

\begin{figure}\centering
\subfigure{\includegraphics[width=0.3253\textwidth]{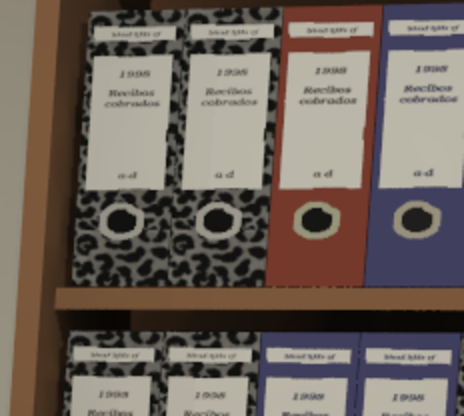}}
\subfigure{\includegraphics[width=0.3253\textwidth]{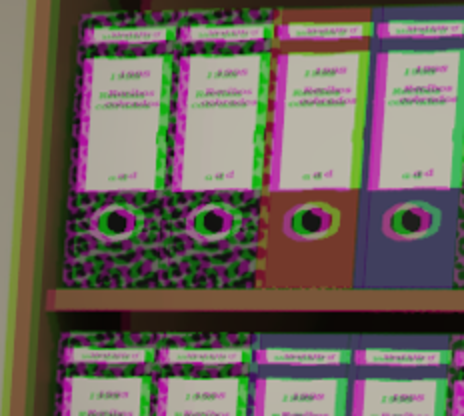}}
\vspace{-7pt}
\subfigure{\includegraphics[width=0.3253\textwidth]{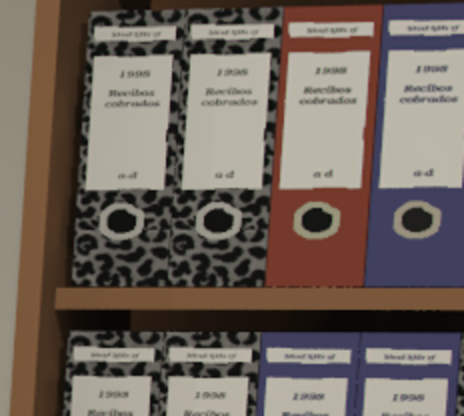}}
\subfigure{\includegraphics[width=0.3253\textwidth]{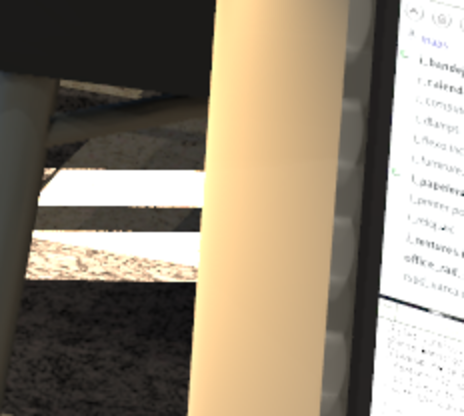}}
\subfigure{\includegraphics[width=0.3253\textwidth]{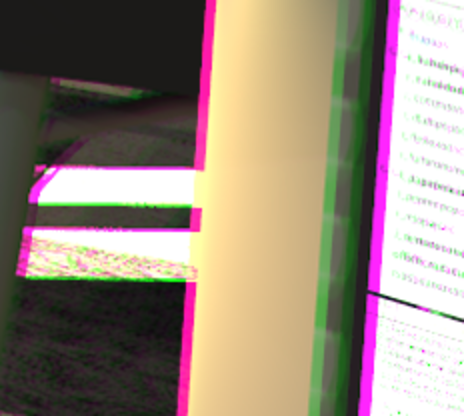}}
\vspace{-7pt}
\subfigure{\includegraphics[width=0.3253\textwidth]{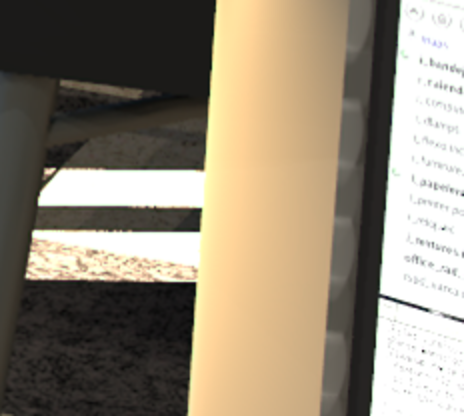}}
\subfigure{\includegraphics[width=0.3253\textwidth]{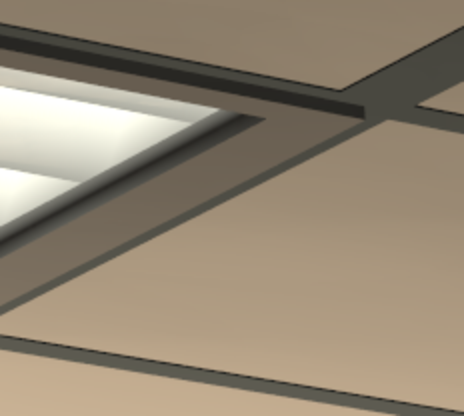}}
\subfigure{\includegraphics[width=0.3253\textwidth]{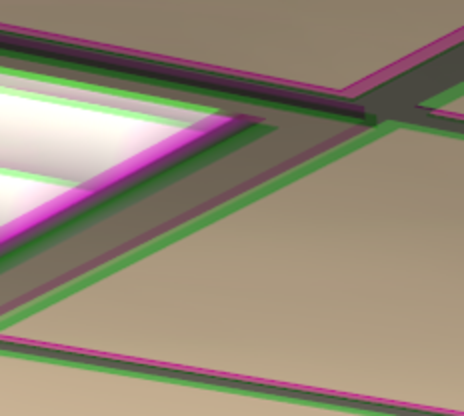}}
\vspace{-7pt}
\subfigure{\includegraphics[width=0.3253\textwidth]{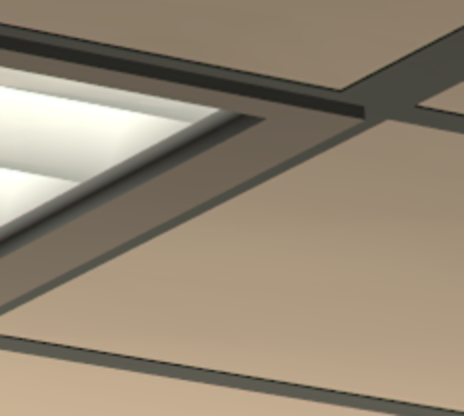}}
\subfigure{\includegraphics[width=0.3253\textwidth]{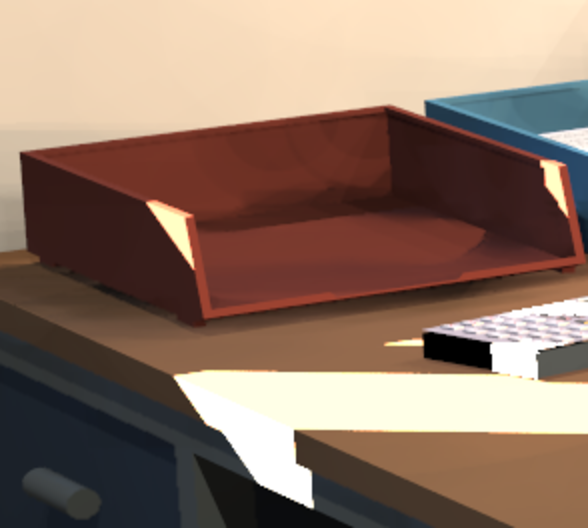}}
\subfigure{\includegraphics[width=0.3253\textwidth]{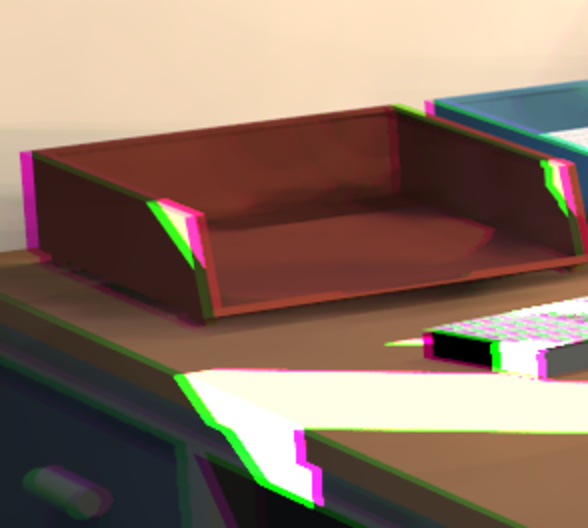}}
\vspace{-7pt}
\subfigure{\includegraphics[width=0.3253\textwidth]{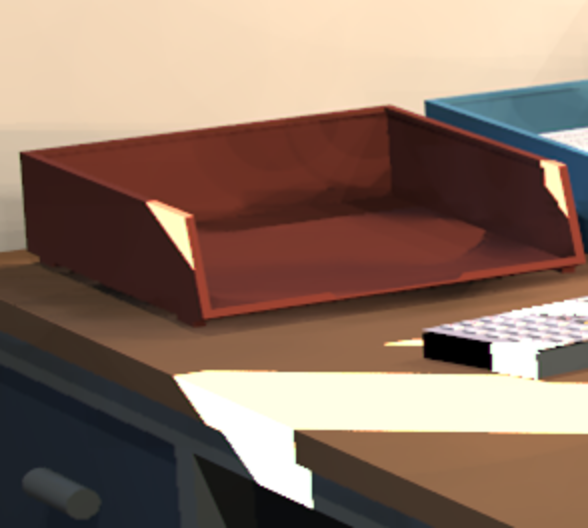}}
\caption[Cropped sections from rendering]{Cropped sections from original (left) distorted (center) and corrected (right) rendering}
\label{fig:rendering-comparison}
\end{figure}

\begin{figure}\centering\includegraphics[width=0.7\textwidth,trim=.3cm 0cm 1.6cm 1.0cm,clip]{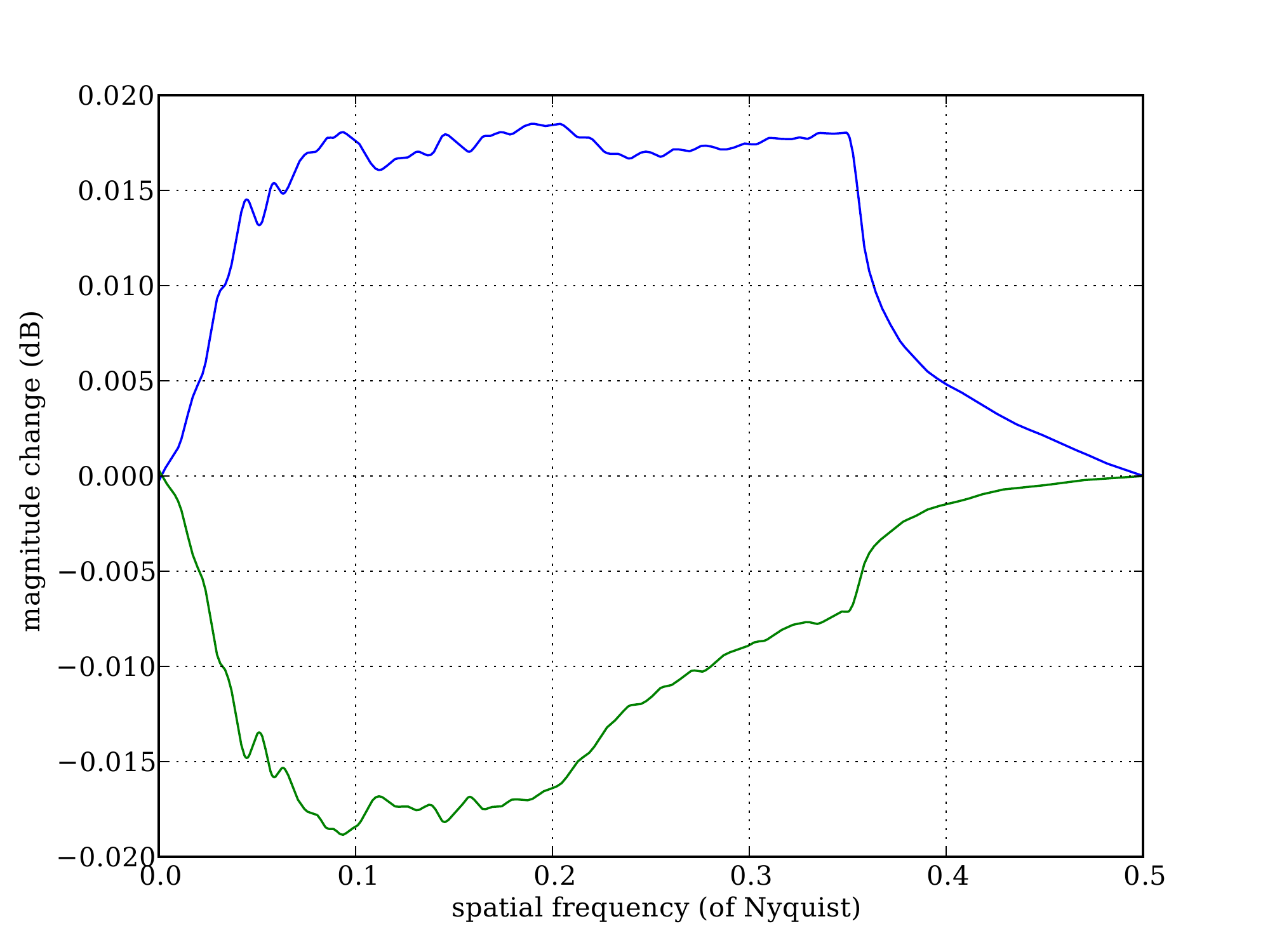}\caption[Change in spatial frequency in rendered images]{Change in spatial frequency from undistorted to distorted (blue) and distorted to corrected (green) rendered images}\label{gph:rendering-improvement}\end{figure}

Distortion on the ray-traced scene is executed with the known coefficients, then the correction algorithm is iterated, shown in Figure~\ref{fig:rendering-comparison}. Figure~\ref{gph:rendering-improvement} illustrates the increase in luminance spatial frequency due to LCA, shown in blue. The high frequencies approaching half-Nyquist (i.e.\ one-pixel width) have been attenuated as expected. The algorithm's correction, shown in green illustrates excellent performance at lower spatial frequencies, mirroring the shape and magnitude of the distortion with approaching zero frequency loss. At spatial frequencies above $f = 0.2N$, significant attenuation to the correction is seen. Since all the pixels in the image have been remapped with interpolation and antialias filtering, this is expected. Tests conducted in the next section will identify the impact of this on real-world image data, where spatial frequency from a lens system and Bayer array is far less than half-Nyquist.

\section{Performance analysis}

\subsection{Real-world test image}

\begin{figure}\centering
\subfigure{\includegraphics[width=0.4960\textwidth]{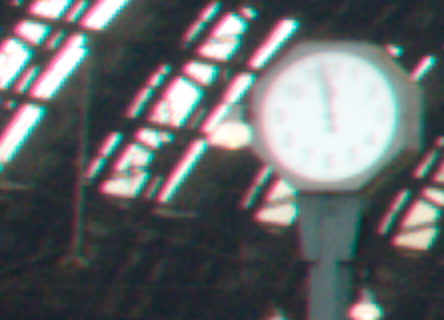}}
\vspace{-7pt}
\subfigure{\includegraphics[width=0.4960\textwidth]{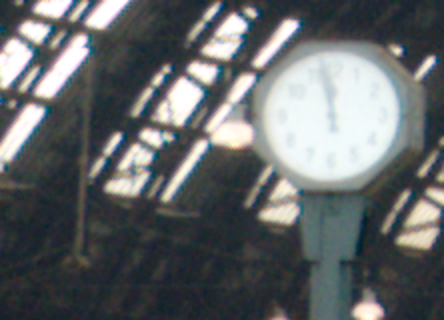}}
\subfigure{\includegraphics[width=0.4960\textwidth]{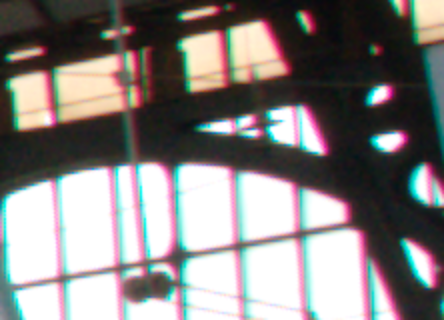}}
\vspace{-7pt}
\subfigure{\includegraphics[width=0.4960\textwidth]{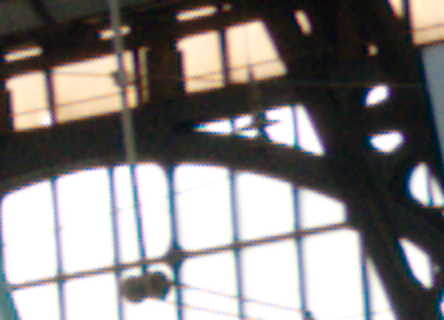}}
\subfigure{\includegraphics[width=0.4960\textwidth]{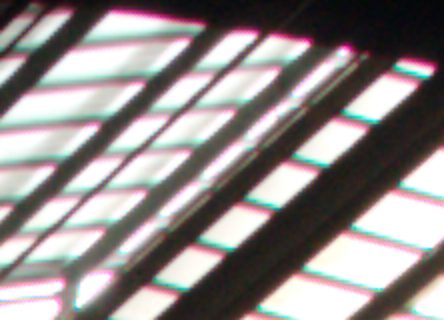}}
\vspace{-7pt}
\subfigure{\includegraphics[width=0.4960\textwidth]{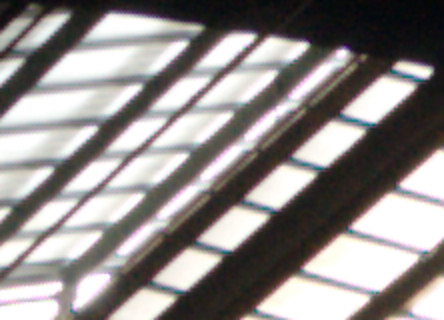}}
\subfigure{\includegraphics[width=0.4960\textwidth]{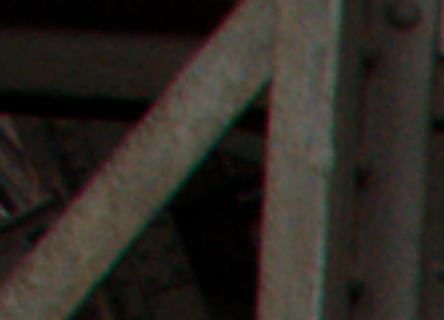}}
\vspace{-7pt}
\subfigure{\includegraphics[width=0.4960\textwidth]{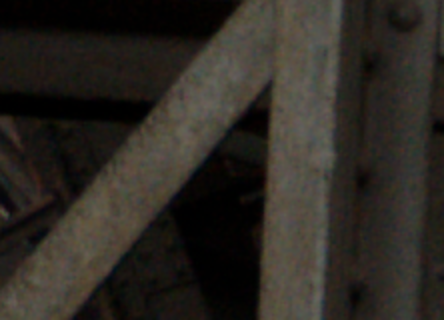}}
\caption[Cropped sections from photograph]{Cropped sections from original (left) and corrected (right) photograph}
\label{fig:lca-comparison}
\end{figure}

Firstly, the real-world example previously shown in Figure~\ref{fig:lca-example} to illustrate LCA was selected. The LCA-correction algorithm was applied, shown in Figure~\ref{fig:lca-comparison}. No residual LCA can be visually observed from the crops, which is the desired outcome. Since simple linear interpolation was used and not more advanced demosaicing algorithms, expected `zipper'-like artifacts are observed; for analysis without this, pixel remapping would have to be performed prior to demosaicing. If this were performed, advanced demosaicing would enhance quality further, rather than simple linear demosaicing.

To visualise the changes, difference maps are generated by taking the absolute difference of each corresponding pixel in the blue and green planes, and separately for the red and green planes; white edges would be seen where chromatic differences exist including due to LCA\@. It is worthwhile noting that even for an image with zero LCA (e.g.~an artificial image), there will be a constant difference map total area which the correction algorithm would not be able to minimise further, thus there is a fixed offset in addition to the edges due to the LCA\@. As a result, it is seen from Figure~\ref{fig:difference-comparison} that there is a significant reduction in structure comparing the uncorrected to corrected difference maps, particularly in the blue-green plane difference.

\begin{figure}\centering\includegraphics[width=0.7\textwidth,trim=.3cm 0cm 1.6cm 1.0cm,clip]{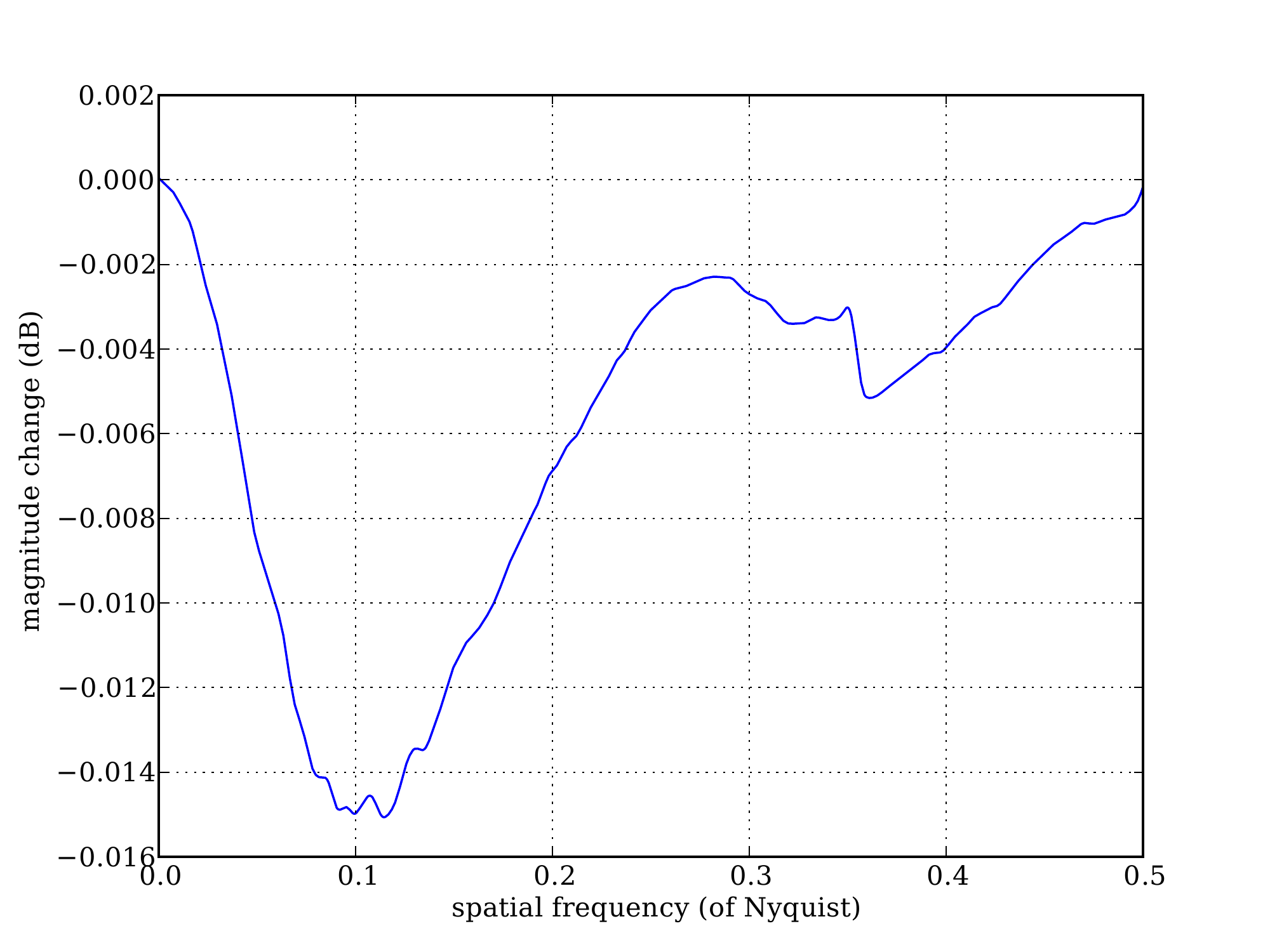}\caption{Change in spatial frequency due to LCA correction}\label{fig:fft-change}\end{figure}

Next the quantification algorithm was executed to quantify the spatial frequency changes between the original and corrected image, shown in Figure~\ref{fig:fft-change}. A significant reduction in spatial frequency of peak 0.015dB is observed. This translates into information which is more useful to the HVS both in luminosity and chromaticity, and the increase in \textit{homogeneity} can be observed by the picture looking `clearer' as a result.

\subsection{Run-time cost}
\begin{figure}
\begin{tikzpicture}[x=0.4mm]
  \foreach \x in {50,100,150,200,250}
    \draw[help lines] (\x,0) node[below,black] {\x} -- (\x,2.1);
 
  \draw (51,0.6) node [fill=white,minimum size=16pt] {};
  
  \foreach \yoff/\start/\end/\name/\color in {
  	1.5/0./2.//blue, 1.5/2./16//green, 1.5/16/257/257/red,
  	0.6/0./2.//blue, 0.6/2/7.75//green, 0.6/7.75/37.8/37.8/red}
  {
    \filldraw[yshift=\yoff cm,fill=\color!50] (\start,-0.3) rectangle (\end,0.3) node [xshift=.6cm,yshift=-0.3cm] {\name};
  }

  \draw [very thick] (0,0) -- (260,0);
  \draw [very thick] (0,0) node[below] {0} -- (0,2.1);
  \node at (125,-0.7) {seconds};

  \draw (0,1.5) node[anchor=east] {correction};
  \draw (0,0.6) node[anchor=east] {quantification};
\end{tikzpicture}
\caption{Runtime breakdown of algorithms}
\label{fig:runtime}
\end{figure}
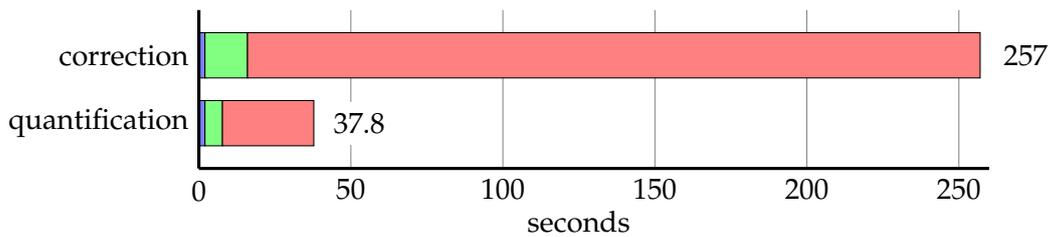

Since the correction and quantification algorithms are highly \textit{computationally intensive} and intentionally avoid subsetting the data they measure for robustness, the runtime can be consequently significant, shown in Figure~\ref{fig:runtime}; time was measured on an Intel Core i5-661 desktop platform.

For the correction algorithm, the runtime cost is dominated by the image size (in pixels) and the efficacy of the L-BFGS-B algorithm minimising the error in a non-linear multivariate problem; this influences the number of iterations of the relatively expensive error function over the image. The majority of time within the error function is spent on non-linear pixel remapping, which produces a lot of out-of-cache access. OpenCV is internally used for the remapping without a highly-optimising compiler, thus this could benefit significantly from optimal load\slash store scheduling and ideal prefetch scheduling to avoid frequent processor pipeline stalls due to waiting for data to arrive from the memory controller. The left-most and small blue bar indicates image loading and plane splitting (0.7\%); the green bar shows the total time subtracting image planes and computing the masked average pixel value (5.4\%); the red bar represents the time generating the pixel remapping offsets performing the interpolated lookups (93.9\%).

The quantification algorithm has a far shorter runtime, as no iteration is performed; the left-most and small blue bar shows that loading both input images has a similar overhead as in the correction algorithm (5.3\%). The two DFT transformations shown in the green bar account for 15\% of the overall time, while the rest of the computation time is shown in the red bar (79.7\%) and is due to binning and accumulating the DFT output based on radius, i.e.~spatial frequency.



\subsection{Suite of images}
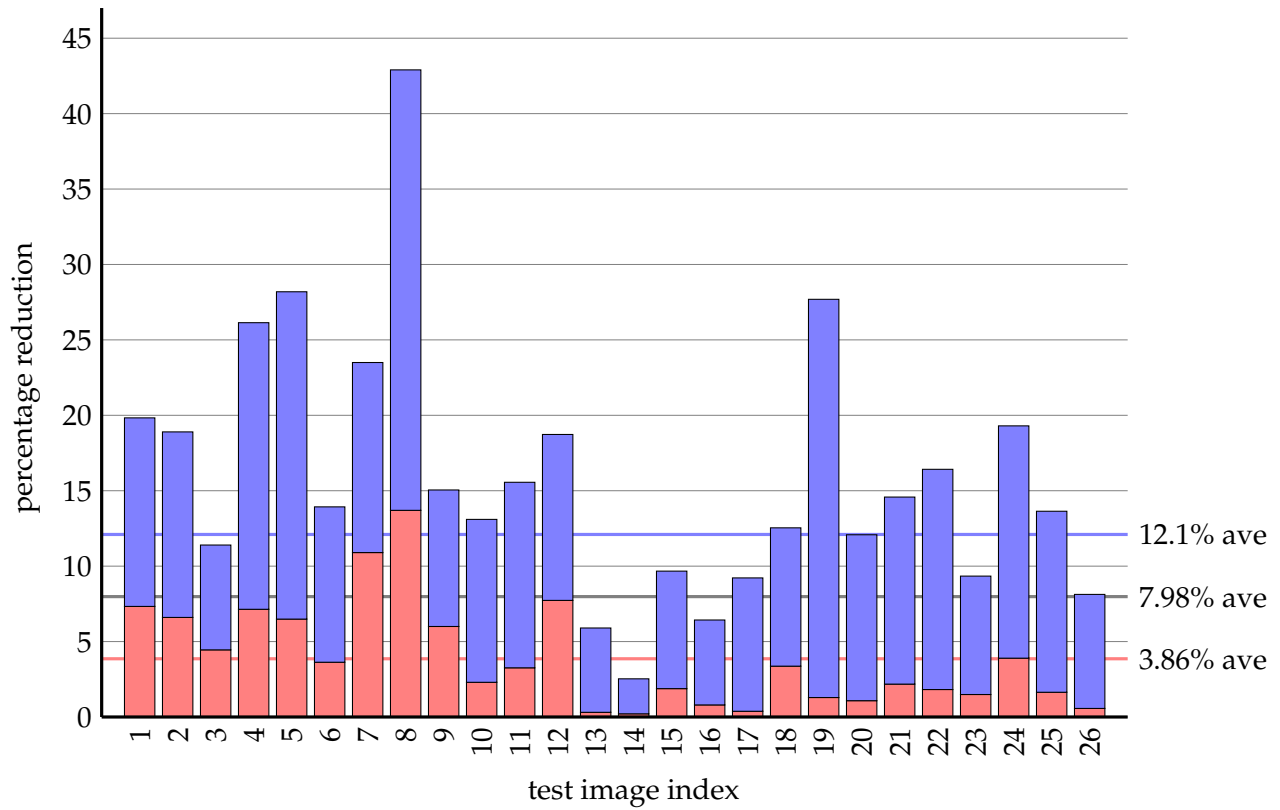
\begin{figure}
\begin{tikzpicture}[y=0.2cm]
  \foreach \y in {0,5,10,15,20,25,30,35,40,45}
    \draw[help lines] (0,\y) node[left,black] {\y} -- (13.5,\y);

  \draw [black!50,very thick] (0,7.98) -- (13.5,7.98) node[right,black] {7.98\% ave};
  \draw [red!50,very thick] (0,3.86) -- (13.5,3.86) node[right,black] {3.86\% ave};
  \draw [blue!50,very thick] (0,12.1) -- (13.5,12.1) node[right,black] {12.1\% ave};

  \foreach \x/\heightA/\heightB in {
1/7.33/12.5,
2/6.60/12.3,
3/4.45/6.95,
4/7.14/19.0,
5/6.49/21.7,
6/3.63/10.3,
7/10.9/12.6,
8/13.7/29.2,
9/6.00/9.05,
10/2.30/10.8,
11/3.26/12.3,
12/7.73/11.0,
13/0.31/5.59,
14/0.21/2.32,
15/1.88/7.79,
16/0.80/5.63,
17/0.38/8.84,
18/3.37/9.17,
19/1.29/26.4,
20/1.08/11.0,
21/2.18/12.4,
22/1.82/14.6,
23/1.49/7.85,
24/3.90/15.4,
25/1.64/12.0,
26/0.567/7.56}
  {
    \draw (\x/2.,0) node[rotate=90,anchor=east] {\x};

    \begin{scope}[xshift=\x cm/2.]

    \filldraw[fill=red!50] (-.2,0) rectangle (.2,\heightA);
    \filldraw[fill=blue!50] (-.2,\heightA) rectangle (.2,\heightA+\heightB);
    \end{scope}
  }

  \draw [very thick] (0,0) -- (0,47);
  \draw [very thick] (0,0) -- (13.5,0);
  \node at (7,-5) {test image index};
  \node at (-1,22.5) [rotate=90] {percentage reduction};

\end{tikzpicture}
\caption[Difference map area reduction]{Difference map area reduction. Blue bars show blue-green edge reduction; red bars show red-green edge reduction.}
\label{fig:improvement}
\end{figure}

The improvement is reflected in the reduction in the area of the plane-difference maps, compared in graphical form in Figure~\ref{fig:improvement}. The reduction in the red to green and the blue to green difference map area are as shown in Table~\ref{tab:edge-reduction}. This illustrates just the reduction in difference areas the correction algorithm was able to achieve; inspection of the resulting corrected images confirmed that all the minima were true global minima and not leading to misconvergence, which would manifest as introduction of additional LCA artifacts. Since the difference area without LCA is not known, this relative reduction in difference area carries no \textit{absolute meaning}. Later, the quantification algorithm will be employed to measure differences with more significance to the HVS.

\begin{figure}\centering\includegraphics[width=0.7\textwidth,trim=.3cm 0cm 1.6cm 1.0cm,clip]{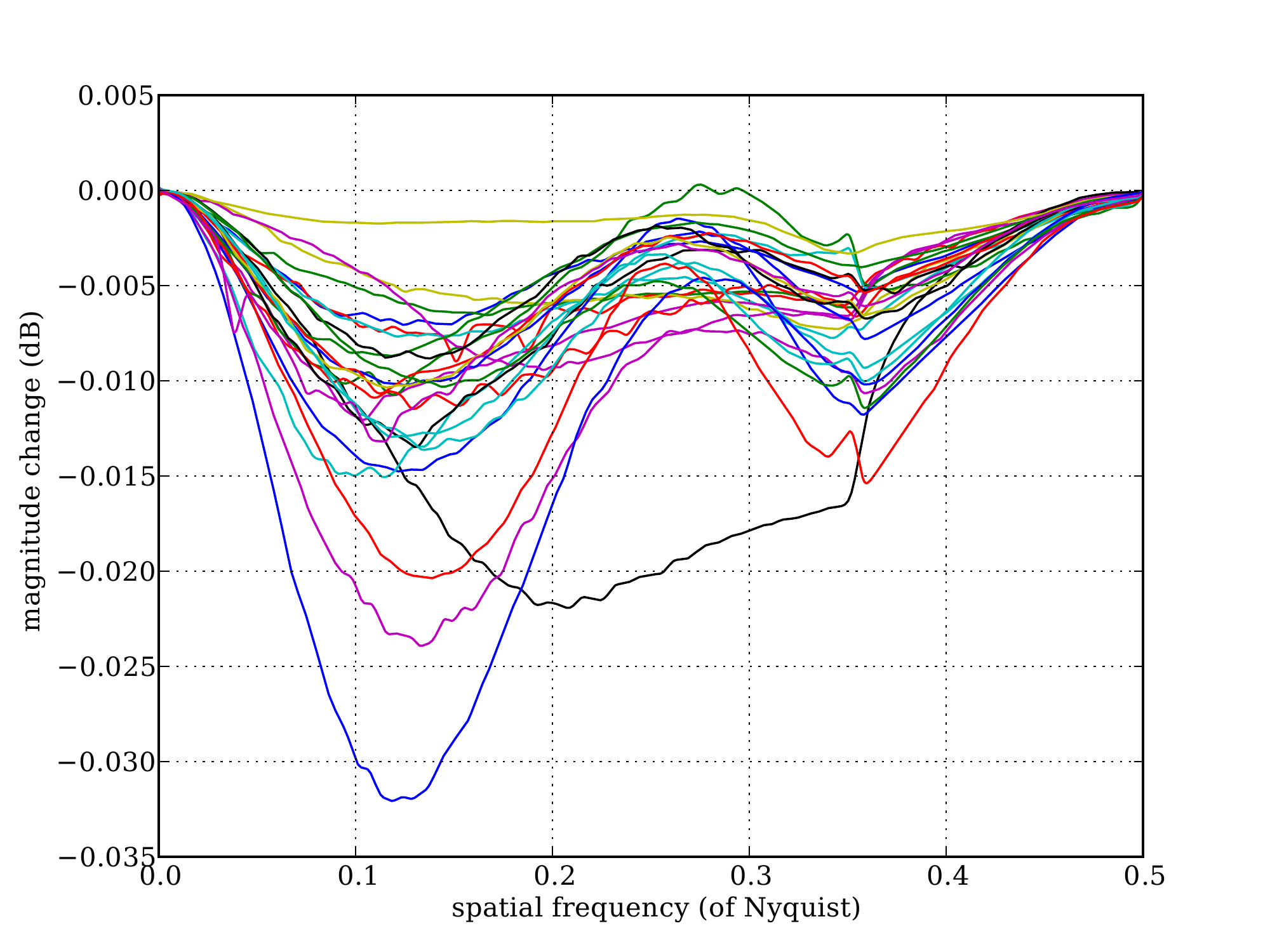}\caption[Change in spatial frequency for test images]{Change in spatial frequency for test images. Colour shows image index.}\label{gph:testimages}\end{figure}

The correction algorithm was executed for all of the images in the bank of representative real-world test images, shown earlier in Figure~\ref{fig:test-images}. The change in spatial frequency for all images was quantified and plotted together, shown in Figure~\ref{gph:testimages}.

\ctable[botcap,caption=Difference map area reduction,label={tab:edge-reduction}]{lll}{}{\FL
Image & R-G reduction & B-G reduction \ML
1 & 7.33\% & 12.5\% \NN
2 & 6.60\% & 12.3\% \NN
3 & 4.45\% & 6.95\% \NN
4 & 7.14\% & 19.0\% \NN
5 & 6.49\% & 21.7\% \NN
6 & 3.63\% & 10.3\% \NN
7 & 10.9\% & 12.6\% \NN
8 & 13.7\% & 29.2\% \NN
9 & 6.00\% & 9.05\% \NN
10 & 2.30\% & 10.8\% \NN
11 & 3.26\% & 12.3\% \NN
12 & 7.73\% & 11.0\% \NN
13 & 0.31\% & 5.59\% \NN
14 & 0.21\% & 2.32\% \NN
15 & 1.88\% & 7.79\% \NN
16 & 0.80\% & 5.63\% \NN
17 & 0.38\% & 8.84\% \NN
18 & 3.37\% & 9.17\% \NN
19 & 1.29\% & 26.4\% \NN
20 & 1.08\% & 11.0\% \NN
21 & 2.18\% & 12.4\% \NN
22 & 1.82\% & 14.6\% \NN
23 & 1.49\% & 7.85\% \NN
24 & 3.90\% & 15.4\% \NN
25 & 1.64\% & 12.0\% \NN
26 & 0.567\% & 7.56\% \ML
Average & 3.86\% & 12.1\% \LL}

\begin{figure}\centering
\subfigure{\includegraphics[width=0.2380\textwidth]{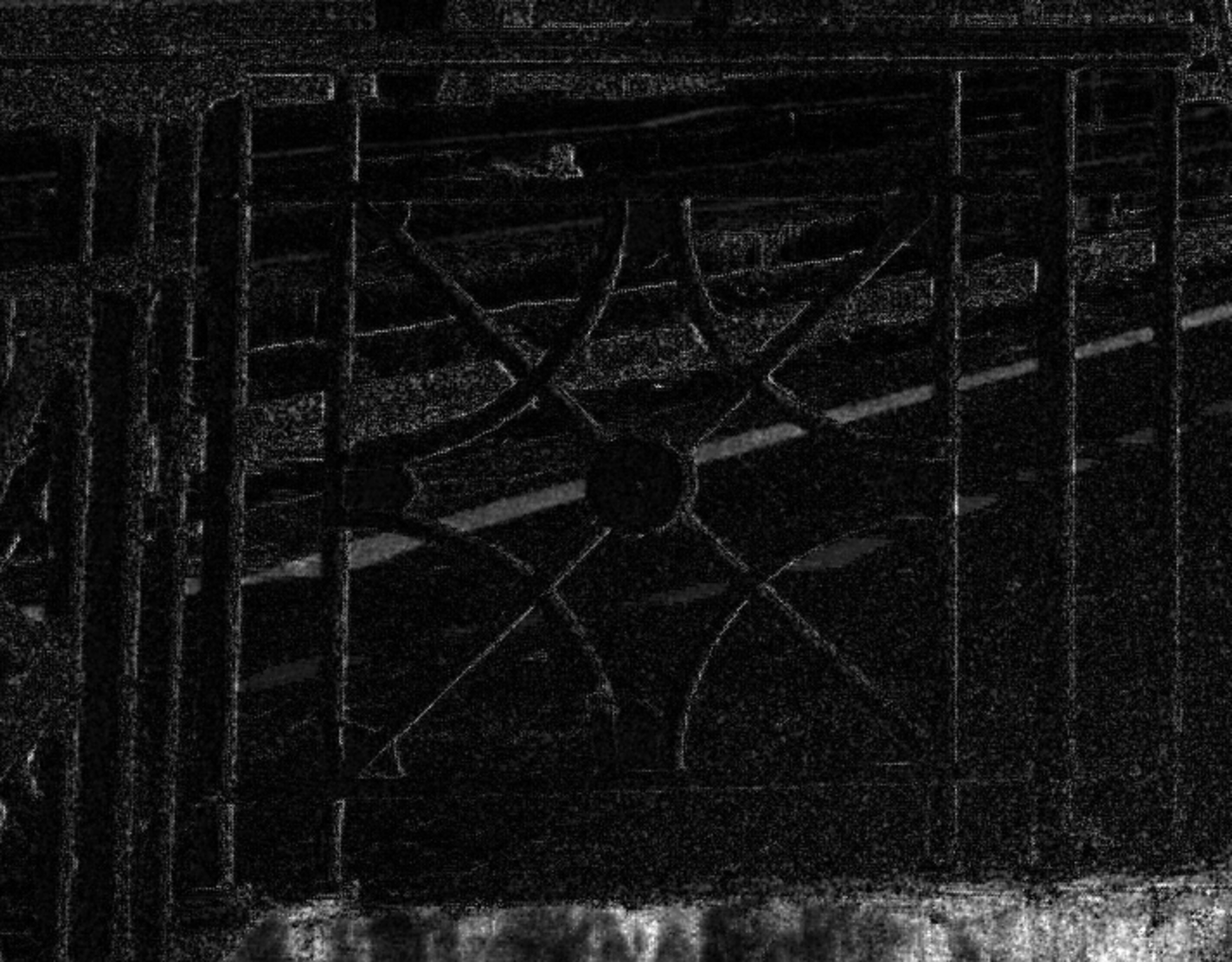}}
\subfigure{\includegraphics[width=0.2380\textwidth]{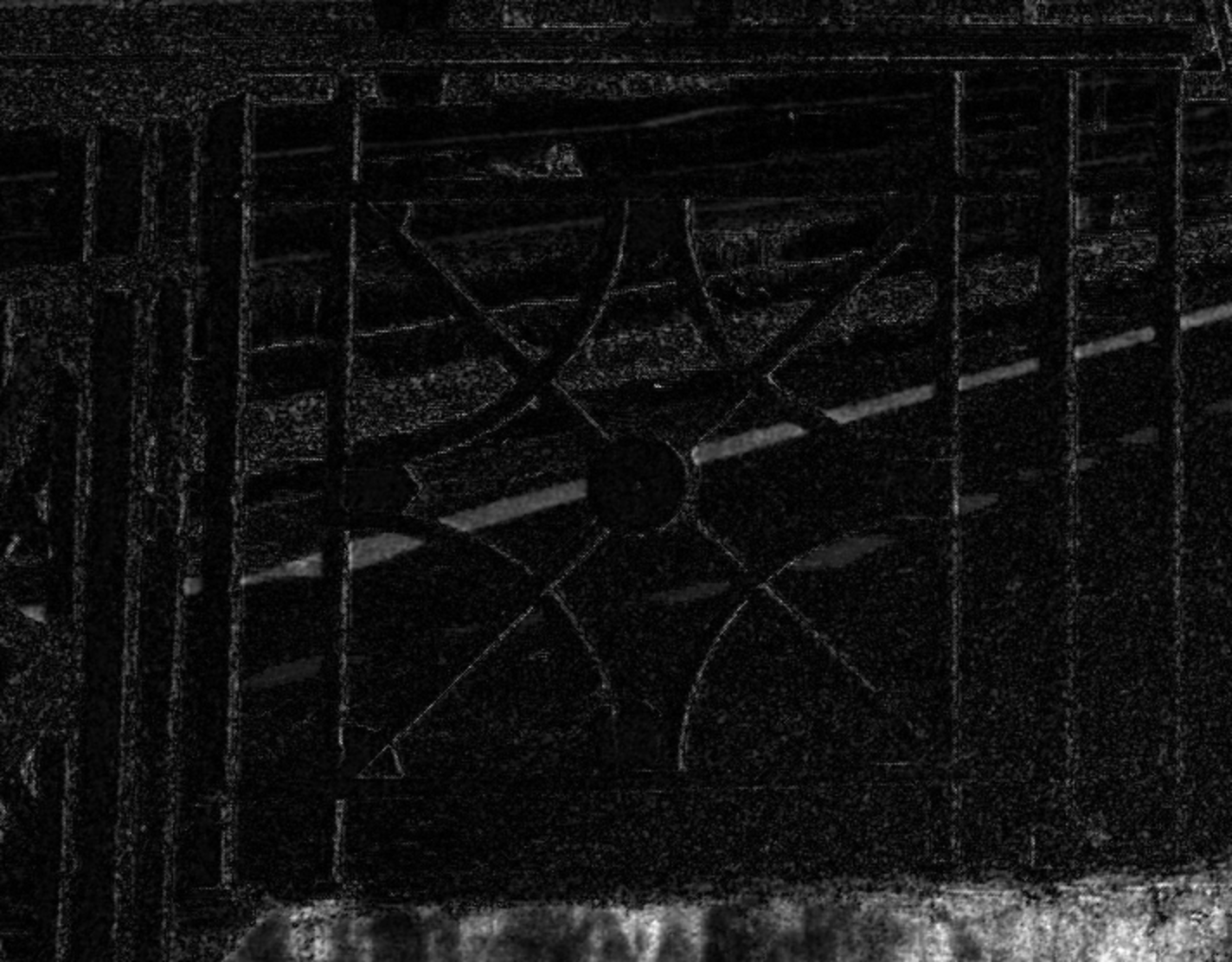}}
\subfigure{\includegraphics[width=0.2380\textwidth]{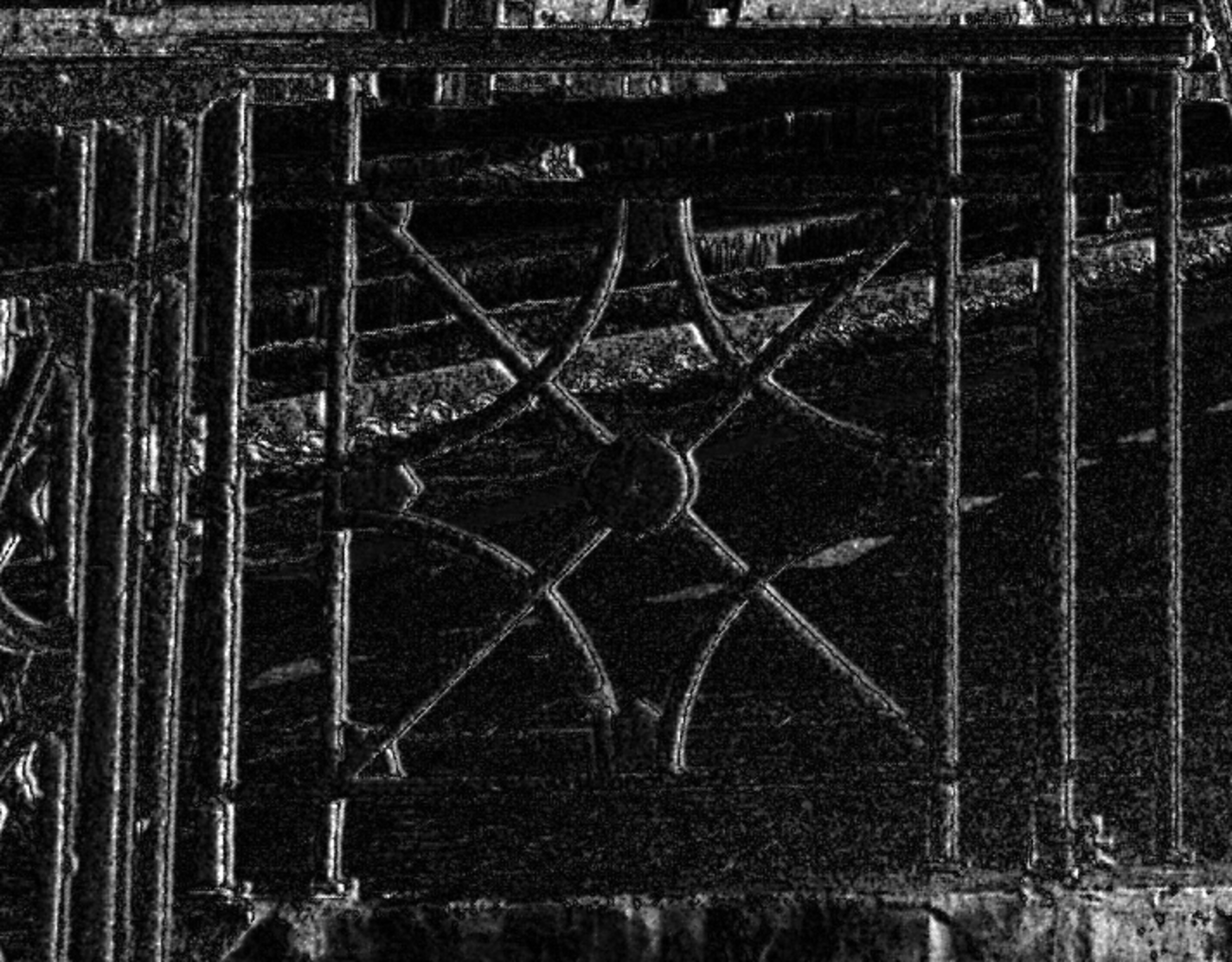}}
\vspace{-7pt}
\subfigure{\includegraphics[width=0.2380\textwidth]{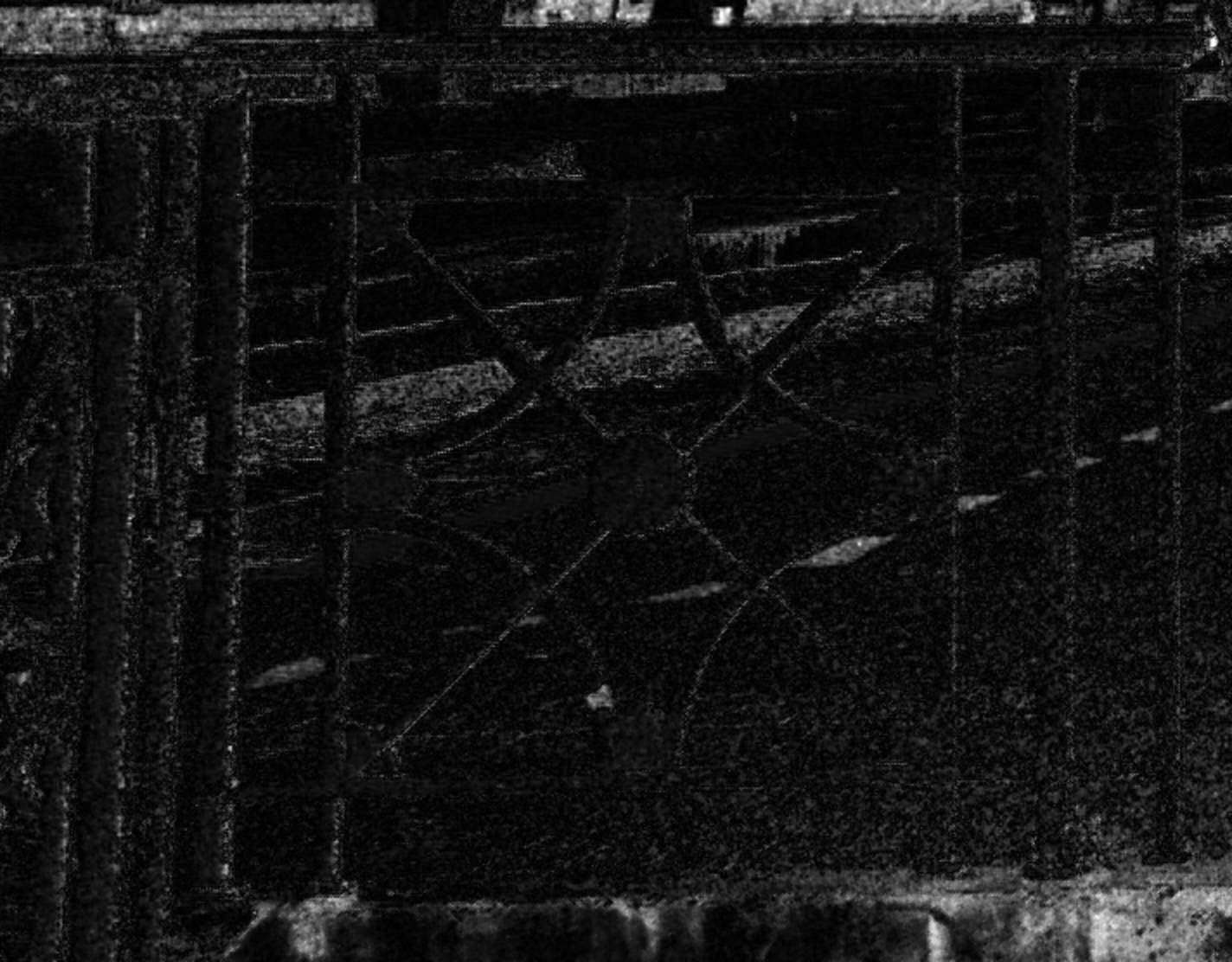}}
\subfigure{\includegraphics[width=0.2380\textwidth]{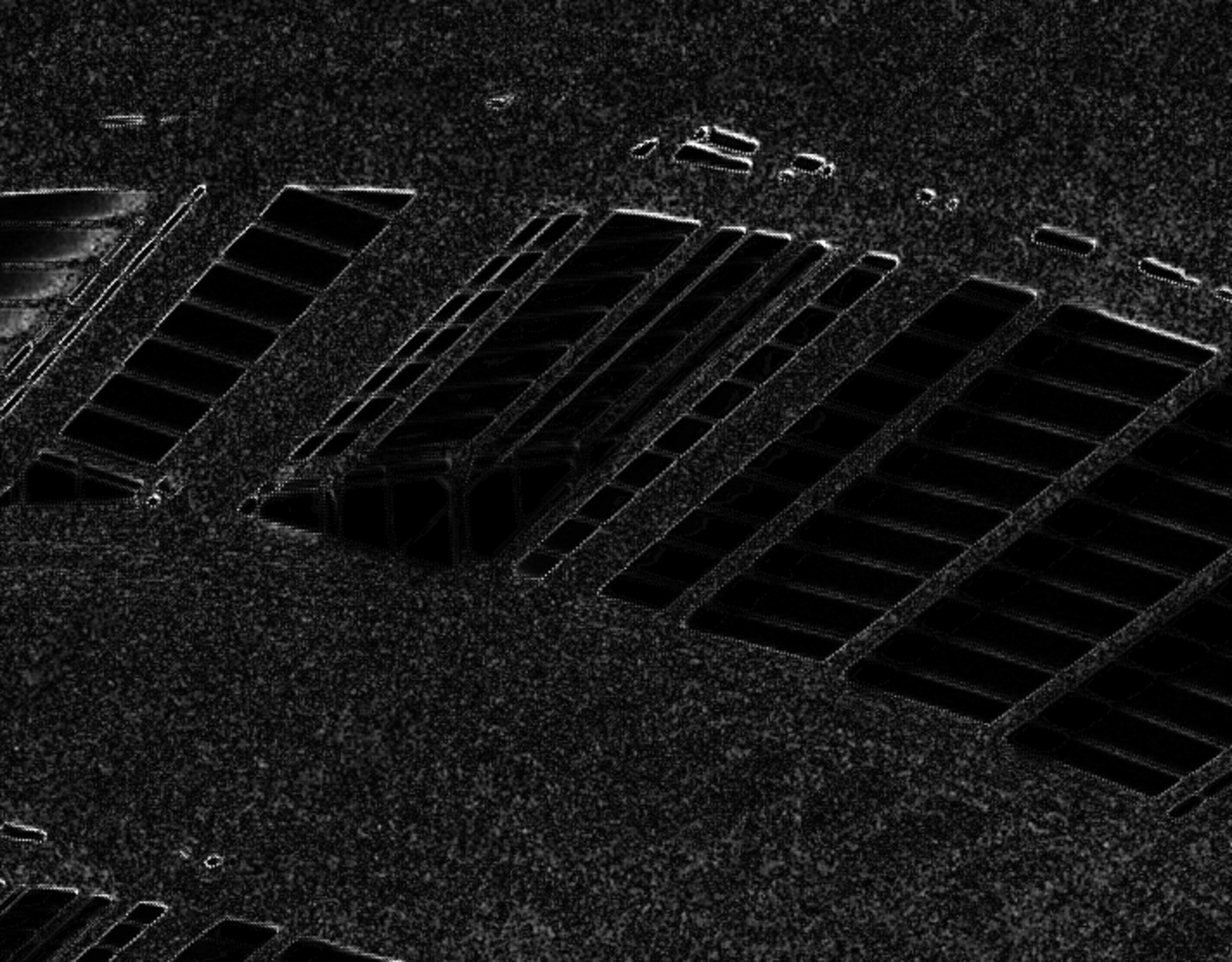}}
\subfigure{\includegraphics[width=0.2380\textwidth]{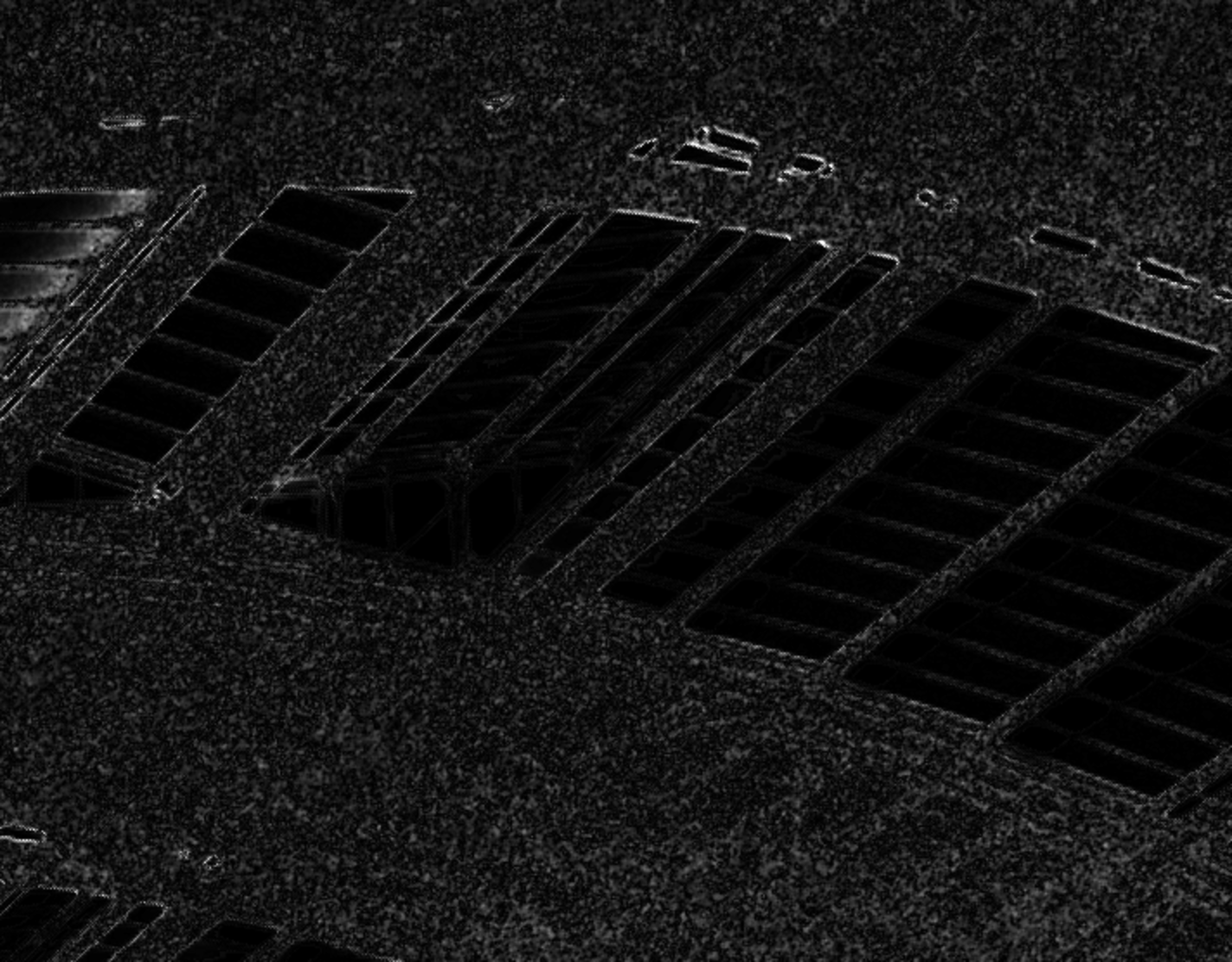}}
\subfigure{\includegraphics[width=0.2380\textwidth]{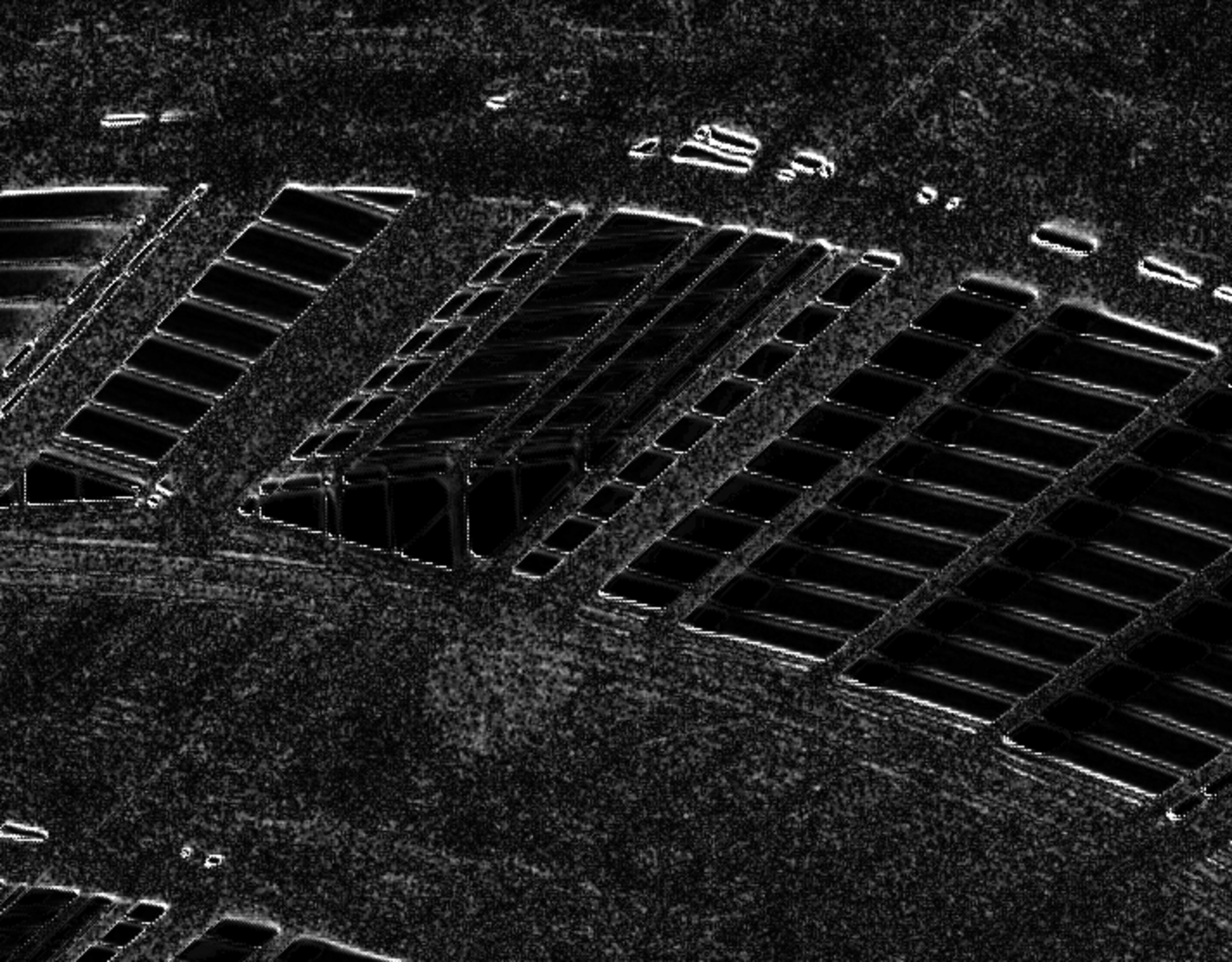}}
\vspace{-7pt}
\subfigure{\includegraphics[width=0.2380\textwidth]{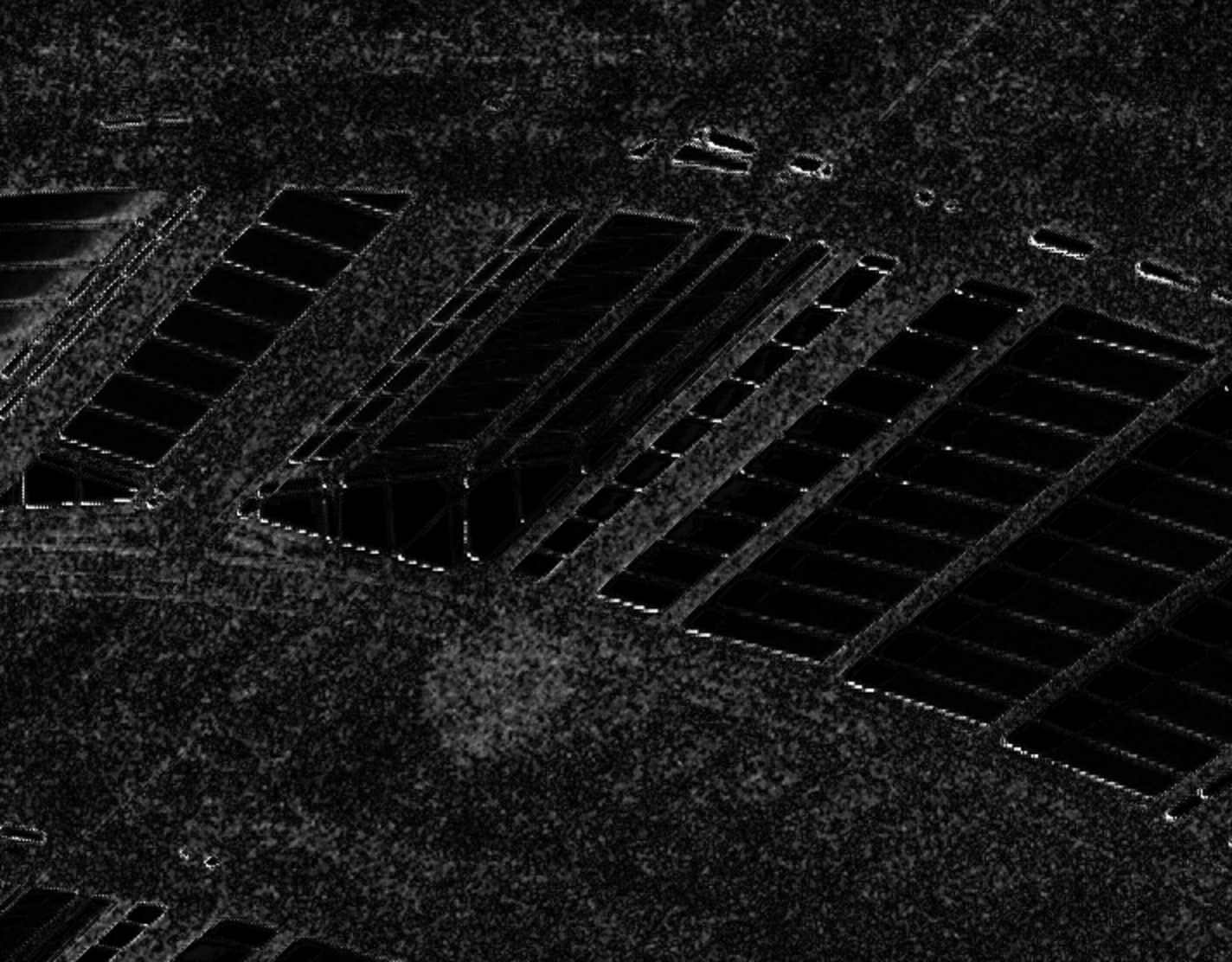}}
\subfigure{\includegraphics[width=0.2380\textwidth]{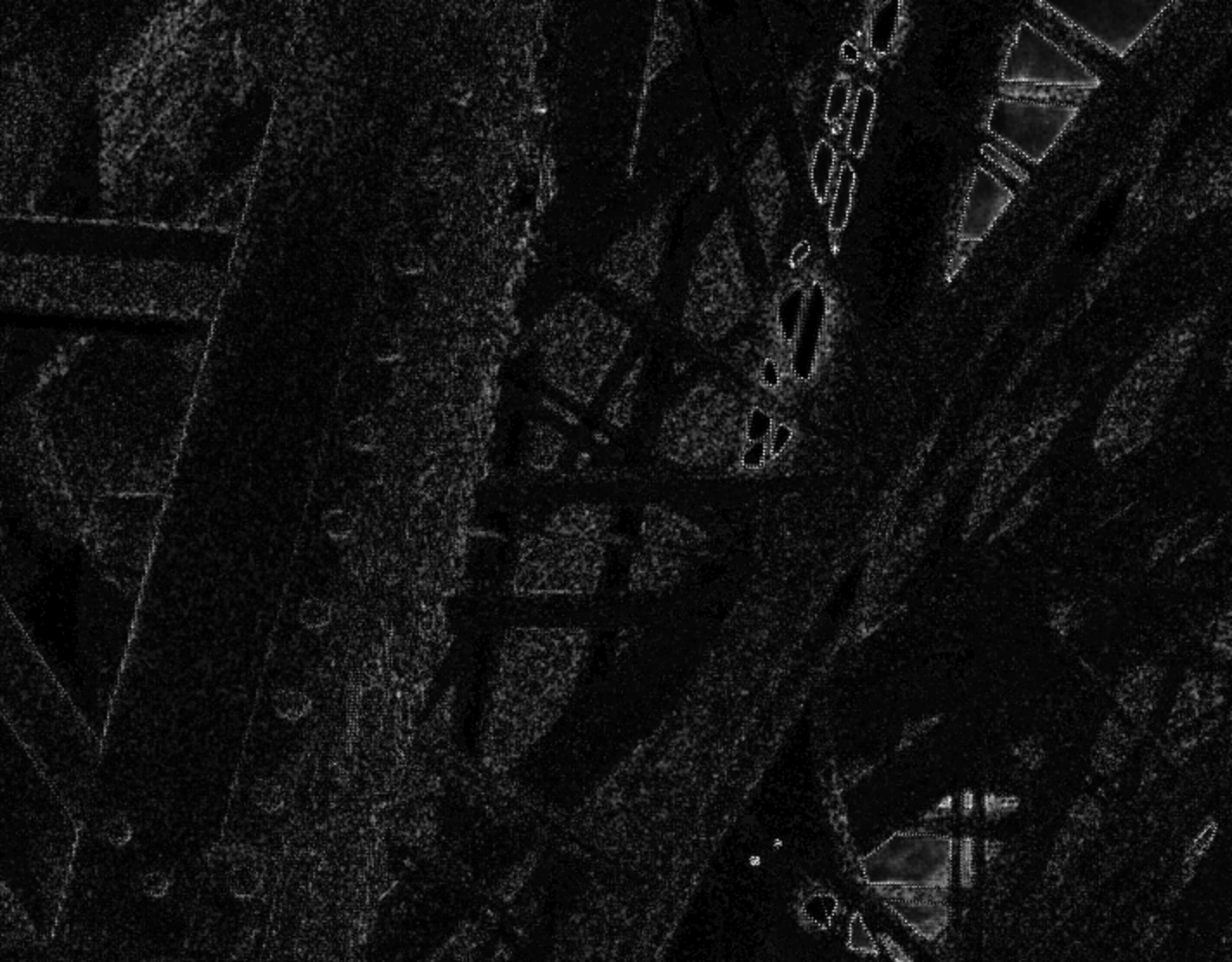}}
\subfigure{\includegraphics[width=0.2380\textwidth]{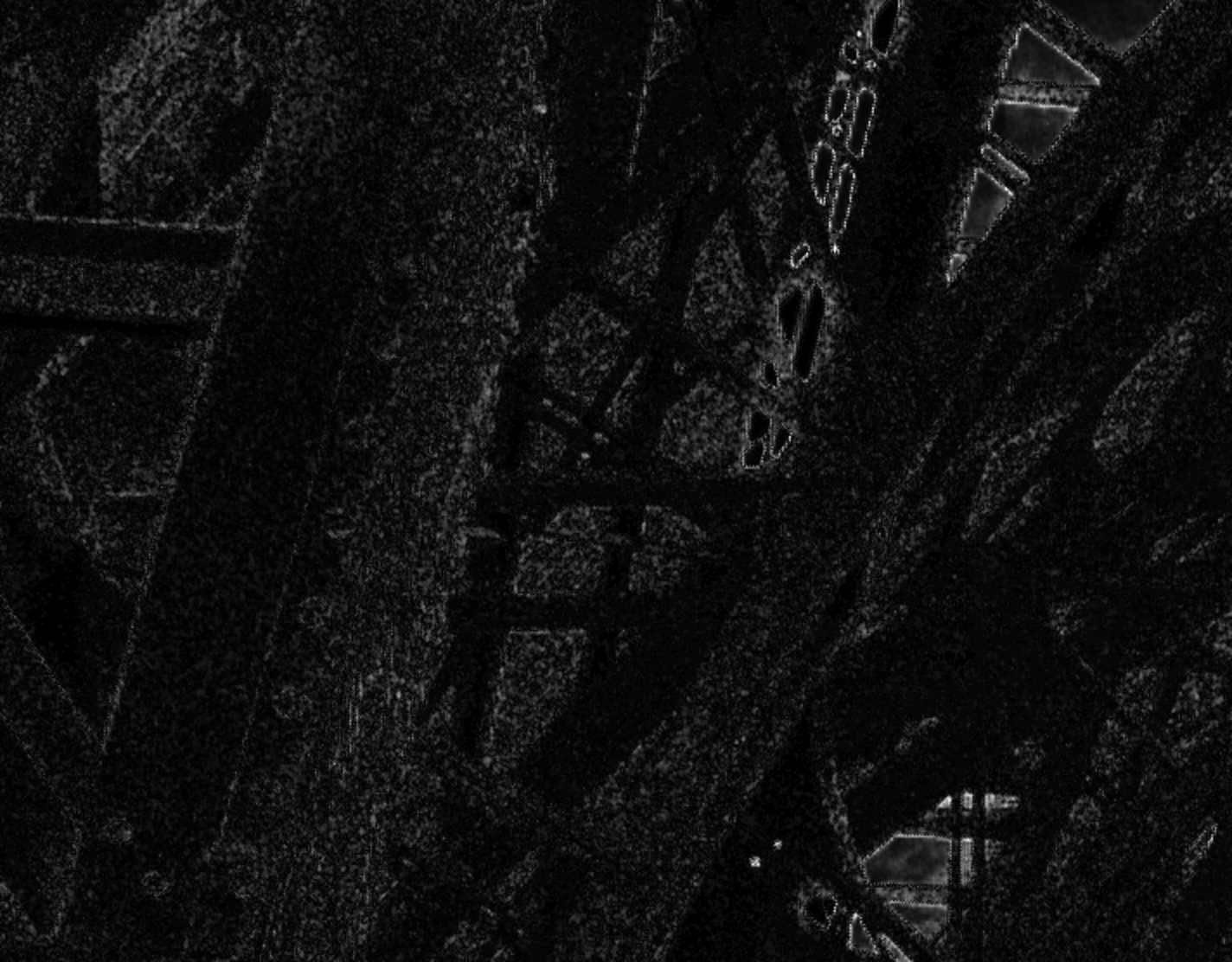}}
\subfigure{\includegraphics[width=0.2380\textwidth]{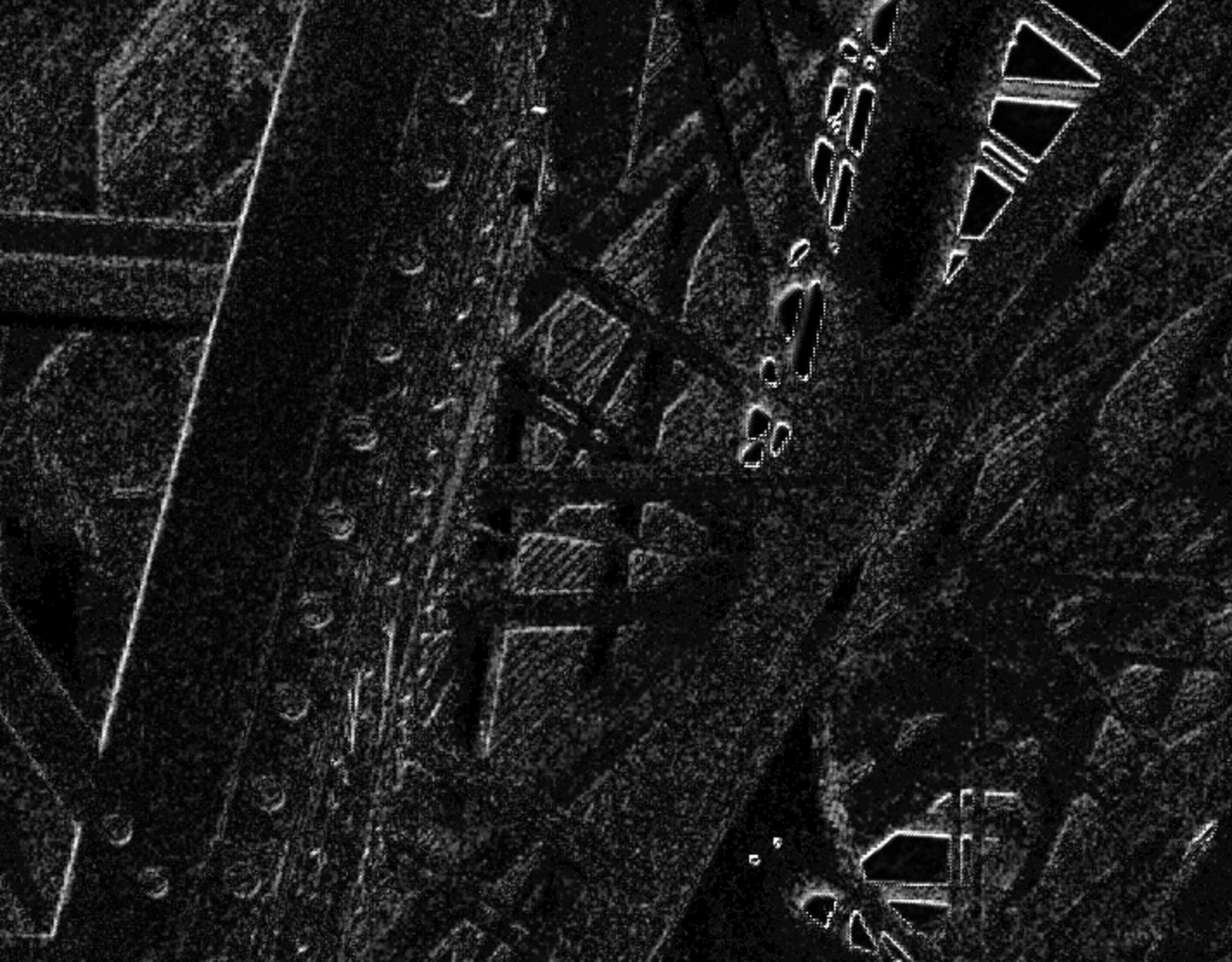}}
\vspace{-7pt}
\subfigure{\includegraphics[width=0.2380\textwidth]{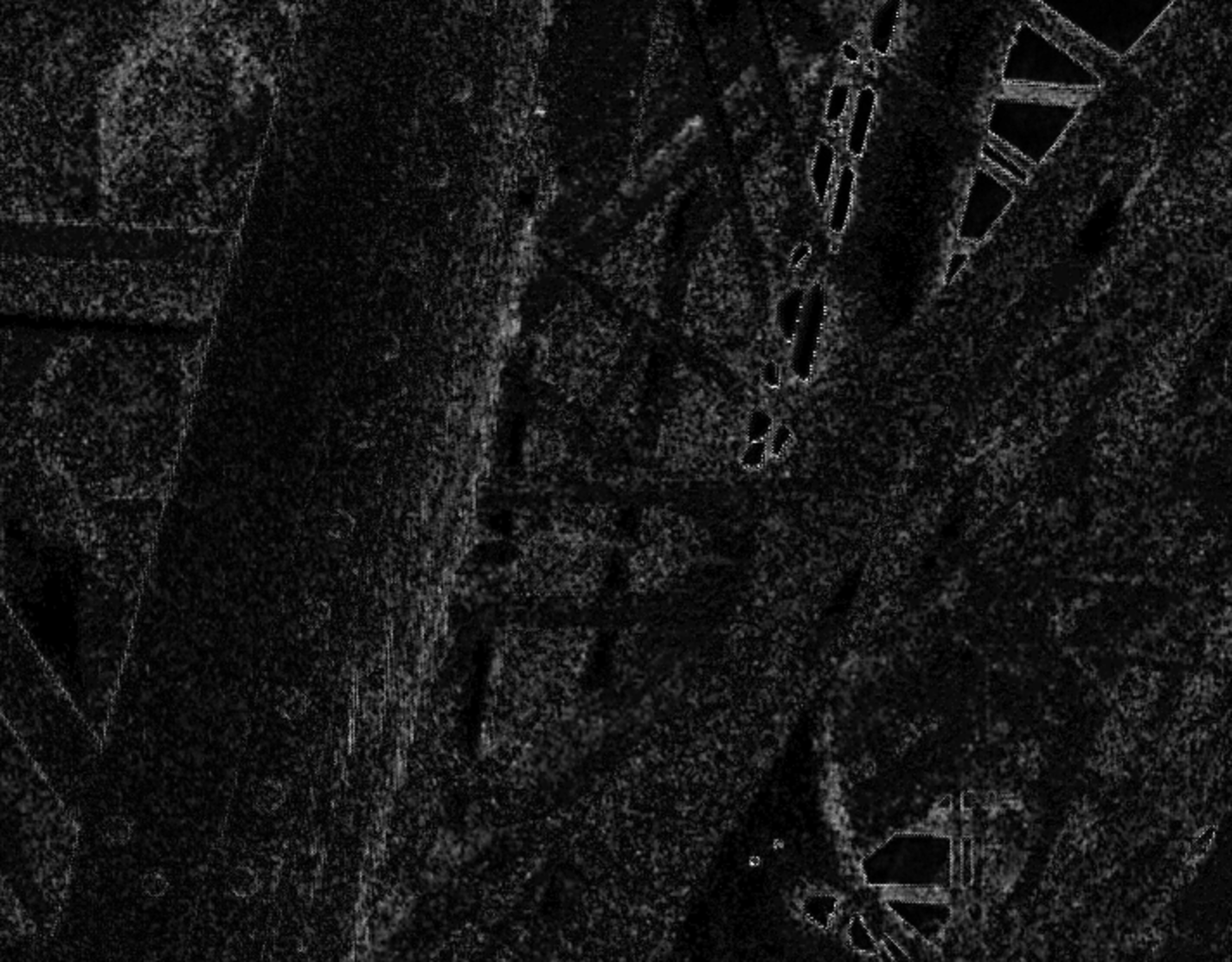}}
\subfigure{\includegraphics[width=0.2380\textwidth]{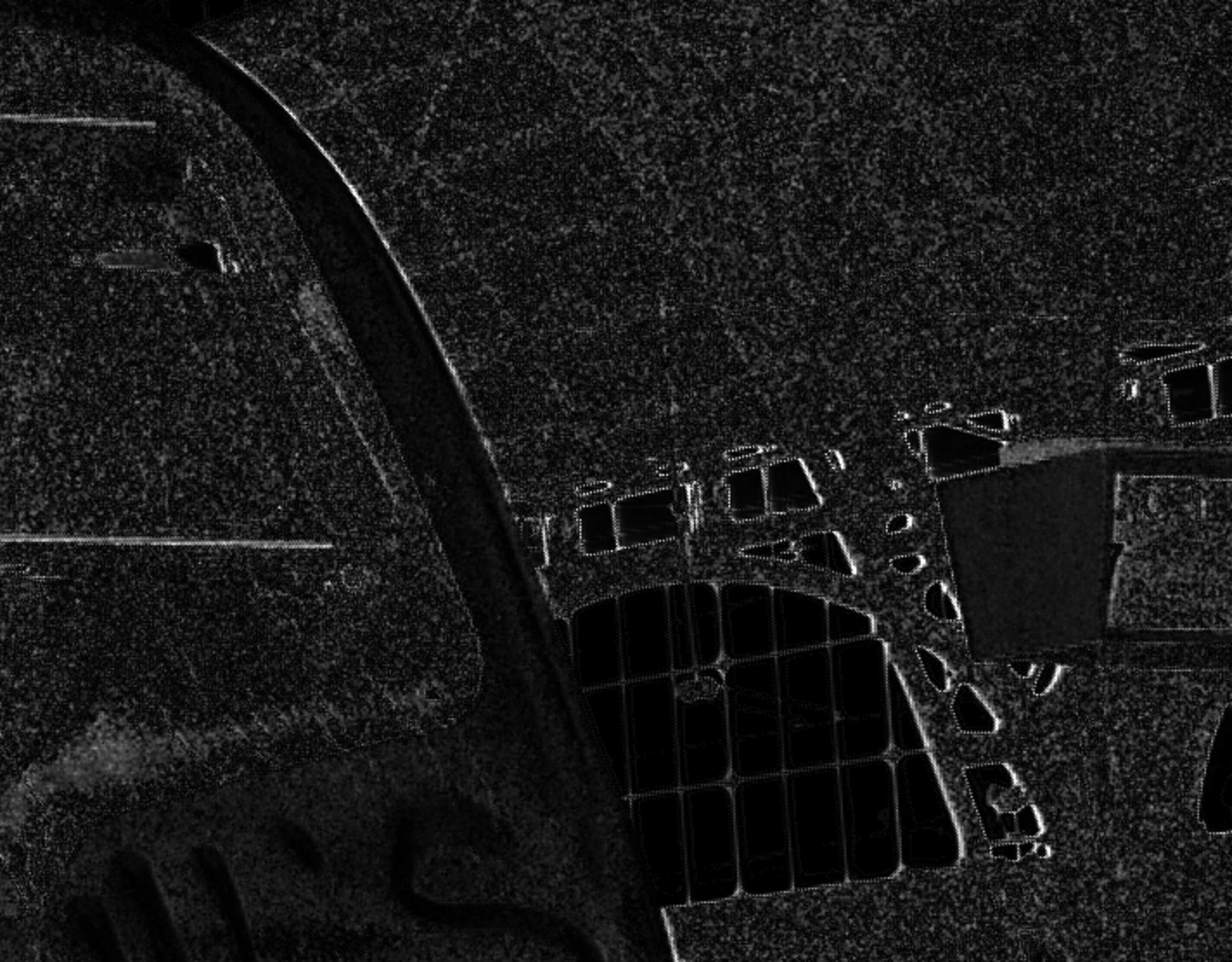}}
\subfigure{\includegraphics[width=0.2380\textwidth]{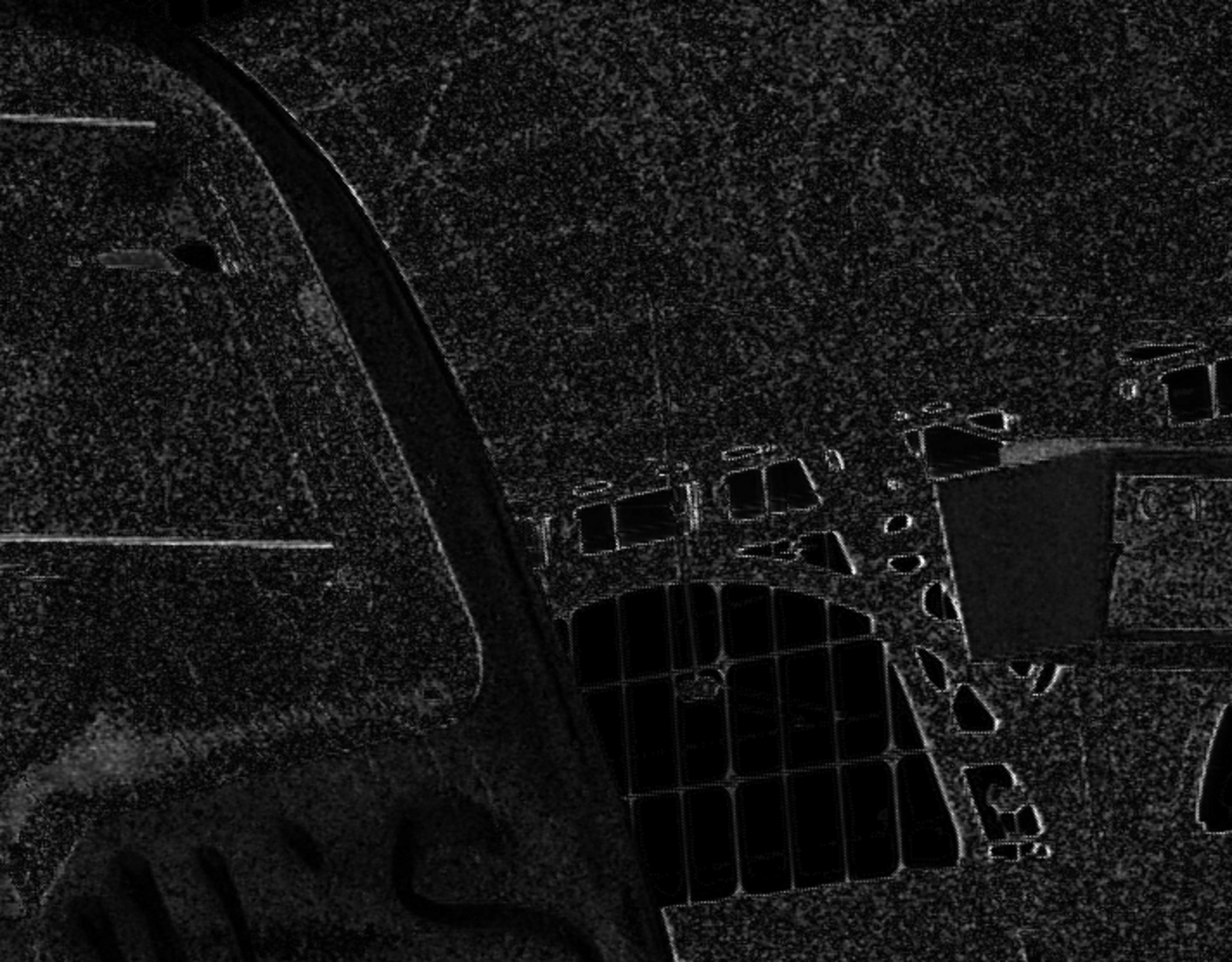}}
\subfigure{\includegraphics[width=0.2380\textwidth]{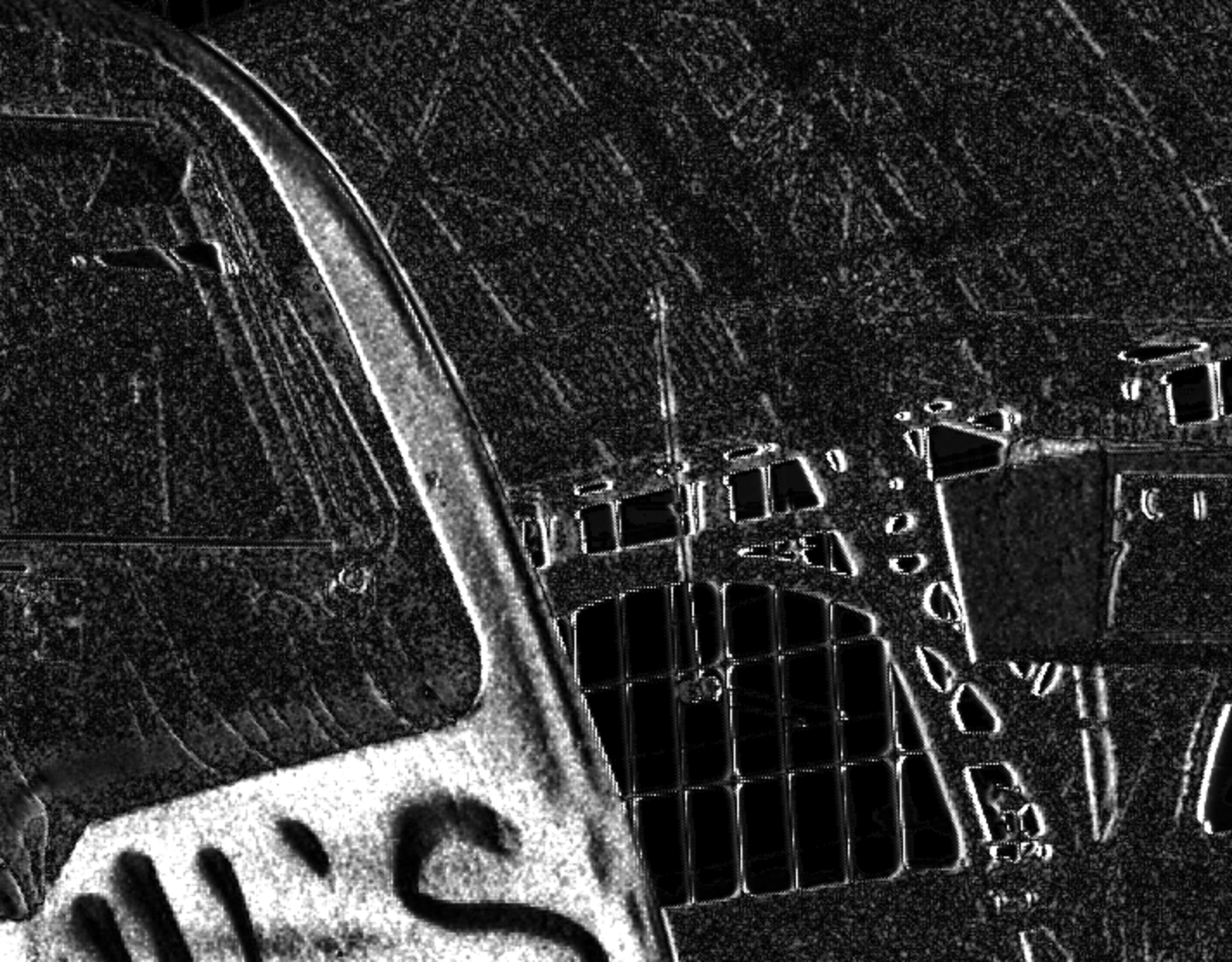}}
\vspace{-7pt}
\subfigure{\includegraphics[width=0.2380\textwidth]{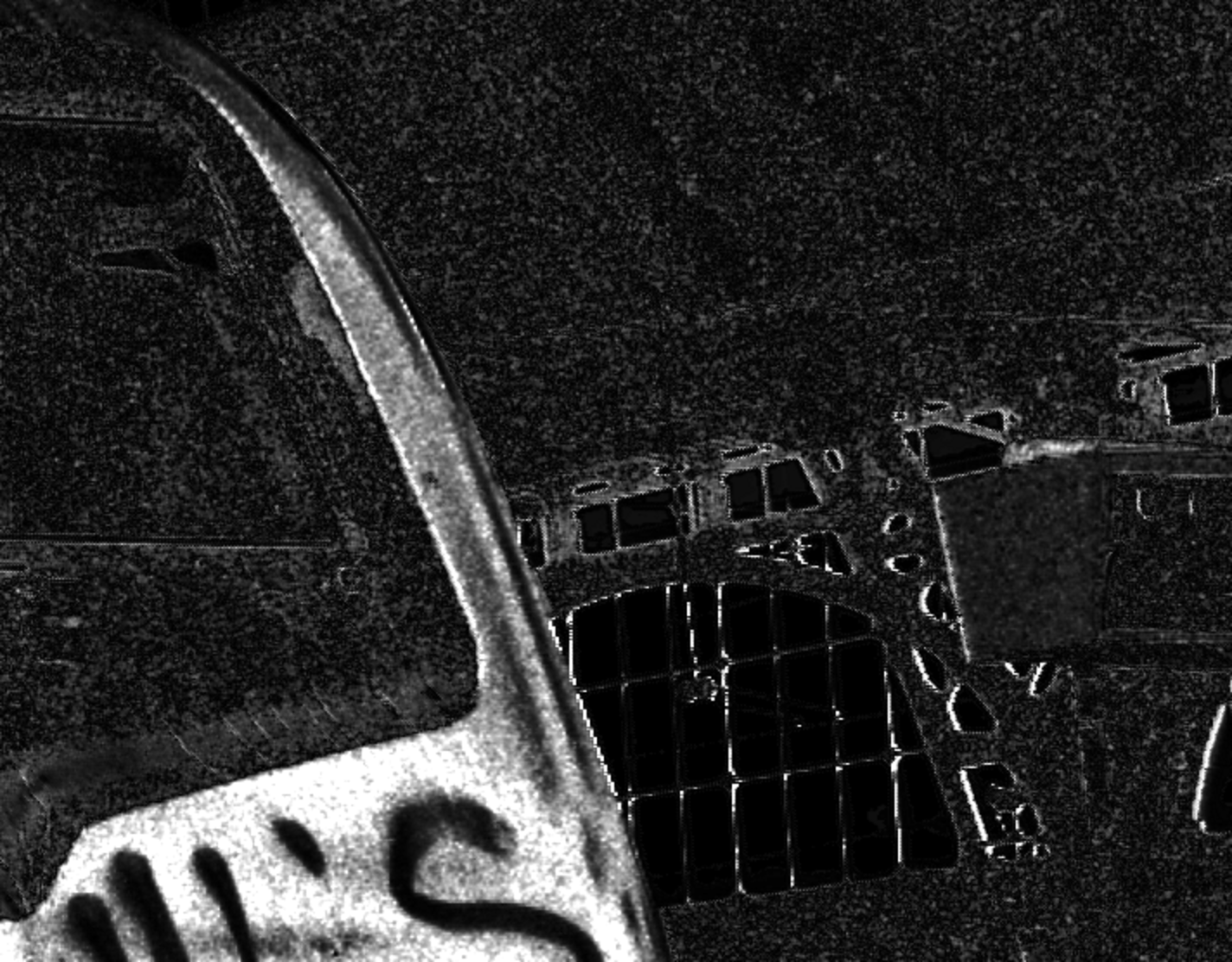}}
\caption[Cropped section comparison from difference maps]{Cropped sections from uncorrected red-green (far left), corrected red-green (left), uncorrected blue-green (right) and corrected blue-green (far right) difference maps}
\label{fig:difference-comparison}
\end{figure}








\section{Comparison with existing methods}
Three images were selected for testing against a widely-used solution for LCA elimination: Adobe Photoshop Creative Suite 4 \cite{Adobe09}. The demosaiced images that were used for proposed correction were also used in the Photoshop LCA correction to avoid comparing differing raw converter and internal processing steps. The procedure employed iteratively refined the three `Green\slash Magenta Fringe', `Blue\slash Yellow Fringe' and `Red\slash Cyan Fringe' controls, until LCA artifacts were eliminated as far as possible by inspection of a 200\% magnification of the image at a range of places throughout the image. Following this, quantification was employed to measure the improvement. The results are analysed in the next section.

\begin{figure}\centering
\subfigure{\includegraphics[width=\textwidth]{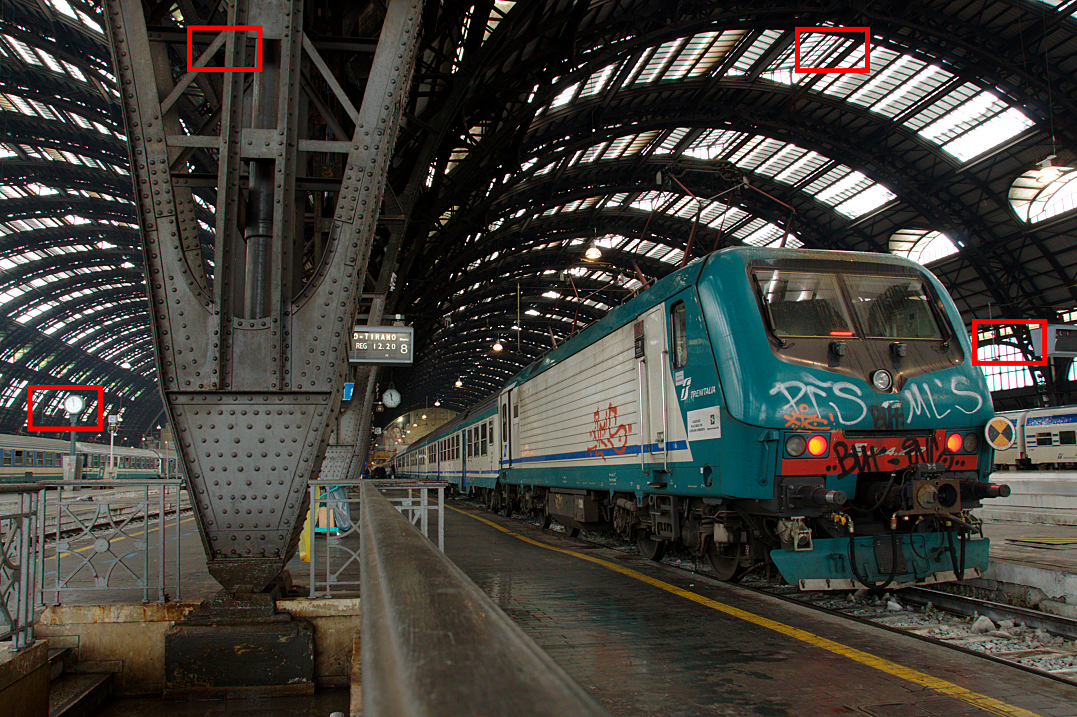}}\vspace{-7pt}
\subfigure{\includegraphics[width=0.3253\textwidth]{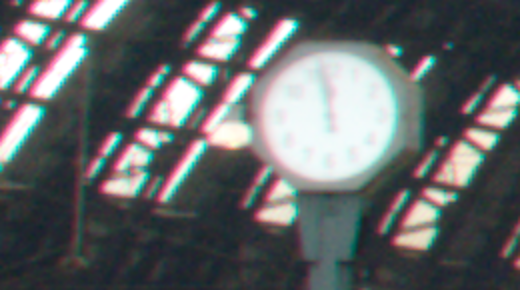}}
\subfigure{\includegraphics[width=0.3253\textwidth]{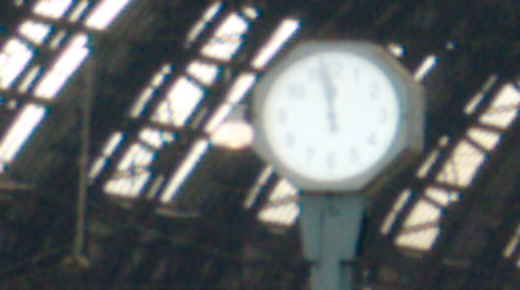}}
\vspace{-7pt}
\subfigure{\includegraphics[width=0.3253\textwidth]{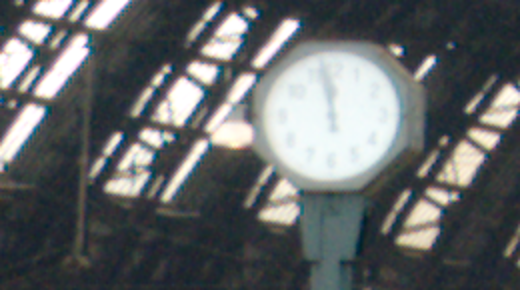}}
\subfigure{\includegraphics[width=0.3253\textwidth]{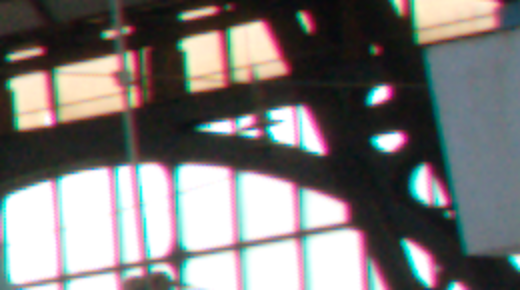}}
\subfigure{\includegraphics[width=0.3253\textwidth]{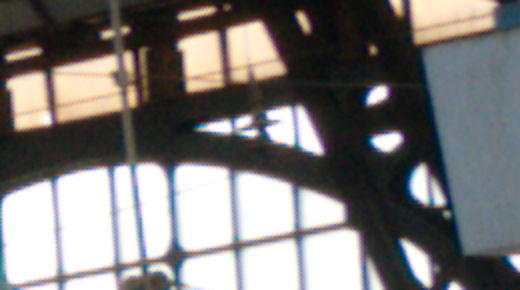}}
\vspace{-7pt}
\subfigure{\includegraphics[width=0.3253\textwidth]{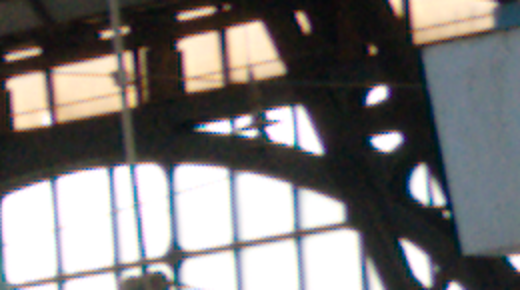}}
\subfigure{\includegraphics[width=0.3253\textwidth]{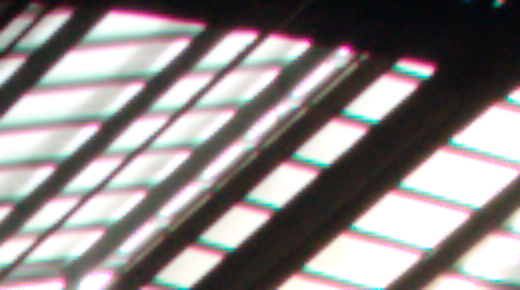}}
\subfigure{\includegraphics[width=0.3253\textwidth]{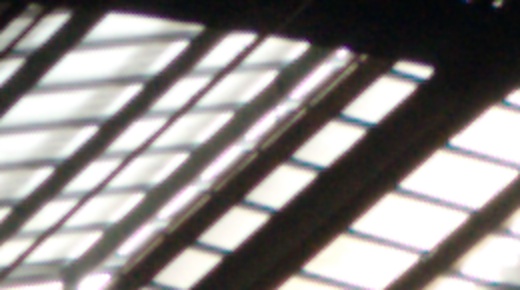}}
\vspace{-7pt}
\subfigure{\includegraphics[width=0.3253\textwidth]{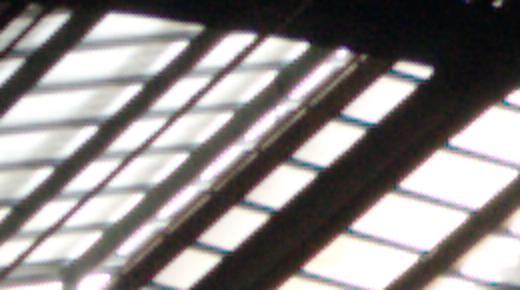}}
\subfigure{\includegraphics[width=0.3253\textwidth]{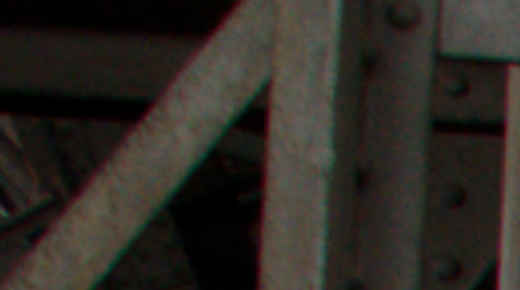}}
\subfigure{\includegraphics[width=0.3253\textwidth]{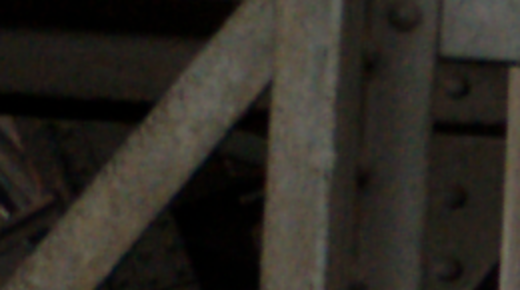}}
\vspace{-7pt}
\subfigure{\includegraphics[width=0.3253\textwidth]{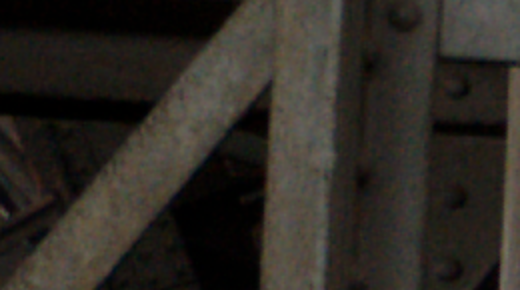}}
\caption[Cropped sections comparing correction methods]{Cropped sections from original (left) Photoshop (center) and proposed correction (right)}
\label{fig:comparison-first}
\end{figure}

\begin{figure}\centering\includegraphics[width=0.8\textwidth,trim=.3cm 0cm 1.6cm 1.0cm,clip]{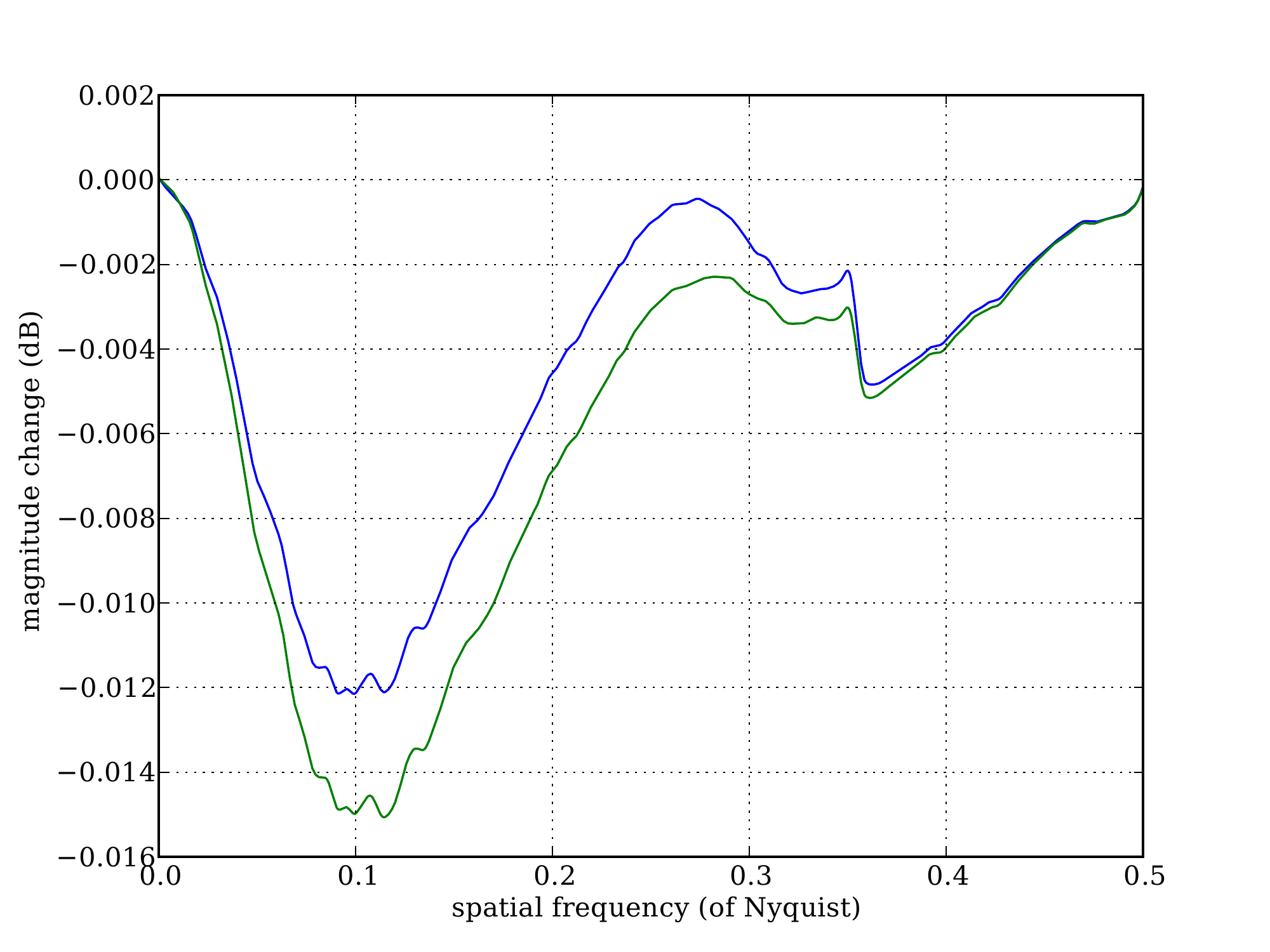}\caption[Change in spatial frequency in the first selected image]{Change in spatial frequency for Photoshop (blue) and proposed (green) correction in the first selected image}\label{gph:comparison-first}\end{figure}

\begin{figure}\centering
\subfigure{\includegraphics[width=\textwidth]{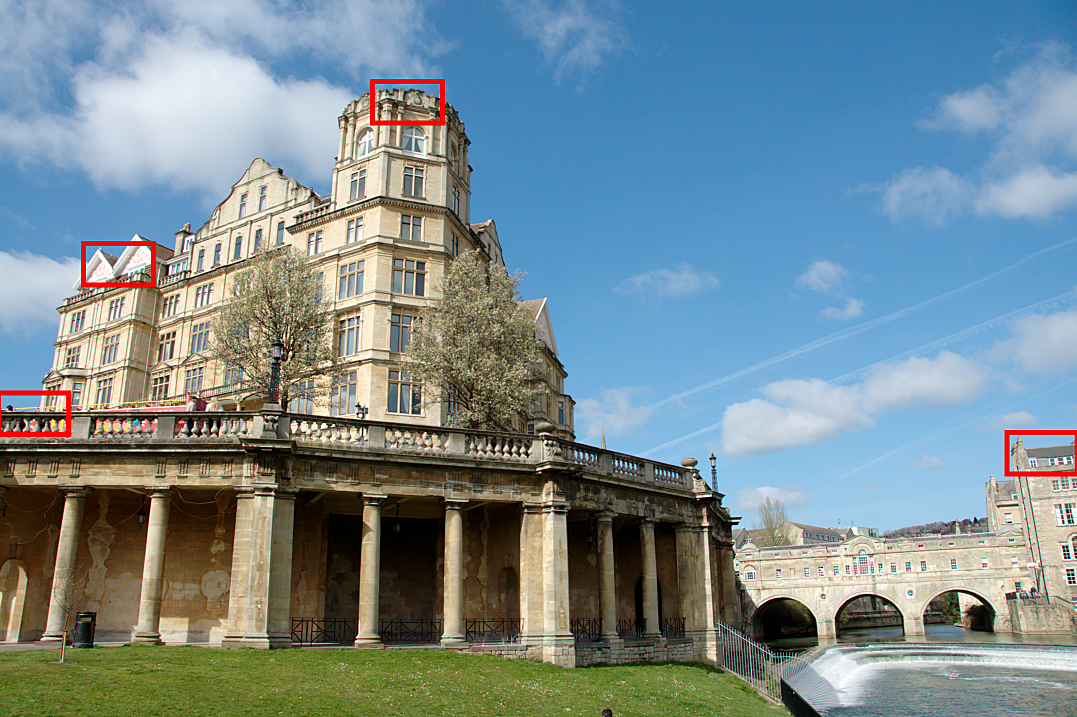}}\vspace{-7pt}
\subfigure{\includegraphics[width=0.3253\textwidth]{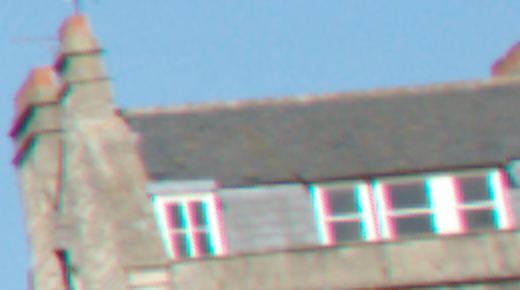}}
\subfigure{\includegraphics[width=0.3253\textwidth]{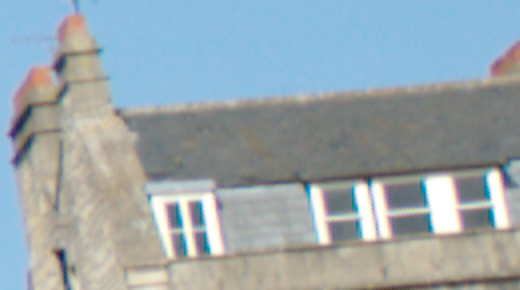}}
\vspace{-7pt}
\subfigure{\includegraphics[width=0.3253\textwidth]{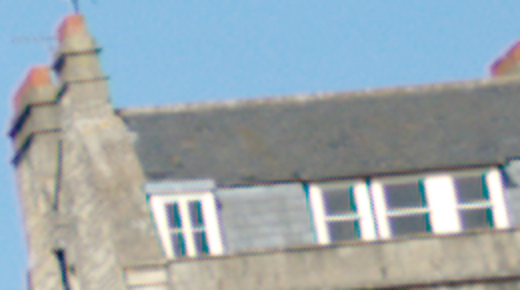}}
\subfigure{\includegraphics[width=0.3253\textwidth]{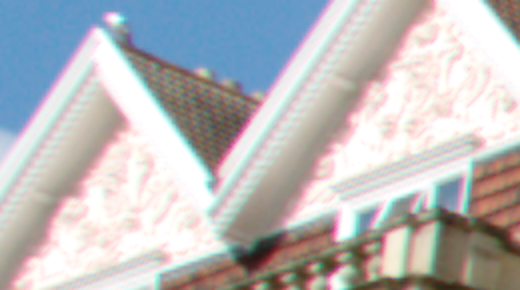}}
\subfigure{\includegraphics[width=0.3253\textwidth]{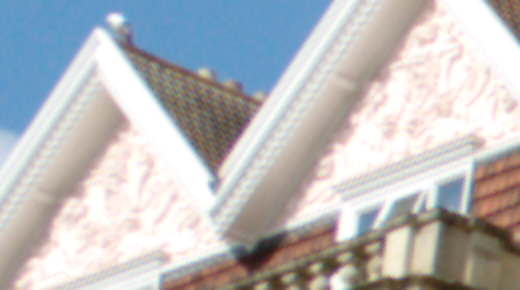}}
\vspace{-7pt}
\subfigure{\includegraphics[width=0.3253\textwidth]{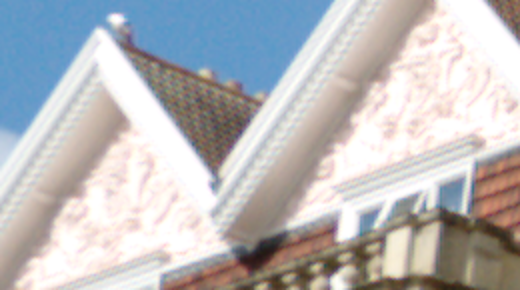}}
\subfigure{\includegraphics[width=0.3253\textwidth]{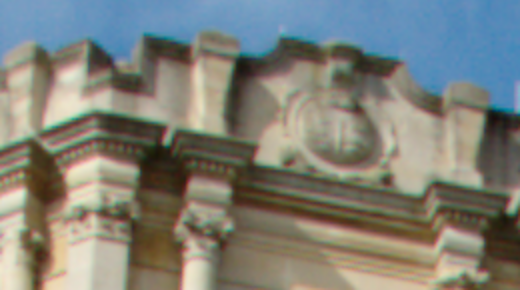}}
\subfigure{\includegraphics[width=0.3253\textwidth]{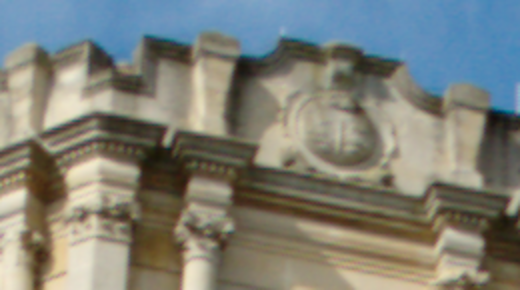}}
\vspace{-7pt}
\subfigure{\includegraphics[width=0.3253\textwidth]{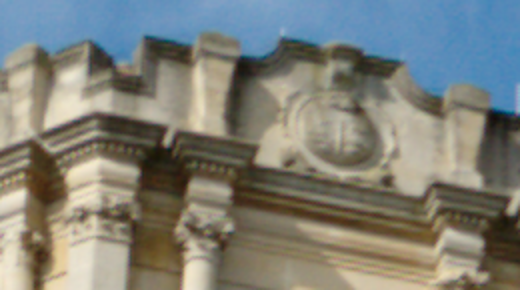}}
\subfigure{\includegraphics[width=0.3253\textwidth]{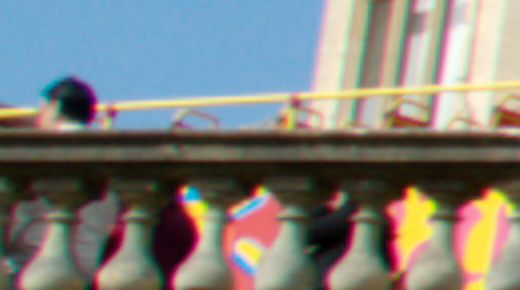}}
\subfigure{\includegraphics[width=0.3253\textwidth]{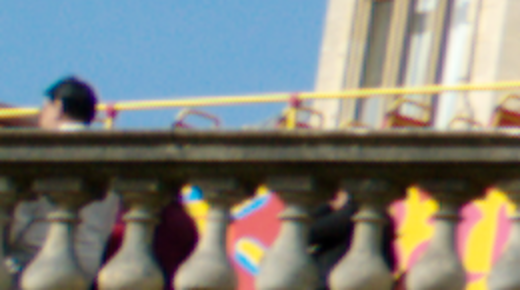}}
\vspace{-7pt}
\subfigure{\includegraphics[width=0.3253\textwidth]{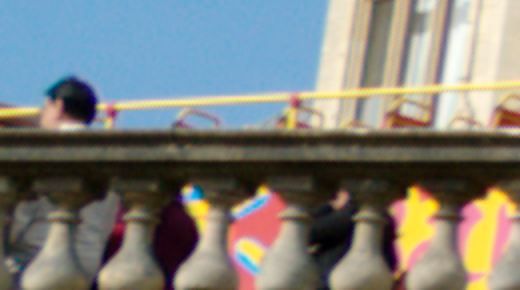}}
\caption[Cropped sections comparing correction methods]{Cropped sections from original (left) Photoshop (center) and proposed correction (right)}
\label{fig:comparison-second}
\end{figure}

\begin{figure}\centering\includegraphics[width=0.8\textwidth,trim=.3cm 0cm 1.6cm 1.0cm,clip]{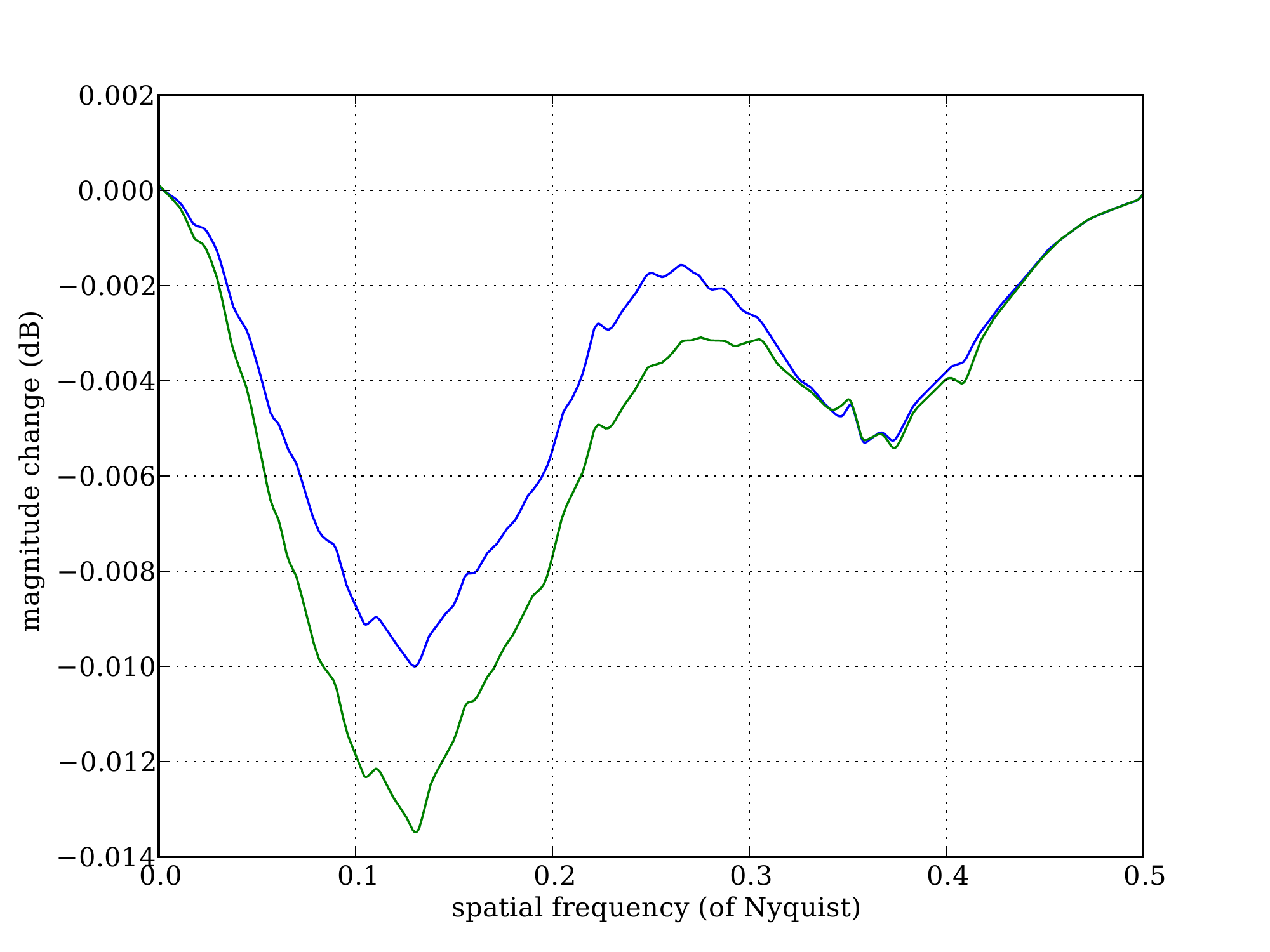}\caption[Change in spatial frequency in the second selected image]{Change in spatial frequency for Photoshop (blue) and proposed (green) correction in second selected image}\label{gph:comparison-second}\end{figure} 

\begin{figure}\centering
\subfigure{\includegraphics[width=\textwidth]{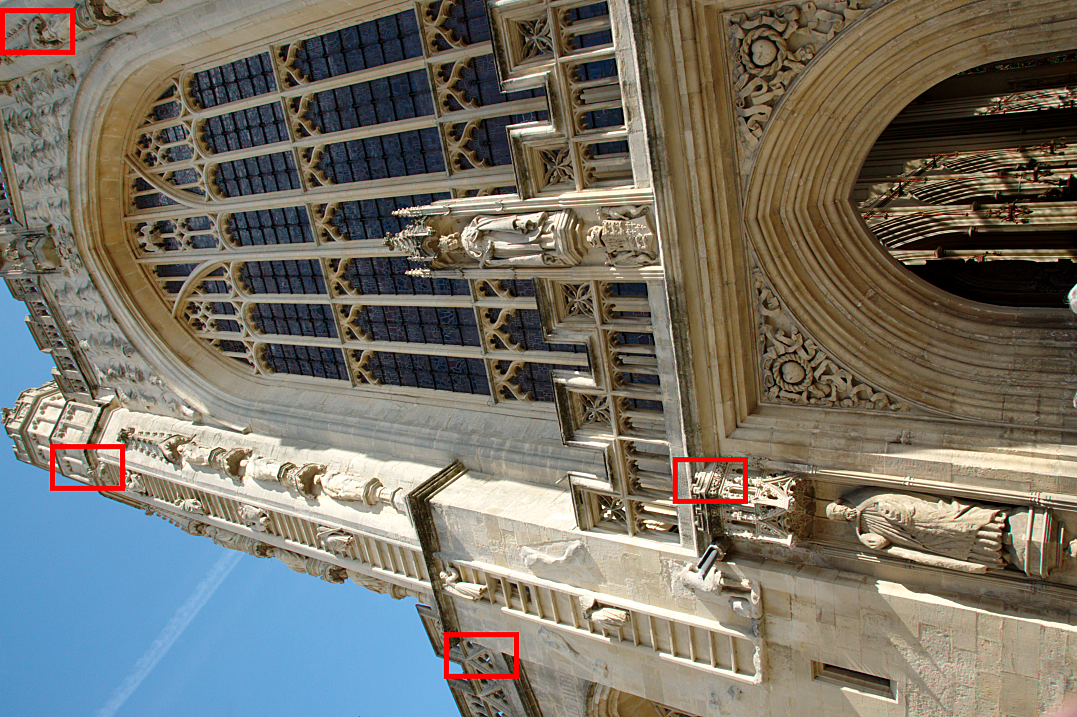}}\vspace{-7pt}
\subfigure{\includegraphics[width=0.3253\textwidth]{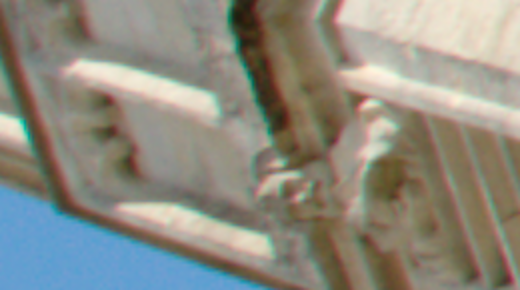}}
\subfigure{\includegraphics[width=0.3253\textwidth]{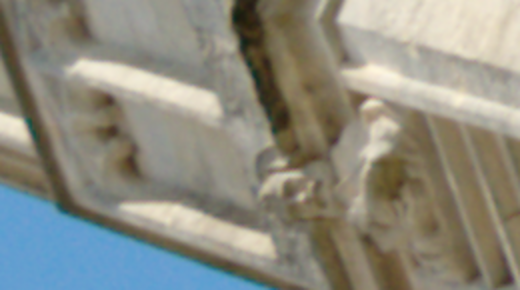}}
\vspace{-7pt}
\subfigure{\includegraphics[width=0.3253\textwidth]{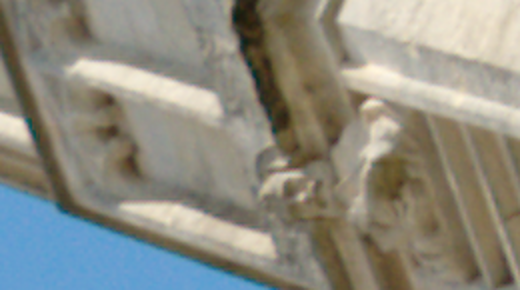}}
\subfigure{\includegraphics[width=0.3253\textwidth]{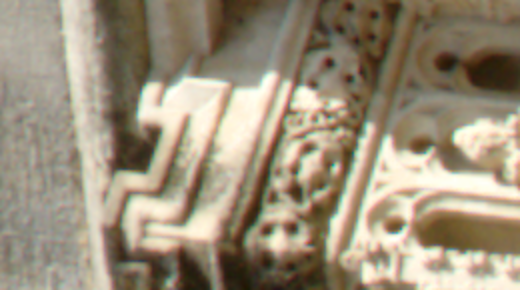}}
\subfigure{\includegraphics[width=0.3253\textwidth]{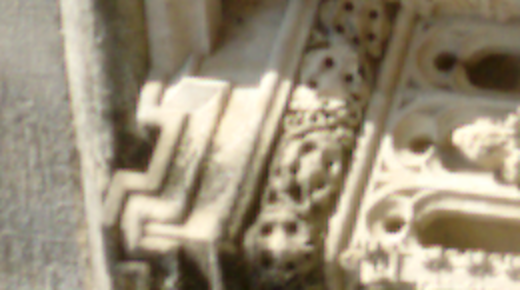}}
\vspace{-7pt}
\subfigure{\includegraphics[width=0.3253\textwidth]{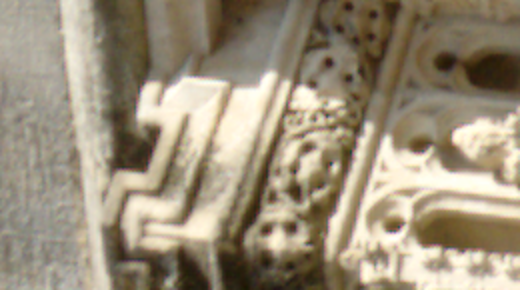}}
\subfigure{\includegraphics[width=0.3253\textwidth]{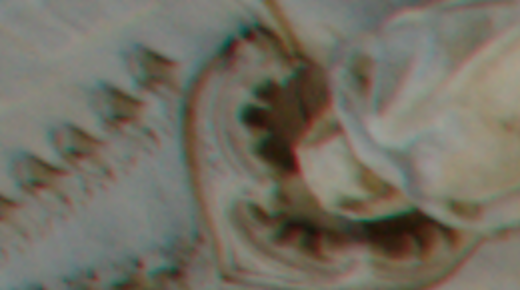}}
\subfigure{\includegraphics[width=0.3253\textwidth]{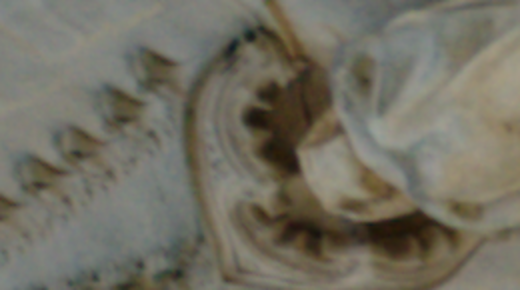}}
\vspace{-7pt}
\subfigure{\includegraphics[width=0.3253\textwidth]{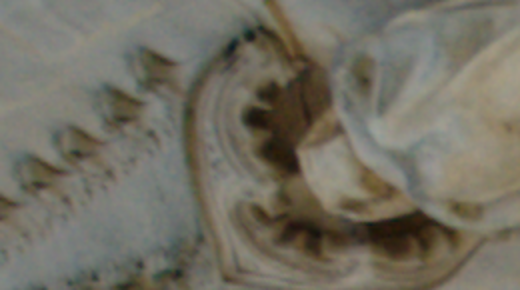}}
\subfigure{\includegraphics[width=0.3253\textwidth]{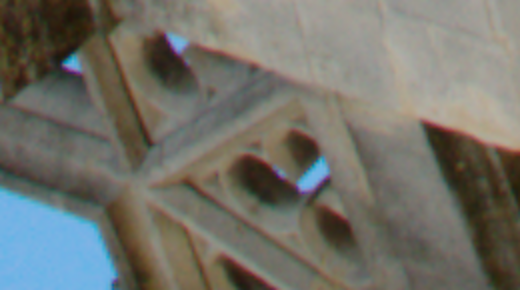}}
\subfigure{\includegraphics[width=0.3253\textwidth]{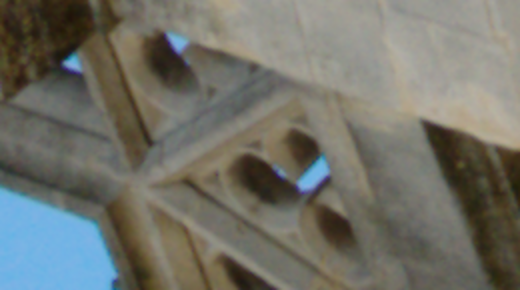}}
\vspace{-7pt}
\subfigure{\includegraphics[width=0.3253\textwidth]{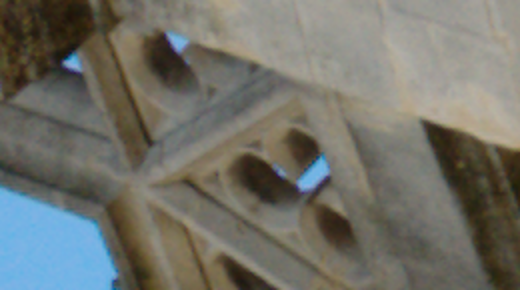}}
\caption[Cropped sections comparing correction methods]{Cropped sections from original (left) Photoshop (center) and proposed correction (right)}
\label{fig:comparison-third}
\end{figure}

\begin{figure}\centering\includegraphics[width=0.8\textwidth,trim=.3cm 0cm 1.6cm 1.0cm,clip]{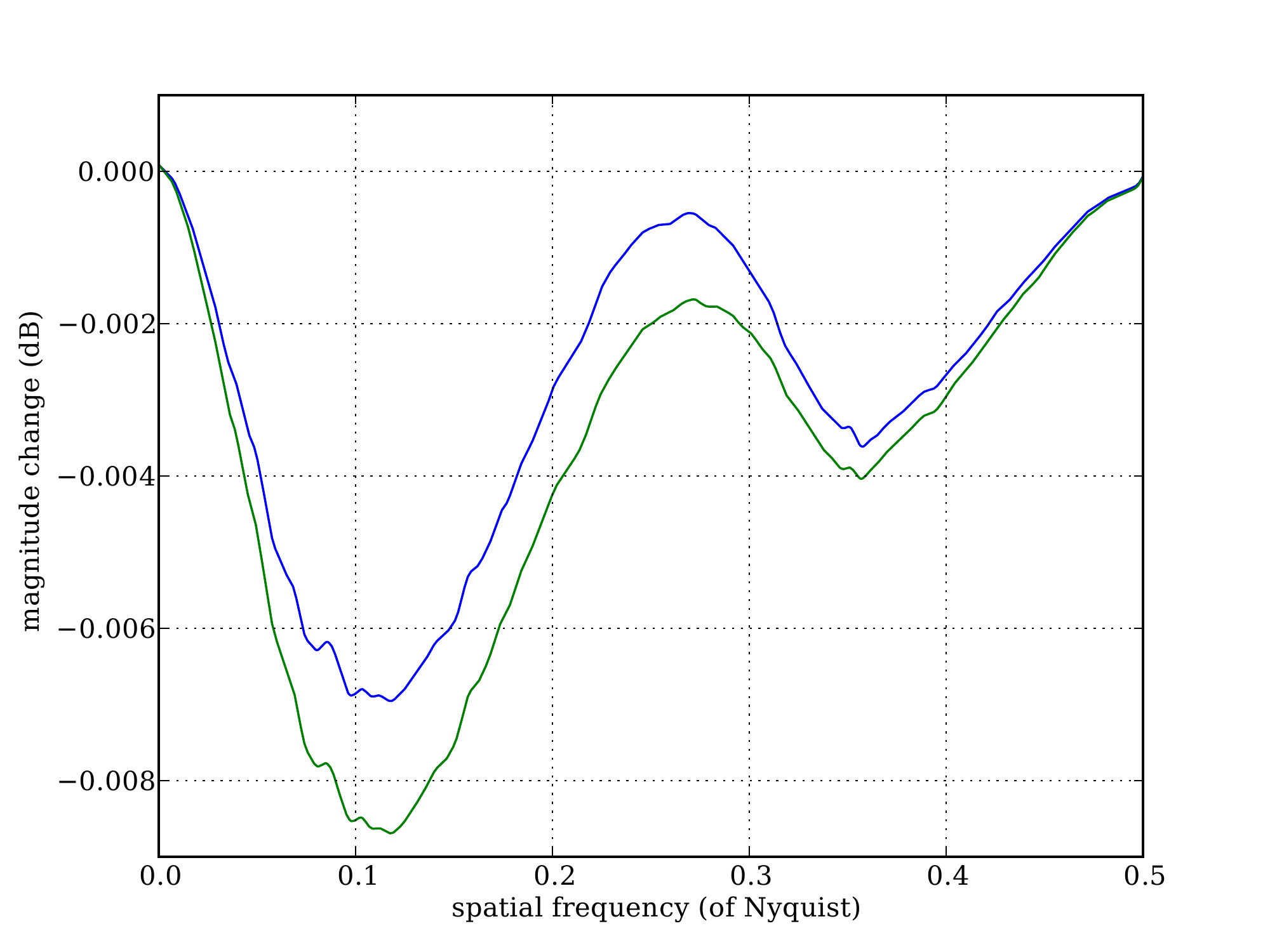}\caption[Change in spatial frequency in the third selected image]{Change in spatial frequency for Photoshop (blue) and proposed (green) correction in the third selected image}\label{gph:comparison-third}\end{figure}

\section{Results discussion}
By inspection of Figure~\ref{gph:testimages}, there are a number of characteristics observed, given by:

\begin{itemize}
\item Spatial frequency is observed to tend towards zero at 0~Hz and half-Nyquist. If this were observed to not be the case, error would have been introduced from the correction algorithm, or mismeasurements from the quantification algorithm
\item There is no spatial frequency (e.g.~texture complexity) increase, i.e.~all the changes serve to reduce spatial frequency; this corroborates with the basis of chromatic regions gaining \textit{homogeneity} and minimising additional spatial frequency due to the reduction of the resultant aliased artifacts
\item There are two observable peak reductions around $0.12 N$ and $0.35 N$, where $N$ is half the sensor Nyquist frequency; these would correspond to LCA features around 1.4 pixels and 4.1 pixels in width, which fits with the LCA feature size of around 4 pixels seen in Section~2
\item For this lens system, Figure~\ref{fig:improvement} shows that significantly more blue edge area is minimised than red edge area from Figure~\ref{fig:improvement}; this suggests that the lens system is optimised for lower LCA at redish wavelengths around 700~nm than blueish around 470~nm, which is a plausible design optimisation for the general use-profile of this lens. An areal reduction on average of 7.45\% was achieved
\end{itemize}

Another technique was initially evaluated and proven to give poor robustness for test image corrections; this algorithm generated a list of image features with Lowe's SIFT \cite{Lowe04} for each image plane, discarding outliers and employed feature correspondence to give an error measure, subsequently minimised via Least-Squares \cite{Marquardt63} to converge the image planes. It was found that there was an insufficient number of \textit{useful} image features to attain \textit{complete} and \textit{accurate} convergence in a significant number of the test cases; while others did not converge to the global minima, but to a local minima, presenting miscorrection. It was found that detecting discrete image features may not give coverage at a sufficient number of different radii from the image centre, which would optimise the image at a sub-range at expense of under-measured areas. The proposed technique avoids these issues by minimising the continuous planar difference, so each differing pixel contributes to the error measure. This provides a far more stable and predictable descent via the error function. Discrete keypoints were used by Cecchetto~\cite{Cecchetto09}, however the use of statistical correlation keypoint pruning was shown to introduce a level of uncertainty leading to subsequent correction error.

From the bank of test images, the proposed algorithm exhibited robust behaviour and showed no misconvergence. This is due to the distortion having radial homogeneity, making the approach sensitive to minima from features which are perpendicular to the line towards the centre of the image. As the lateral chromatic aberration is an image-global phenomena, after taking into account any lens decentering, it is experienced uniformly at all angles at a given distance from the centre of the image, thus local LCA features contribute small influence over the error minimisation surface. This property is fully exploited by performing the algorithm over the whole image, rather than just a subsector of the image, or on discrete image features. Whilst it is entirely possible to prepare a theoretical image which would defeat this mechanism e.g.~via red, green and blue concentric circles of differing but near radii, this has not occured in practice, or nor is likely to. Per-channel \textit{histogram equalisation} was introduced as a step in the algorithm to balance out the potentially uneven influence each channel e.g.\ due to a colour cast from a non-white light source.

For extremal or unexpected cases, the algorithm limits coefficient excursions to plausible values by employing bounded error minimisation \cite{Zhu94}. For any of the distortion polynomial components to be greater in magnitude than 0.05, creates a level of displacement that would only be useful in correcting an unrealistic level of LCA, thus limits are retained. Secondarily, this provides a set of constraints for the optimiser to work within, allowing searching for the minima to converge quicker than an unconstrained search \cite{Zhu94}.

At present, the demosaicing step is prior to the displacement correction, itself which uses linear interpolation to avoid introducing additional uncertainty and artifacts. From the uncorrected and corrected crops zipper-like artifacts are consequently visible at chromatic edges arising from the simple linear interpolation not exploiting spatial locality to detect local contours, as done in more advanced demosaicing algorithms, as previously addressed. With the impact of LCA present in a picture, at chromatic edges, advanced demosaicing would introduce false edges due to the demosaicer design goal of local homogeneity; should the correction step be performed prior to demosaicing, even further gains would be achieved.

\section{Summary}
Validation of the proposed LCA minimisation algorithm was performed incrementally with an artificial checkerboard pattern, followed by an artificial ray-traced image to prove basic operation. These were used to provide the basis for real-world testing, and provide an initial model within which the real-world examples would correlate. Following this, the initial example correction was presented, followed by correction of the test images.

The data from the image correction was quantified and graphed, allowing numerical inspection of the results. The data on the graphs show that higher-order luminance spatial frequencies were attenuated, indicating false aliasing from LCA was removed. Visual inspection confirmed this, and numeric comparison with a leading method showed a significant improvement with this novel approach.

\cleardoublepage
\chapter{Conclusion and further work}
\label{cha:future}
\section{Review}
In this work, the imaging performance of practical photography systems was reviewed and current limitations were examined. The poorly addressed and increasingly significant limitations that \textit{LCA} introduce were analysed in detail. Various solutions have been developed \cite{Boult92,Willson91,Kaufmann05,Cronk06,Mallon07,Heuvel06,Krause04,Luhmann06,Fryer86,Remondino06,Taehee07,Cecchetto09}, however many present limitations on \textit{generality}, \textit{application area} or impose impractical constraints for the correction of arbitrary images.

Real-world examples of LCA in photographs were presented along with discussion of parameter coverage. Next, the goals and scope of the solution needed were defined. The correction algorithm methodology was then broken down and presented into steps with justification.

A quantification framework was developed to measure LCA, then later the correction algorithm was incrementally validated with test subjects from a classic chequerboard test pattern and a high-complexity rendered scene. The quantification framework was used to measure the improvement to artificial and real-world test cases, and the performance of a leading LCA correction method was compared with the proposed method. Results were presented that showed visual elimination of LCA\@. Finally, the results including change in edge difference and spatial frequency were presented and discussed.

\subsection{Mechanics}
The algorithm for the correction of LCA presented in this work is based around minimising the sum of absolute difference between the reference green plane and the target red and blue planes; this achieves the goal of minimising LCA.

Firstly, preprocessing the source planes with histogram equalisation was performed to ensure the magnitude of the per-plane pixel values was comparable; this was introduced to address the problem of converging on false minima from images with strong colour casts, such as shot under sodium or other poor artificial lighting.

Convergence on minimal LCA is then achieved by iteratively varying the Taylor series coefficients in a radial distortion function; this is used to warp the target plane to align up with the reference plane, mitigating the lateral pixel displacement caused by differing paths due to diffraction. Error is measured by taking the average absolute difference of the target and reference planes for each pixel, with an inner boundary of the two planes to mask outlying pixels. L-BFGS-B descent was employed to take advantage of \textit{a priori} knowledge of extreme LCA limits that would not be seen in practise. This presented robust convergence and required fewer descent steps than other techniques, which were less suitable due to the non-linearity of this problem.

\begin{figure}\centering
\includegraphics[width=0.6\textwidth]{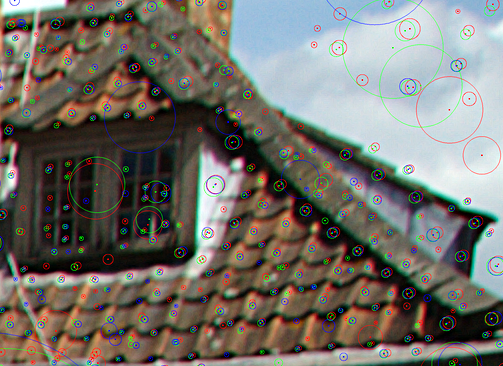}
\caption{SIFT used to detect feature points per channel}
\label{fig:lca-sift}
\end{figure}

Notably, early testing showed that using discrete feature points failed to provide adequate information for stable and accurate convergence on minimal LCA; Figure~\ref{fig:lca-sift} is an upper-right crop showing good feature-point correlation in the shadowed area under the roof, but disparity elsewhere. The primary advantage of the final algorithm is therefore the maximal use of image information to achieve accuracy and robustness through calculating inter-plane absolute difference. This also mitigated uncertainty and instability in the decent process.

\subsection{Results}
The results are separated into three areas. Firstly, from setting the context and presenting the algorithm steps and rationale, initial absolute difference crops given in Figure~\ref{fig:difference-comparison} show the design principle --- comparing the absolute difference in two planes on the LCA-uncorrected image and the LCA-corrected image. It can be seen that the corrected absolute difference map is overall darker, due to more coincident subtraction; this concurs with expectation.

Next, a simple testcase was constructed by rendering a classic chess-board calibration target. This was distorted with a Taylor-series polynomial and corrected; crops were compared to validate the algorithm. Visually, the chess-board converged to within sub-pixel expectations. Quantification in Figure~\ref{gph:pattern-comparison} showed correct inversion of the frequency changes, though some frequency loss was noted. Executing validation on a high-complexity ray-traced scene with spatial frequency exceeding Bayer array capability showed that convergence was visually accurate, such that no error could be observed in the crops. This was understood due to the increased level of available visual information, and highlights the architectural benefit of using maximum information from all pixels. Quantifying the frequency change in Figure~\ref{gph:rendering-improvement}, there is notable frequency loss due to aliasing error. Both observations of spatial frequency loss are not significant when dealing with images from Bayer arrays; the red and blue channels are sub-sampled by a factor of four within Bayer architecture. The green channel is passed through unmodified, so experiences no change.

Subsequently, LCA correction was performed on all of the images in the test set given in Figure~\ref{fig:test-images}. Overall inspection confirmed no divergent correction. As a result of misconvergence being observed in the twelfth test image due to coloured glass windows while developing the algorithm, analysis revealed descent to a false minima; the cause was found to be the uneven relative weighting of the image planes being differenced. Addition of the per-channel histogram equalisation step addressed this. A further step was taken to introduce non-linear bounded minimisation via L-BFGS-B minimisation algorithm.

Inspecting the corrected test images at pixel-level confirmed that convergence was accurate such that the visible chromatic error was wholly due to the simple (linear interpolation) demosaicing algorithm; this demonstrated that the minima was central within error from linear interpolation demosaicing. Quantification of all the images in Figure~\ref{gph:testimages} showed an interesting trend: that there are two peaks in spatial frequency reduction, at around 0.15 Nyquist and 0.35 Nyquist. This is understood to be due to the frequency and geometry of LCA feature size. Nevertheless, all pictures responded positively to improvement.

In order to test algorithm performance against a well understood existing approach, LCA correction of the images was undertaken in Photoshop CS4; best-effort manual correction was executed, and the same source image was processed with the proposed algorithm. Crops were extracted and compared. Visually, LCA is shown to be eliminated as far as possible in both corrected images, as expected. Employing quantification revealed a 20\% to 22\% peak reduction in spatial frequency over the Photoshop approach, thus converging on chromatic homogeneity down to the \textit{sub-pixel} level in an optimal manner.

\subsection{Summary}
A number of significant observations and findings were drawn from the outcome of this work:
\begin{itemize}
\item When correcting a large sequence of pictures, it was found that there were exactly three lens parameters at capture time that affected the LCA correction coefficients: \textit{focal plane}, \textit{focus distance} and \textit{aperture size}. The first parameter is natural, as there is movement of lens elements when this is varied. The last two parameters are less intuitive when varied, since focusing moves internal lens elements a short distance; changing the aperture size doesn't move lens elements. Critically, both do alter the path of light.
\item Detecting discrete feature points on the image colour planes and finding correspondence from the feature descriptor similarity was found to locate the order of hundreds of inter-plane stable features, and was used in the initial approach. Later this proved to be inadequate for \textit{accurate} convergence on the minimum-LCA solution. Following this, the architecture was modified to exploit areal differences to deliver the sensitivity and accuracy anticipated.
\item Performing LCA minimisation on a \textit{subsampled} image (e.g.\ quarter-size) demonstrated the same robust convergence on the true minimum-LCA solution, though an increase in error at a sub-pixel level was observed. Since performance was secondary to accuracy and correctness in this work, this was left aside as a future potential optimisation, using phased convergence with various pyramid sub-sample levels.
\item A number of other camera systems were tested during late development with only a small set of images; as expected, LCA correction showed correct convergence and produced satisfying results
\item The algorithm proved to deliver robust convergence on minimum LCA, while still allowing very wide convergence bounds for experimental lens systems
\item Convergence was accurate to the sub-pixel level to the extent that demosaicing was the limiting factor for image quality; this limitation could be surpassed by performing LCA minimisation and correction prior to demosaicing
\item From the set of test images shown in Figure~\ref{fig:test-images}, 3.86\% average edge reduction was found for the red plane and 12.1\% for the blue plane; this indicated the Nikkor lens used to capture the test images is optimised for lower red-green LCA than blue-green LCA
\item Comparable attenuation was observed at higher spatial frequencies in the proposed method; further analysis with a rendered target and Bayer simulation would allow accurate quantification of this loss, and a similar level was observed from the Photoshop method
\item Finally, significant image noise was found to not disrupt or hinder accurate convergence; this is considered due to the stochastic nature of sensor noise, averaging to zero across small patch areas, small with respect to the whole image area used in minimisation
\end{itemize}

\subsection{Limitations}
Due to the iterative and \textit{whole image} analysis and warping, the cost of each step is high, and thus the overall minimisation of LCA on an image is relatively expensive: with single-threaded execution, it took between 100 and 300 seconds to process an image. For this reason, the framework to re-use correction data was developed. It was noted that around 80\% of processing time (\textit{Pareto principle}) is spent in the spatial image remapping; since there is no data dependency, this is highly suitable for processing on a highly-parallel system e.g.\ a Graphics Processing Unit (GPU), and would achieve almost linear speedup.

This computational constraint precludes application onto an embedded hardware target, such as a camera, unless implemented in a custom image processing engine in silicon or with a Field Programmable Gate Array (FPGA)\@. This doesn't present a problem in practise, since the lenses available for a particular camera platform can be profiled and a database built, or correction performed offline.

Lastly, due to spatially remapping the red and blue planes, error is introduced from pixel aliasing; this was previously demonstrated with an artificial rendered image. While Bayer arrays are used in the vast majority of camera systems, some exist which capture full RGB data \textit{per-pixel}, such as using the Foveon X3 sensor. If a lens was employed which had a Nyquist frequency exceeding the sensor's Nyquist, correction would incur spatial frequency loss in the red and blue planes for areas in focus. This is considered a reasonable tradeoff, and is the cost of post-processing (i.e.\ non-optical) correction. This limitation was also seen in the Photoshop method.

\section{Claims and contributions}
The contributions of this work are divided into three areas:
\begin{itemize}
\item An algorithm delivering optimal LCA correction via robust convergence using arbitrary images. The associated error is minimised by the design of using all the information available from the image at every iteration, ensuring accuracy is maximised. The initial application of this work is photogrammetry, however this can equally apply to the areas found in broader literature, notably including microscopy and telescopy.

\item Mechanism to quantify the shift in spatial frequencies in a way useful to the HVS was developed. This provided a platform to develop and test the LCA correction algorithm, and allowed correction stability to be understood at low frequency and high up to half-Nyquist spatial frequency, where visual inspection would be clearly deficient.

\item A system for resulting correction data being associated with extracted image parameters for efficient offline correction was presented. Realising this in practise was enabled by locating near correction parameters, with further improvement possible via \textit{multi-variate} interpolation.
\end{itemize}

\section{Future work}
\subsection{Enhanced interpolation}
Though the presented algorithm uses Nearest Neighbour Search to locate the most suitable of correction coefficients, this is clearly non-optimal for two cases. Firstly, where the number of seen images and thus spread of correction coefficients is low, the average distance between lens parameter values is unfavourably high; this will correlate to higher average error, and may limit whole-system correction accuracy. Error from lens element position hysteresis and sensor discretisation are assumed small. Secondly, to achieve optimal correction, some level of coefficient interpolation must be used. There are a set of problems associated with this, which are briefly discussed: since the data is not at regular \textit{grid} intervals, it is considered sparse and thus classic linear interpolation between data points cannot be used. Instead, Delaunay triangulation must be performed in the three dimensions of the parameter space to locate the tetrahedron covering the location in which the point of interest lies \cite{Delaunay34}; this is the basis for Natural Neighbour interpolation \cite{Sibson81}. Further, polynomial interpolation must be used, as the correction coefficients cannot be interpolated individually. Methods such as Aitken-Neville \cite{Aitken32,Neville34} would be suitable.

Lastly, a simple step to achieve less information loss from pixel aliasing, is to use a more advanced interpolation algorithm for pixel remapping during warping, such as in the sinc family, e.g.\ Lanczos \cite{Lanczos50}.

\subsection{Parameter modelling}
It should be noted that as the error in the correction coefficients is low due to accurate convergence, interpolation between points is sufficient rather than construction of a model of the correction coefficients mapping onto the lens parameters. Algorithm~\ref{alg:coefficient-interpolation} is presented as a solution using first order linear interpolation.

\begin{algorithm}
\SetKwFunction{NearestNeighbour}{NearestNeighbour}
\SetKwFunction{Database}{Database}
\SetKwFunction{OpenDatabase}{OpenDatabase}
\SetKwFunction{lowerCoeffs}{lowerCoeffs}
\SetKwFunction{upperCoeffs}{upperCoeffs}
\SetKwFunction{PolyFit}{PolyFit}
\SetKwFunction{LoadImage}{LoadImage}
\SetKwFunction{return}{return}

\SetKwData{Image}{Image}
\SetKwData{ImageEXIF}{Image.EXIF}
\SetKwData{LensIdentifier}{LensIdentifier}
\SetKwData{Aperture}{Aperture}
\SetKwData{FocalLength}{FocalLength}
\SetKwData{FocusDistance}{FocusDistance}
\SetKwData{Database}{Database}
\SetKwData{Radius}{Radius}
\SetKwData{parameter}{parameter}
\SetKwData{ParameterSpace}{ParameterSpace}
\SetKwData{lowerDist}{lowerDist}
\SetKwData{upperDist}{upperDist}
\SetKwData{radius}{radius}
\SetKwData{interval}{interval}
\SetKwData{average}{average}
\SetKwData{newCoeffs}{newCoeffs}
\SetKwData{points}{points}
\SetKwData{Lower}{lower}
\SetKwData{Upper}{upper}
\SetKwData{VarA}{a}
\SetKwData{VarB}{b}

\SetKwInOut{Input}{input}
\SetKwInOut{Output}{output}

\Input{source image for LCA correction, complete with EXIF data}
\Input{database of previously computed correction coefficients for a range of images}
\Output{interpolated correction coefficients}

\Image $\leftarrow$ \LoadImage{}\;
\LensIdentifier, \Aperture, \FocalLength, \FocusDistance $\leftarrow$ \ImageEXIF\;
\Database $\leftarrow$ \OpenDatabase{\LensIdentifier}\;
\Radius $\leftarrow$ \Database{``MaxRadius''}\;
\tcp{graph average of both polynomials}
\For{\parameter $\in \{``Aperture'', ``FocalLength'', ``FocusDistance''\}$}{
	\tcp{locate two nearest neighbours in Euclidean space}
	\lowerDist $\leftarrow$ \NearestNeighbour{\Database, \parameter, \Lower}\;
	\upperDist $\leftarrow$ \NearestNeighbour{\Database, \parameter, \Upper}\;
	\BlankLine
	\tcp{graph average of both polynomials}
	\For{\interval $\leftarrow 0\ \KwTo$ \Radius}{
		\VarA $\leftarrow$ \lowerDist{\interval}\;
		\VarB $\leftarrow$ \upperDist{\interval}\;
		\average\big[\parameter\big] $\leftarrow \frac{\VarA \times (1 - \lowerDist) + \VarB \times (1 - \upperDist)}{2}$\;
	}
}

\tcp{use standard polynomial curve fitting}
\newCoeffs $\leftarrow$ \PolyFit{\points $\in$ \average}\;
\return{\newCoeffs}
\label{alg:coefficient-interpolation}
\caption{First-order coefficient interpolation}
\end{algorithm}

One solution for further consideration could be selected at the expense of implementation complexity and may be more suitable if convergence was inaccurate: parametric surface fitting to model the lens parameters. Earlier steps create database entries for a particular lens, storing the computed correction coefficients and relevant lens parameters at image-capture time. The correction coefficients are computed by minimising an error function which sums the vector of the fitted points to the ground truth data. Together, these are used to derive the correction coefficients for an unseen image. The full list of steps are: 

\textit{Database storage}: A database record is constructed with sufficient information to identify the camera lens and related information such as lens mounting (thus crop factor). This is later used to select the relevant correction parameters for a particular lens and mount.

\textit{Parametric surface fitting}: Once the data from each image in the input library has been stored into the database, each database record is accessed to identify what lens it was taken with; only images with the selected lens are considered in the model, thus images taken with other lenses are not considered and skipped.

A parametric surface equation is used to model the focal length and distortion relationship. For a given focal length, the experimental approach Fang et al \cite{Fang07} take, finds first-order fitting sufficient for their application; the optical design theory calculated by Gross et al \cite{Gross05} concurs with this. However it has been seen in practice, that simple first-order LCA correction is insufficient. In order to design in scope for this, a higher-order polynomial is used in both axes. Where second order is sufficient, the third order coefficients will be observed to be small.

The resulting parametric model with the constant term and cross-products removed gives six coefficients:

\begin{equation}
z = bx + cy + dx^2 + ey^2 + dx^3 + ey^3
\end{equation}

Marquardt's least-squares fitting \cite{Marquardt63} is used to converge the coefficients, minimising error.

\textit{Profile generation}: The parametric surface coefficients are stored along with the lens data, so can later be retrieved for offline correction.

\subsection{Implementation performance}
The last identified area for future work, is that an implementation leveraging GPU hardware would enable interactive. As previously revealed, around 80\% of the processing time is spent in the plane remapping function, of which is highly suited to GPU and highly-parallel architectures. Since there is no data dependency between the microkernels that process patches of destination pixels in the remapping function, the parallel scheduling and execution will efficiently use processing resources. With fine-grained hardware scheduling, throughput and efficiency far higher than on a general-purpose processor would be seen, resulting in low execution time.

\section{Summary and concluding remarks}
LCA correction is clearly a powerful and compelling step to reduce the loss that lens systems inevitably impose. It has been found that the proposed solution addresses the need of general high-fidelity photography and provides a solid foundation for further development, e.g.\ to realise faster processing or correction parameter interpolation. Since the presented algorithm can be applied in general to any imaging system, it has wide application across a variety of fields, including but not limited to the significant areas of telescope and microscope imaging.

The author believes in the free availability and openness of software and will make the source code available for use, along with this thesis. A future step would be to integrate this work with Zabolotny's \textit{Lensfun} optical correction library \cite{Zabolotny08}, which itself is used in a number of popular image manipulation tools for converting from raw sensor data. Photographers and scientists would be able to generate and further submit \textit{high-quality} lens LCA correction data for the library to automatically use, further realising the benefit. $\blacksquare$

\cleardoublepage
\phantomsection 
\addcontentsline{toc}{chapter}{Bibliography}
\bibliographystyle{abbrvnat}
\bibliography{thesis}

\end{document}